%% file: neurips_2026.tex
\documentclass{article}

\usepackage[main, final]{neurips_2026}

\input{preamble}

\author{%
  Stas Syrota, Albert Lopez I Serrano, Johanne Franck, Søren Hauberg \\
  Technical University of Denmark \\
  \texttt{\{stasy,alise,---,sohau\}@dtu.dk}
}

\title{Identifiable alignment of unpaired representations}

\begin{document}

\maketitle

\begin{abstract}
\input{sections/abstract}
\looseness=-1
\end{abstract}

\input{sections/intro}

\input{sections/mmspaces}

\input{sections/identifiability}

\input{sections/recovery}
\input{sections/ngma}

\input{sections/experiments}

\input{sections/related}
\input{sections/outro}

\paragraph{Acknowledgements.}
This work was supported by a research grant (42062) from VILLUM FONDEN. This project received funding from the European Research Council (ERC) under the European Union’s Horizon research and innovation programme (grant agreement 101125993). The work was partly funded by the Novo Nordisk Foundation through the Center for Basic Machine Learning Research in Life Science (NNF20OC0062606). We acknowledge \citep{DTU_DCC_resource} for awarding this project access to the LUMI supercomputer, owned by the EuroHPC Joint Undertaking, hosted by CSC (Finland) and the LUMI consortium.

\bibliography{bibfile}
\bibliographystyle{plainnat}
\newpage
\clearpage

\appendix
\input{sections/appendix}

\end{document}

%% file: preamble.tex
\usepackage[utf8]{inputenc} 
\usepackage[T1]{fontenc}    
\usepackage[hidelinks]{hyperref}       
\usepackage{url}            
\usepackage{booktabs}       
\usepackage{amsfonts}       
\usepackage{nicefrac}       
\usepackage{microtype}      
\usepackage{xcolor}         
\usepackage{algorithm,algpseudocode}
\usepackage{amsmath}
\usepackage{amssymb}
\usepackage{mathtools}
\usepackage{amsthm}
\usepackage{pifont}
\usepackage{lipsum}
\theoremstyle{plain}
\newtheorem{theorem}{Theorem}[section]
\newtheorem{proposition}[theorem]{Proposition}
\newtheorem{lemma}[theorem]{Lemma}
\newtheorem{corollary}[theorem]{Corollary}
\theoremstyle{definition}
\newtheorem{definition}[theorem]{Definition}

\theoremstyle{remark}
\newtheorem{example}[theorem]{Example}
\newtheorem{remark}[theorem]{Remark}

\usepackage{tabularx}
\usepackage{booktabs}
\usepackage{tikz}
\usetikzlibrary{positioning, arrows.meta, shapes.geometric}
\usepackage{enumitem}

\usepackage[most]{tcolorbox}
\tcbuselibrary{skins,breakable}

\tcbset{textmarker/.style={%
        enhanced,
        parbox=false,boxrule=0mm,boxsep=0mm,arc=0mm,
        outer arc=0mm,left=6mm,right=3mm,top=7pt,bottom=7pt,
        toptitle=1mm,bottomtitle=1mm,oversize}}

\newtcolorbox{hintBox}{textmarker,
    borderline west={6pt}{0pt}{yellow},
    colback=yellow!10!white}
\newtcolorbox{importantBox}{textmarker,
    borderline west={6pt}{0pt}{red},
    colback=red!10!white}
\newtcolorbox{noteBox}{textmarker,
    borderline west={6pt}{0pt}{green},
    colback=green!10!white}
\definecolor{figborder}{HTML}{D8A078}

\newtcolorbox{hypbox}{
  enhanced,
  breakable,
  frame hidden,
  colback=white,
  borderline west={2pt}{0pt}{figborder},
  left=3pt, right=0pt, top=0pt, bottom=0pt,
  boxsep=0pt,           
  before skip=3pt,      
  after skip=3pt,       
}

\newcommand{\rz}{\mathrm{z}}
\makeatletter
\newcommand{\stag}[1]{\textbf{(#1)}\phantomsection\def\@currentlabel{#1}\label{#1}}
\makeatother

\usepackage{todonotes}
\usepackage{url}

%% file: sections/abstract.tex
Aligning two models without paired data requires a distance that both share, yet existing methods use Euclidean or spherical geometry. This is either an arbitrary choice or imposed by their loss function. We show that, under a precise form of the manifold hypothesis, the pullback of the data metric gives such a distance for any two models of one data-generating process. This holds whether the models are encoders, decoders or both, and whatever their architecture, loss, prior or training data. The alignment map preserves this distance and the distribution of representations, and is identifiable if and only if the data distribution breaks every symmetry of the data manifold. We show that this symmetry breaking holds generically. Identifiability makes the alignment coupling the unique optimum of Gromov--Wasserstein and of its weighted combinations with a linear program. The linear program alone has the same optimum if almost every data point has a distinct distribution of distances. The coupling, and with it the alignment map, is recovered exactly from permuted samples, even without identifiability, and consistently from independent ones. Our method, Neural Gromov--Monge Alignment (NGMA), uses no pairs or labels and outperforms every unpaired and label-informed baseline.
It relies on pullback distances to align image models with different priors and single-cell models trained on disjoint cells, where Euclidean distances fail.

%% file: sections/intro.tex
\section{Introduction}\label{sec:intro}

Building multimodal foundation models \citep{merullo2023linearly}, sharing knowledge across data-privacy walls \citep{lotfollahi2022mapping}, and auditing proprietary vector-database security \citep{jha2025harnessing} all rely on learning an \emph{alignment map} between two representation spaces, which sends the representation of each data point in one space to the representation of the same data point in the other.
With paired data, this is a regression problem.
Without pairs, individual representations are meaningless in absolute terms.
What remains in each space are the relative distances between points and their distribution, which unpaired methods use to recover the alignment.

This idea rests on the Platonic representation hypothesis \citep{platonics}, which posits that independently trained models converge to the same \emph{similarity structure} on the data, defined as a kernel on pairs of data points computed from their representations.
Given two sets of representations, unpaired methods therefore learn a map that matches the kernels, in the hope that it is the alignment.

Recent methods align text encoders whose loss function places the representations on a sphere \citep{wang2020understanding,jha2025harnessing, minivec2vec}, where equal kernels make the alignment map a rotation.
Their cross-modal counterpart \citep{rosetta} extends this to vision encoders by imposing a spherical geometry on their representations before alignment.
The advances therefore predominantly cover one family of loss functions and two modalities, and whether they extend beyond these is an open problem.
The case to settle first is a single modality under arbitrary loss functions, which any account of representational convergence must pass before it can extend across modalities.

The immediate problem is that both signals, the relative distances and the distribution of the representations, depend on the choice of geometry.
Intuitively, leveraging them requires a distance tied to the intrinsic geometry of the data, invariant to arbitrary modelling choices.
Unfortunately, the field almost universally assumes the Euclidean or spherical geometry for the representations, which the data do not necessarily determine.
Standard contrastive losses build the sphere into training \citep{wang2020understanding}.
\citet{balestriero2026platonic} prove that, at their optimum, whitened joint-embedding predictive architectures (JEPAs) trained on the same finite set of data points with the same positive pairs share a metric, but only on those points, not on their generating distribution.
In both cases, the loss function sets the geometry, so models trained with different loss functions need not share it.
For general loss functions, models can freely reparameterise their representations, which distorts Euclidean distances \citep{syrota2025, hauberg2018onlybayes}.

\begin{figure}[!b]
  \vspace{-3mm}
  \centering
  \includegraphics[width=0.95\textwidth]{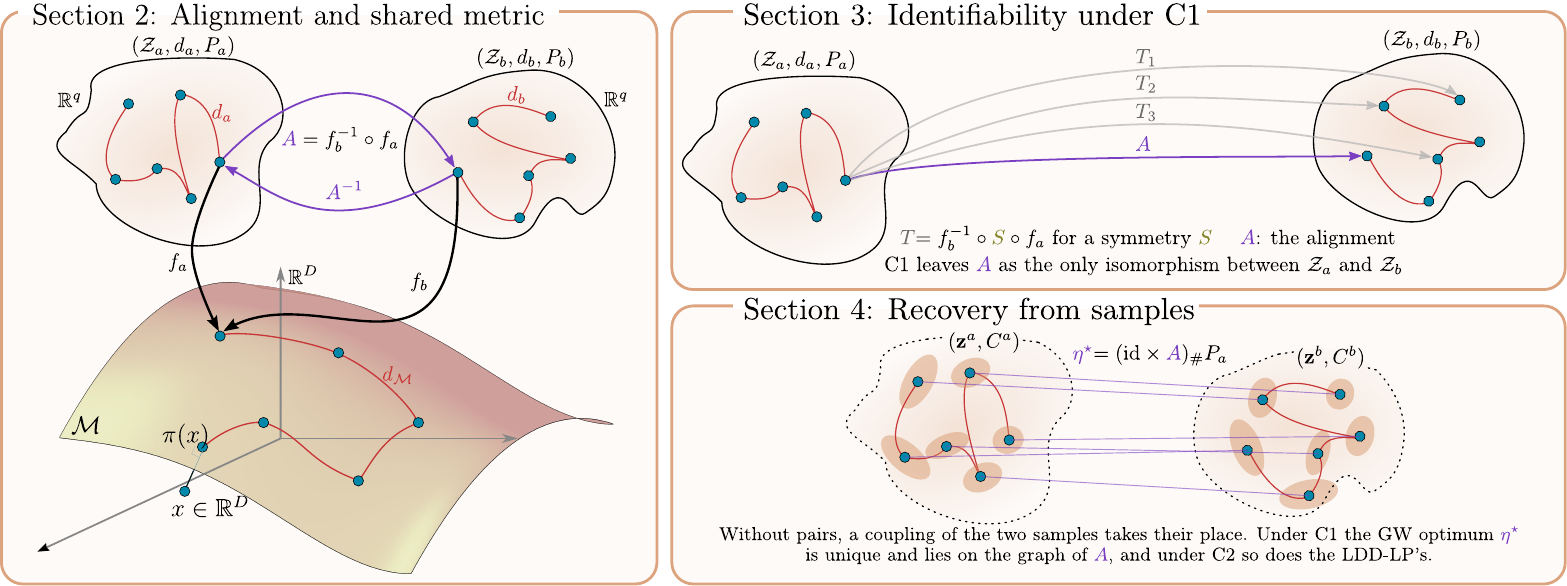}
  \vspace{-2mm}
  \caption{\textbf{Outline of Sections~\ref{sec:latentmmspaces}--\ref{sec:recovery}:} the setting, identifiability and recovery from samples.}
  \label{fig:overview}
\end{figure}

More fundamentally, the literature offers no general account of why trained models should share a similarity structure and what that structure is.
Without one, we cannot say how the alignment map is defined, whether a map that matches the kernels is unique, or whether its recovery is well-posed.

\textbf{In this paper}, we show that aligning two models of the same data without pairs is well-posed only when both representations induce the same metric on the data. A formalised manifold hypothesis defines this metric, and treating models as charts of the data manifold defines the alignment map.
With this metric, the map is uniquely identifiable if and only if the data distribution breaks the symmetries of the underlying manifold, a condition we prove holds generically.
We turn this identifiability into recovery guarantees by casting the problem as transporting one distribution of representations onto the other while preserving pairwise distances.
Finally, we instantiate our theory as an architecture-agnostic alignment method for representations of the same dimension.
Figure~\ref{fig:overview} visualises the main contributions of the paper, the relationships between them and how they are presented.

\textbf{Section~\ref{sec:latentmmspaces}} states the \emph{metric manifold hypothesis} (\ref{H1}) and treats each model as a chart of the data manifold (\emph{model coordinates}, \ref{H2}), so the alignment map is a coordinate change.
It preserves latent distances only if both models assign the same distances to data points, as the pullback of the data metric always does.
With the distribution of representations, each latent space becomes a \emph{metric measure space} (mm-space) \citep{gromov1999metric,sturm2006geometry}, and the alignment map an isomorphism.

\textbf{Section~\ref{sec:identifiability}} shows that $A$ is identifiable, the only isomorphism between the latent mm-spaces, if and only if no isometry of the data manifold other than the identity preserves the data distribution.
This \emph{symmetry breaking} (\ref{C1}) depends on the data-generating process alone and holds generically.

\textbf{Section~\ref{sec:recovery}} recovers $A$ by fitting a map to a coupling of the samples.
Under \ref{C1}, the coupling uniquely solves a non-convex transport problem \citep{memoli11}.
Under \emph{distinguishable points} (\ref{C2}), a linear program finds the same optimum. We show that fused GW, which combines the two, keeps the unique optimum under \ref{C1} without requiring \ref{C2}.
Recovery is exact when both models encode the same observations and consistent for independent samples.

\textbf{Section~\ref{sec:ngma}} instantiates the theory as Neural Gromov--Monge Alignment (NGMA), which computes the coupling with entropic optimal transport solvers \citep{peyre2016gromov, cuturi2013sinkhorn} and fits a map that preserves the pullback metric.
\textbf{Section~\ref{sec:experiments}} tests each claim on the sphere and on image and single-cell models.
NGMA outperforms unpaired and label-informed baselines, and the pullback geometry is what makes it work across priors and training cells.

%% file: sections/mmspaces.tex
\section{The metric manifold hypothesis defines the alignment and latent geometry}

\label{sec:latentmmspaces}

Representation learning is often motivated by the manifold hypothesis \citep{bengio2013representation}, which posits that the high-dimensional data concentrates near a lower-dimensional submanifold of the observation space.
This idea has proven effective in, e.g., learning causal representations \citep{Schoelkopf2021CausalRepresentation}, biological mechanisms \citep{scVI}, text embeddings \citep{mikolov2013efficient}, and image generation \citep{rombach2021highresolution}.

Each of these successes uses the learned structure as a space.
Clusters are found through distances, latent codes are interpolated along paths, and the representations of two models are compared by how far apart their points lie.
All of this implicitly requires the data manifold to be connected, so that any two points are joined, and compact, so that distances stay bounded.
It also requires the data to cover the manifold, since a model learns only the part of it that the data reach.
Our \emph{metric manifold hypothesis} (\ref{H1}) states these assumptions explicitly.

\begin{hypbox}
\textbf{Metric manifold hypothesis \stag{H1}.}
Observations $x\sim P_{\mathcal{X}}$ lie in $\mathcal{X}=\mathbb{R}^D$, equipped with a Riemannian metric $g^{\mathcal{X}}$, Euclidean unless stated otherwise.
We further assume:
\begin{enumerate}[label=(\roman*), leftmargin=*, itemsep=0pt, topsep=1pt]
  \item Data concentrate near a submanifold $\mathcal{M}\subset\mathcal{X}$ of dimension $\dim\mathcal{M}<D$ that is smooth, compact and connected, possibly with a boundary \citep{FeffermanMitterNarayanan2016, homology_learn}.\looseness=-1
  \item The noise-free data follow a distribution $P_{\mathcal{M}}$ with a smooth positive density
on $\mathcal{M}$.
  \item Each observation $x$ has a unique nearest point $\pi(x)$ on $\mathcal{M}$, and it is the
noise-free point $x$ came from.
This holds when noise moves points perpendicular to $\mathcal{M}$ by less than its reach, the largest distance within which every point has a unique nearest point on $\mathcal{M}$ \citep{federer1959curvature, estimate_reach}.
\end{enumerate}
\end{hypbox}
Distances on $\mathcal{M}$ are measured with the restriction $g^{\mathcal{M}}$ of $g^{\mathcal{X}}$.
The resulting shortest-path distance $d_{\mathcal{M}}$ is finite because $\mathcal{M}$ is connected and bounded because it is compact.
Compactness also makes the reach
positive, so (iii) allows noise of positive size. Projecting the observations onto $\mathcal{M}$ recovers the noise-free data (Remark~\ref{rem:projection}), so $P_{\mathcal{M}}$ is the distribution of $\pi(x)$ for $x\sim P_{\mathcal{X}}$, the \emph{pushforward} $\pi_{\#}P_{\mathcal{X}}$. A metric space with a probability measure is a \emph{metric measure space}, or mm-space \citep{gromov1999metric, sturm2006geometry}. We call $(\mathcal{M}, d_{\mathcal{M}}, P_{\mathcal{M}})$ the \emph{data mm-space}. 

What it means for a model to represent this structure is also usually left implicit.
Using latent distances as distances between data points presupposes that distinct data points have distinct representations and that nearby representations belong to nearby data points. Identifiability theory assumes this explicitly \citep{hyvarinen1999nonlinear, khemakhem2020variational, xi2023, syrota2025}.
Our \emph{model coordinates} hypothesis states it precisely:

\begin{hypbox}
\textbf{Model coordinates \stag{H2}.}
A model provides coordinates on $\mathcal{M}$ through a parameterisation, a diffeomorphism $f\colon\mathcal{Z}\to\mathcal{M}$ from a latent space $\mathcal{Z}$ of the same dimension onto $\mathcal{M}$, that is, a smooth bijection with a smooth inverse.
Then $\mathcal{Z}$ inherits the compactness, connectedness and boundary of $\mathcal{M}$.
A parameterisation can be built from either an encoder or a decoder (Lemma~\ref{lem:decoder-chart}):
\begin{enumerate}[label=(\roman*), leftmargin=*, itemsep=0pt, topsep=1pt]
    \item A decoder $\operatorname{dec}\colon\mathbb{R}^{q}\to\mathcal{X}$, injective with a
Jacobian of rank $q$ and image containing $\mathcal{M}$.
Then $f = \operatorname{dec}|_{\mathcal{Z}}$ is a parameterisation on $\mathcal{Z}:= \operatorname{dec}^{-1}(\mathcal{M})$ if $\mathcal{Z}$ is compact.

\item An encoder $\operatorname{enc}\colon\mathcal{X}\to\mathbb{R}^{q}$ with
$\operatorname{enc} = \operatorname{enc}|_{\mathcal{M}}\circ\pi$, injective on $\mathcal{M}$ with a Jacobian of rank $\dim\mathcal{M}$ there.
Then $f = (\operatorname{enc}|_{\mathcal{M}})^{-1}$ is a parameterisation on $\mathcal{Z}:= \operatorname{enc}(\mathcal{M})$.

\end{enumerate}
\end{hypbox}
Since $\dim\mathcal{M}$ is unknown, decoders usually have $q>\dim\mathcal{M}$, and compactness of $\mathcal{Z}$ keeps neighbouring data points from having distant representations.
We write $f_a, f_b$ for the parameterisations of two models and $\mathcal{Z}_a, \mathcal{Z}_b$ for their latent spaces.
Two models of the same data are two coordinate systems on one manifold, and aligning them is changing coordinates.
\begin{definition}[Alignment map]\label{def:alignment}
Under \ref{H1}--\ref{H2}, the \emph{alignment map} $A\colon\mathcal{Z}_a\to\mathcal{Z}_b$ is:
\begin{equation}\label{eq:alignment}
    A = f_b^{-1}\circ f_a.
\end{equation}
$A$ sends each representation in model $a$ to the representation of the same data point in model $b$.
\end{definition}

Although $A$ has a closed form, it cannot be evaluated from unpaired samples and must be inferred from distances and distributions.
Both depend on the geometry, and since $A$ only changes coordinates, it preserves a geometry only if both models measure the same distances on the data manifold $\mathcal{M}$.

\begin{proposition}[A shared data metric]\label{prop:shared-metric}
Under \ref{H1}--\ref{H2}, let $d_a$ and $d_b$ be metrics on $\mathcal{Z}_a$ and $\mathcal{Z}_b$.
The alignment preserves them, $d_b(A(z), A(z')) = d_a(z,z')$ for all $z,z'\in\mathcal{Z}_a$, if and only if both models assign the same distance to every pair of data points,
$$d_a\big(f_a^{-1}(x), f_a^{-1}(y)\big) = d_b\big(f_b^{-1}(x), f_b^{-1}(y)\big) \qquad\text{for all } x,y\in\mathcal{M}.$$
\end{proposition}
Euclidean or spherical latent distances satisfy this condition only when $A$ extends to a rigid motion, which is what the methods of Section~\ref{sec:intro} assume and what the data alone do not guarantee.
\citet{balestriero2026platonic} prove that at the optimum, whitened JEPAs trained alike on the same data points share a Euclidean metric on them.
This metric satisfies Proposition~\ref{prop:shared-metric}, but only models trained with the same loss on the same data points share it.

\textbf{Latent metric measure spaces.}
By Proposition~\ref{prop:shared-metric}, the metric has to be one that both models induce on $\mathcal{M}$ regardless of their training, and $\mathcal{M}$ already carries a canonical one, $d_{\mathcal{M}}$, from \ref{H1}.
Each model reaches it through its own Jacobian, via the pullback metric $g_a(u,v):= g^{\mathcal{M}}(\mathrm{d}f_a u, \mathrm{d}f_a v)$, written $f_a^*g^{\mathcal{M}}$.
The length $d_a$ of the shortest curve in $\mathcal{Z}_a$ under $g_a$ then satisfies $d_a(z,z') = d_{\mathcal{M}}(f_a(z),f_a(z'))$.
We pair it with the distribution of the representations, $P_a:= (f_a^{-1})_{\#}P_{\mathcal{M}}$, and likewise for $b$.
Two mm-spaces are \emph{isomorphic} if a bijection between them preserves distances and carries one measure onto the other.
By construction, $f_a$ is such an isomorphism from the \emph{latent mm-space} $(\mathcal{Z}_a, d_a, P_a)$ onto the data mm-space, and so is $f_b$.
Composing the two gives a key result:

\begin{proposition}\label{prop:alignment}
Under \ref{H1}--\ref{H2}, the alignment map is an isomorphism of the latent mm-spaces.
\end{proposition}

Unlike the Euclidean metric of whitened JEPAs, the pullback metric is shared by any two models satisfying \ref{H2}, whatever their architecture, loss function or training points.

%% file: sections/identifiability.tex
\section{Identifiability of the alignment map}
\label{sec:identifiability}

\begin{figure}[b]
  \centering
  \includegraphics[width=\textwidth]{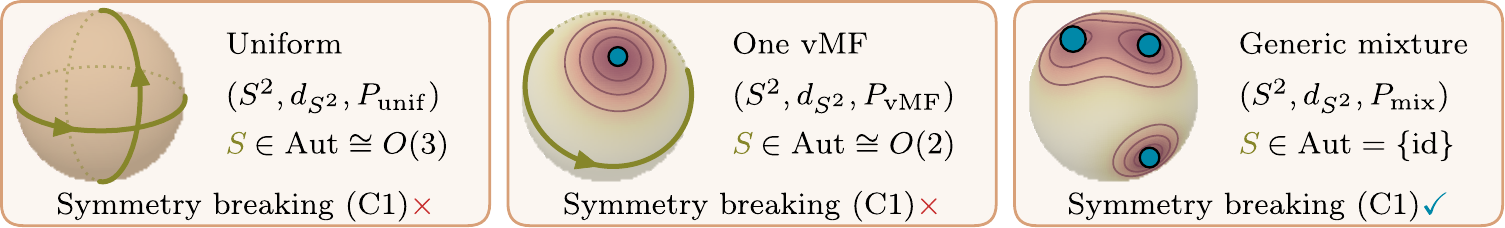}
  \vspace{-2mm}
  \caption{Three data mm-spaces on the round sphere, with the symmetries $S$ that preserve each distribution (olive), the vMF centres (teal, area proportional to weight) and iso-density contours. $O(3)$ consists of all isometries of the sphere, and $O(2)$ of those that fix the vMF centre.}
\label{fig:sphere-cases}
\end{figure}

Proposition~\ref{prop:alignment} tells us to search for an isomorphism of the latent mm-spaces, but not whether the search will find $A$.
Without pairs, the samples reveal only distances and distributions, which every such isomorphism, including $A$, matches.
If several exist, the search may return any of them.
We call $A$ \emph{identifiable} if it is the only isomorphism.
Since both latent mm-spaces inherit their geometry from the data mm-space, identifiability is a property of the data-generating process rather than of the models, and we show that it holds generically.\footnote{A property is generic if an arbitrarily small change can make it hold and no small change can break it. An example is invertibility of square matrices, where a small change of the entries makes a singular matrix invertible and keeps an invertible one invertible. Formally, it holds on an open and dense set.}
\looseness=-1

To illustrate when multiple isomorphisms exist, we build a running example of three data mm-spaces, each the sphere $S^2\subset\mathbb{R}^3$ with the round metric and a different distribution (Figure~\ref{fig:sphere-cases}).
The hypersphere is the geometry that contrastive losses \citep{wang2020understanding} and alignment methods \citep{rosetta} impose on representations, even when the data manifold itself is not a sphere.
The example therefore mirrors such models, but all results hold for any data mm-space satisfying the metric manifold hypothesis (\ref{H1}).

On each sphere, we take two parameterisations, $f_a = \mathrm{id}$ and $f_b = R^{-1}$ for a fixed rotation $R$, so model $b$ represents every data point rotated by $R$ and the alignment map is $A = f_b^{-1}\circ f_a = R$.
Rotations preserve the round metric, so the latent mm-spaces $(S^2, d_{S^2}, P)$ and $(S^2, d_{S^2}, R_{\#}P)$ are isomorphic.
For the uniform distribution, $P = P_{\mathrm{unif}}$ (Figure~\ref{fig:sphere-cases}, left), every rotation $S$ other than the identity moves points but leaves the distances between them and their distribution unchanged.
The map $f_b^{-1}\circ S\circ f_a = RS$ is then an isomorphism of the latent mm-spaces just like $A$, and the search may return either.
Concentrating the data around one point with a von Mises--Fisher (vMF) distribution $P_{\mathrm{vMF}}$ \citep{fisher1953dispersion} rules out some of these symmetries (middle), and the mixture $P_{\mathrm{mix}}$ of three vMF distributions rules out the rest (right), so $A$ is the only isomorphism (Appendix~\ref{app:vmf-rigidity}).

A \emph{symmetry} of the data mm-space is a bijection $S\colon\mathcal{M}\to\mathcal{M}$ that preserves distances and $P_{\mathcal{M}}$, and the symmetries form the \emph{automorphism group} $\operatorname{Aut}(\mathcal{M}, d_{\mathcal{M}}, P_{\mathcal{M}})$.
Each symmetry $S$ yields an isomorphism $f_b^{-1}\circ S\circ f_a$ that is inseparable from the alignment map without pairs, but differs from $A$ unless $S$ is the identity.
The following condition ensures that $A$ is identifiable:

\begin{hypbox}
\textbf{Symmetry breaking \stag{C1}.}
The identity is the only symmetry of the data mm-space.
\end{hypbox}

It is also necessary, since every isomorphism of the latent mm-spaces comes from a symmetry:

\begin{proposition}
\label{prop:identifiable}
Under \ref{H1}--\ref{H2}, the isomorphisms of the latent mm-spaces are exactly the maps $f_b^{-1}\circ S\circ f_a$ with $S\in\operatorname{Aut}(\mathcal{M}, d_{\mathcal{M}}, P_{\mathcal{M}})$, so $A$ is identifiable if and only if \ref{C1} holds.
\end{proposition}

The condition may seem demanding, since the uniform sphere already violates it.
Yet symmetric densities are the exception on any data manifold:

\begin{proposition}
\label{prop:trivial_automorphism}
Fix $\mathcal{M}$ and $g^{\mathcal{M}}$.
The densities in $C^\infty_{>0}(\mathcal{M})$ for which the data mm-space satisfies \ref{C1} form an open, dense subset in the $C^\infty$ topology.
\end{proposition}

A symmetry $S$ is fragile, because the density $\rho$ of $P_{\mathcal{M}}$ must satisfy $\rho\circ S=\rho$ at every point.
An arbitrarily small bump at a point that $S$ moves, but not at its image, breaks $S$.
On the sphere, a first vMF bump on $P_{\mathrm{unif}}$ breaks every symmetry that moves its centre.
The rotations about its axis and the reflections in planes containing it remain, as for $P_{\mathrm{vMF}}$ (Figure~\ref{fig:sphere-cases}, middle).
A second, smaller bump off that axis breaks these rotations, and only the reflection in the plane through both centres remains.
A third, even smaller bump off that plane breaks this reflection, which maps the third centre to a point without a bump (right).
On any $\mathcal{M}$, well-placed bumps lower the dimension of the continuous symmetries until none remain.
Since the isometry group of $\mathcal{M}$ is compact, finitely many symmetries are left, and one more bump breaks each.
Densities that satisfy \ref{C1} are therefore dense (Appendix~\ref{app:proofs}).

Conversely, densities that satisfy \ref{C1} form an open set, because a small change cannot create a symmetry.
Every isometry of the sphere far from the identity changes the density of $P_{\mathrm{mix}}$ somewhere by at least a fixed amount.
A small rotation changes it at a rate set by the slope of the density along the rotation, which is nonzero since no rotation preserves $P_{\mathrm{mix}}$.
A perturbation whose values and slope stay well below this amount and this rate, such as a slight shift of a centre, creates no symmetry.
On any $\mathcal{M}$, the argument uses only that the isometries form a compact Lie group, like $O(3)$.

The pullback metric is what makes identifiability a property of the data-generating process alone.
With Euclidean or spherical distances, $A$ preserves them only if $A$ is a rigid motion, which limits alignment to models trained with the same dot-product loss (Section~\ref{sec:latentmmspaces}).
Even then, $A$ is unique only if the models spread their representations without symmetry.
Contrastive losses and the LeJEPA regulariser do the opposite.
They push the representations towards the uniform distribution on the hypersphere or towards $\mathcal{N}(0,I)$, which every orthogonal map preserves \citep{wang2020understanding, balestriero2025lejepa}.
Where these losses give two models a shared metric, they do so through this symmetry, which leaves $A$ inseparable from $QA$ for every orthogonal $Q$.

%% file: sections/recovery.tex
\section{Recovery of the alignment from finite samples}
\label{sec:recovery}

Under symmetry breaking (\ref{C1}), which holds generically, the latent mm-spaces determine $A$, so estimating it without pairs is well-posed.
We see only independent samples of the two latent mm-spaces, with no one-to-one correspondence between them, so regression is out of reach.
Instead, we first estimate a \emph{coupling} and then fit a map to it, akin to neural optimal transport methods \citep{klein2024genot}.
We now discuss recovery guarantees for this two-stage procedure.

A coupling is a probability measure $\eta\in\Pi(P_a,P_b)$ on $\mathcal{Z}_a\times\mathcal{Z}_b$ with marginals $P_a$ and $P_b$.
Since the alignment map is an isometry, the coupling induced by $A$ preserves all pairwise distances.
We therefore estimate it by solving the \emph{Gromov--Wasserstein} (GW) problem \citep{memoli11}:
\begin{equation}\label{eq:GWdist}
\min_{\eta\in\Pi(P_a,P_b)} J_{\mathrm{GW}}(\eta),\qquad J_{\mathrm{GW}}(\eta):= \iint \big|d_a(z,z') - d_b(w,w')\big|^2 \,\mathrm{d}\eta(z,w)\,\mathrm{d}\eta(z',w').
\end{equation}
It seeks the coupling that distorts distances least, and under \ref{C1} it has exactly one solution:

\begin{proposition}\label{prop:unique-gw-coupling}
Under \ref{H1}--\ref{H2}, the optimal couplings of Equation~\ref{eq:GWdist} are exactly the couplings $(\operatorname{id}\times T)_{\#}P_a$ induced by isomorphisms $T$ of the latent mm-spaces.
Under symmetry breaking (\ref{C1}), the optimal coupling is unique and equals $\eta^\star = (\operatorname{id}\times A)_{\#}P_a$.
\end{proposition}

Since $A$ is continuous and $P_a$ has full support, the coupling $\eta^\star$ is supported on the graph of $A$, so $\eta^\star$ determines the map.
Under \ref{C1}, both stages therefore have a unique target, $\eta^\star$ for the coupling and $A$ for the map, so the two-stage procedure is well-posed.
The proofs are in Appendix~\ref{app:proofs}.

\subsection{Computing the coupling}
\label{sec:ldd_sinkhorn}

The uniqueness in Proposition~\ref{prop:unique-gw-coupling}, however, does not guarantee that a solver finds the optimum.
The GW problem is non-convex \citep{kravtsova2026complexity}, so solvers return stationary points that need not be global optima.
We therefore turn to Mémoli's third lower bound on GW \citep{memoli11}, which compares the \emph{local distance distributions} (LDDs) of points instead of their pairwise distances.
It is a convex problem, so every stationary point is a global optimum, but it is coarser than GW.

The LDD of a point $z\in\mathcal{Z}_a$ is the distribution function of its distance to a random representation, $\Phi_a(z,r):= P_a\big(\bar B_a(z,r)\big)$, where $\bar B_a(z,r)$ is the closed ball of radius $r$ around $z$, and likewise $\Phi_b$ on $\mathcal{Z}_b$.
Under \ref{H1}--\ref{H2}, both latent mm-spaces are isomorphic to the data mm-space, so $\Phi_a(z,\cdot)=\Phi(x,\cdot)=\Phi_b(A(z),\cdot)$ for the data point $x=f_a(z)=f_b(A(z))$, where $\Phi(x,r):=P_{\mathcal{M}}\big(\bar B_{\mathcal{M}}(x,r)\big)$ is its LDD in the data mm-space.
The alignment map thus preserves LDDs, so we can match them instead of distances by solving
\begin{equation}\label{eq:sinkhorn_problem}
\min_{\eta\in\Pi(P_a,P_b)} J_{\mathrm{LDD}}(\eta),\qquad J_{\mathrm{LDD}}(\eta):= \int W_2^2\big(\Phi_a(z,\cdot),\Phi_b(w,\cdot)\big)\,\mathrm{d}\eta(z,w).
\end{equation}
Equation~\ref{eq:sinkhorn_problem} is a linear program in $\eta$, which we call the LDD-LP, and its cost is the squared $L^2$ distance between the quantile functions of the two LDDs.

Each LDD collapses a point's distances to all others into one distribution, so the LDD-LP tells points apart less finely than GW.
If two data points share an LDD, the LDD-LP can exchange them at no cost, so we require almost every point to have a unique LDD:

\begin{hypbox}
\textbf{Distinguishable points \stag{C2}.}
The LDD map $x\mapsto\Phi(x,\cdot)$ is injective $P_{\mathcal{M}}$-almost everywhere, i.e., the collision set $\mathcal{C}:= \{x: \Phi(x',\cdot) = \Phi(x,\cdot) \text{ for some } x'\neq x\}$ is $P_{\mathcal{M}}$-null.
\end{hypbox}

\ref{C2} implies \ref{C1}, because a symmetry preserves every LDD, but not conversely, since almost every point can share its LDD with another without a symmetry (Appendix~\ref{app:ldd-multi-scale}).
If the data mm-space has distinguishable points, we can use the LDD-LP to find the coupling $\eta^\star$:

\begin{proposition}\label{prop:convex-alignment}
Under \ref{H1}--\ref{H2} and \ref{C2}, $\eta^\star$ is the unique optimal coupling of the LDD-LP.
\end{proposition}

The two problems trade the fidelity of GW against the convexity of the LDD-LP, and neither has both.
We propose to combine them as in fused GW \citep{vayer2019optimal}, with the LDD-LP in place of the feature term and a weight $\beta\in[0,1]$ on it:
\begin{equation}\label{eq:fused}
J_\beta(\eta):= (1-\beta)\,J_{\mathrm{GW}}(\eta)+\beta\,J_{\mathrm{LDD}}(\eta).
\end{equation}
Both terms are nonnegative and vanish at $\eta^\star$, so under \ref{C1}, $\eta^\star$ is the unique minimiser of $J_\beta$ for $\beta<1$ (Proposition~\ref{prop:fused-unique}), and under \ref{C2} also for $\beta=1$.
Therefore, a solver can lower $\beta$ from 1 to 0 without moving the minimiser, but $J_\beta$ is non-convex for $\beta<1$, and we do not prove that the path ends at $\eta^\star$.

Since LDDs are estimated from samples and carry sampling error, they must be not only distinct but well separated.
\ref{C2} itself cannot be checked from samples, but the smallest distance between the LDDs of two data points at least $t$ apart can be estimated:

\begin{definition}\label{def:margin}
For $0 < t \le \operatorname{diam}(\mathcal{M})$, the \emph{margin} of the data mm-space at separation $t$ is
\begin{equation}\label{eq:margin}
m(t):= \inf\big\{\,\|\Phi(x,\cdot) - \Phi(x',\cdot)\|_\infty \;:\; x,x'\in\mathcal{M},\ d_{\mathcal{M}}(x,x')\geq t\,\big\}.
\end{equation}
\end{definition}

From samples, the DKW inequality \citep{DKW, massart2} gives a level-$\delta$ margin test of whether two query points at separation $t$ share an LDD.
A rejection says nothing about the pairs the test does not query, so it can occur where $m(t)=0$.
In low dimension, however, the queries can cover $\mathcal{M}$, and the estimate then bounds $m(t)$ from below if the LDDs are Lipschitz (Appendix~\ref{app:dkw}).

\subsection{Finite-sample guarantees}
\label{sec:finiterecovery}

In practice we see only samples, $\mathbf{z}^a = (\rz^a_1, \dots, \rz^a_n)$ in $\mathcal{Z}_a$ and $\mathbf{z}^b$ in $\mathcal{Z}_b$, with distance matrices $C^a_{ij} = d_a(\rz^a_i, \rz^a_j)$ and $C^b$.
The samples define the empirical distributions $\hat P_a$ and $\hat P_b$, and with the distance matrices, they carry the two signals that remain without pairs and determine the empirical LDDs.
We minimise $J_\beta$ for some $\beta \in [0,1]$ over the couplings of $\hat P_a$ and $\hat P_b$, which gives the empirical version of the problem.

In the \emph{independent regime}, the two samples are drawn independently from $P_a$ and $P_b$, so almost surely no point of one is the image of a point of the other.
In the \emph{permutation regime}, both models encode the same observations in unknown order, so that $\rz^b_{\sigma(i)} = A(\rz^a_i)$ for a permutation $\sigma$.

The permutation regime is the easier one, because $A$ is an isometry and the two distance matrices therefore agree up to the unknown order, $C^b_{\sigma(i)\sigma(k)} = C^a_{ik}$.
Almost surely the pairwise distances of the sample are distinct, so only $\sigma$ reorders one matrix into the other or pairs points with equal empirical LDDs (Lemma~\ref{lem:multiset}).
The empirical problem therefore recovers $\sigma$ exactly for every $\beta$:

\begin{proposition}\label{prop:perm-exact}
In the permutation regime, under \ref{H1}--\ref{H2} and with $n\geq 3$, the empirical $J_\beta$ almost surely has the unique optimal coupling $\frac1n\sum_i\delta_{(\rz^a_i,\,\rz^b_{\sigma(i)})}$ for every $\beta\in[0,1]$.
\end{proposition}

The proposition needs neither \ref{C1} nor \ref{C2}, so it holds even for the uniform distribution $P_{\mathrm{unif}}$ on the sphere (Figure~\ref{fig:sphere-cases}), where no method identifies $A$ from independent samples.
With $\sigma$ recovered, fitting $A$ is supervised regression, and $\dim\mathcal{M}+1$ pairs fix the alignment among isometries if they are in general position and close enough that the shortest paths between them are unique (Corollary~\ref{cor:pinning}).

In the independent regime, no coupling of the two samples is exact at finite $n$, and only consistency remains (Proposition~\ref{prop:consistency}).
Global minimisers of the empirical $J_\beta$ converge to $\eta^\star$ under \ref{C1} for $\beta<1$, and under \ref{C2} also for $\beta=1$, by compactness and the uniqueness of the population minimiser (Propositions~\ref{prop:unique-gw-coupling}, \ref{prop:convex-alignment} and~\ref{prop:fused-unique}).
In each case, a map fitted to the coupling over a uniformly compact class containing $A$ converges to $A$.
Consistency leaves two questions open.
First, it covers only global minimisers, while solvers return stationary points that need not be global.
Second, it gives no rate of convergence.

%% file: sections/ngma.tex
\section{Neural Gromov--Monge Alignment (NGMA)}
\label{sec:ngma}
Our Neural Gromov--Monge Alignment (NGMA) method instantiates the two-stage recovery procedure with choices informed by Section~\ref{sec:recovery}.
Its input is the samples $\mathbf{z}^a$, $\mathbf{z}^b$, their distance matrices $C^a$, $C^b$, and the pullback metric tensors $G_a$, $G_b$ at the sample points.

NGMA computes the coupling $\hat\eta$ with entropic GW \citep{peyre2016gromov} applied to $J_\beta$ in four stages, $\beta = 1$, $0.8$, $0.5$ and $0$, that lower the entropy weight together with $\beta$, each starting from the coupling of the previous one (Appendix~\ref{app:ngma}).
The first stage is the convex LDD-LP at $\beta=1$ and the last is entropic GW at $\beta=0$, so NGMA returns a stationary point of entropic GW, reached from the solution of the convex problem rather than from the independent coupling.

Given $\hat\eta$, NGMA estimates $A$, a smooth isometry that carries $P_a$ onto $P_b$ (Proposition~\ref{prop:alignment}).
Such a map solves the non-convex Gromov--Monge problem \citep{memoli2022}, which NGMA avoids by fitting a smooth map to $\hat\eta$, the coupling that Section~\ref{sec:recovery} recovers.

A smooth isometry is also a bijection.
Architectures with a closed-form inverse would build this in, but they restrict the class of maps, so NGMA fits two plain networks, $\phi$ for $A$ and $\psi$ for $A^{-1}$.
Each is an affine map plus a residual network, with the affine part initialised at the Procrustes solution $\min_{Q,v}\sum_{i,j}\hat\eta_{ij}\|Q\rz^a_i+v-\rz^b_j\|^2$ \citep{schonemann1966procrustes}, or its inverse for $\psi$, and the residual at zero.
Both are trained on $\mathcal{L}(\phi,\psi)=\ell_{ab}(\phi,\psi)+\ell_{ba}(\psi,\phi)$, where
\begin{equation}\label{eq:ngma-objective}
\ell_{ab}(\phi,\psi)=\frac{1}{n_a}\sum_{i}\bigg[\frac{\|J_\phi^\top G_b\,J_\phi-G_a\|_F^2}{\|G_a\|_F^2}+\lambda_{\mathrm{fit}}\|\phi(\rz^a_i)-\tau^a_i\|^2+\lambda_{\mathrm{cyc}}\|\psi(\phi(\rz^a_i))-\rz^a_i\|^2\bigg],
\end{equation}
$J_\phi$ and $G_a$ are evaluated at $\rz^a_i$, $G_b$ at the point of $\mathbf{z}^b$ nearest to $\phi(\rz^a_i)$, $\tau^a_i$ is the $\rz^b_j$ with the largest $\hat\eta_{ij}$, and $\ell_{ba}$ exchanges the roles of $a$ and $b$. Weights are set at $\lambda_{\mathrm{fit}}=0.3$ and $\lambda_{\mathrm{cyc}}=0.1$.

The first term penalises deviations from $A^*g_b=g_a$, the last pushes $\psi$ towards $\phi^{-1}$, and among maps that nearly satisfy both, the fit term favours the one that follows the coupling, which at $\eta^\star$ is $A$.

We hold out $n_0$ random points of $\mathbf{z}^a$ and every point of $\mathbf{z}^b$ whose largest entry of $\hat\eta$ falls on one of them, train on the remaining $n_a$ and $n_b$ points, and keep the checkpoint at which the most held-out points map to a mutual nearest neighbour in the other sample.
The coupling's targets are wrong for many individual points, so, as in unsupervised word translation \citep{conneau2018word, artetxe2018robust}, we refit both maps for up to three rounds on the pairs they agree on, the mutual nearest neighbours under $D(i,j)=\omega_a\|\phi(\rz^a_i)-\rz^b_j\|+\omega_b\|\psi(\rz^b_j)-\rz^a_i\|$.
Both distances are divided by their medians, and each weight is proportional to the squared fraction of mutual nearest neighbours its map finds.
We keep the round with the most held-out mutual nearest neighbours under $D$ and rank test candidates by $D$ (Appendix~\ref{app:ngma}). 
Code is available at \url{https://github.com/mustass/identifiable-alignment}.

%% file: sections/experiments.tex
\section{Experiments}
\label{sec:experiments}

First, we test the claims of Sections~\ref{sec:latentmmspaces}--\ref{sec:ngma} on the sphere, where we choose $A$, and the geometry is in closed form.
Then we train VAEs \citep{VAE} on three image datasets (MNIST \citep{lecun1998gradient}, FashionMNIST \citep{xiao2017fashion} and CelebA \citep{liu2015deep}) with a Gaussian prior or a VampPrior \citep{pmlr-v84-tomczak18a}, and scVI models \citep{scVI} on Tabula Sapiens lung cells \citep{tabulasapiens2022}. 

On real data, we split each dataset into three disjoint subsets: one to train the models, one to fit the alignment and one to test it.
All image models are trained on the same training set, across priors and seeds, while the lung models are trained on disjoint sets of cells.
We draw the two samples from the alignment subset, in regimes ranging from disjoint to fully shared (Section~\ref{sec:finiterecovery}), and repeat this with different seeds.
NGMA and the baselines are fitted to each draw and scored on the test subset by instance retrieval (R@1), the fraction of test points whose own representation in the other model is ranked first.
For all models, we measure latent distances with the pullback metric of an ensemble of eight decoders \citep{syrota2024decoder}, pulling back the Euclidean metric of pixel space for the images and the Fisher metric of scVI's negative-binomial likelihood for the cells.
Conceptually, the ensemble keeps the geometry on the compact latent space $\mathcal{Z}$ of \ref{H2}, since its metric grows where the decoders disagree, which is away from the data \citep{hauberg2018onlybayes, syrota2024decoder}.
Instead of computing geodesics, we use the lengths of straight segments under this metric, which bound the geodesic distances from above, need no optimisation per pair and lose little accuracy (Appendix~\ref{app:geodesic-segments}).
On the sphere, the two samples are drawn from the measure itself, and we score the fitted rotation against the true $R$.
Each paragraph below states one finding, with the main results in Table~\ref{tab:alignment} and Figures~\ref{fig:sphere} and~\ref{fig:lung_main}.

\input{figures/sphere.tex}

\textbf{Symmetry breaking (\ref{C1}) decides identifiability.}
On each sphere of Figure~\ref{fig:sphere-cases}, model $a$ represents the data as they are and model $b$ rotates them by a fixed $R$, so $A=R$.
We add a mixture with equal weights and concentrations, which satisfies \ref{C1} but whose LDDs collide along curves, so its margin is zero (Appendix~\ref{app:ldd-multi-scale}).
The NGMA coupling, the LDD-LP and GW are solved on $k$ landmarks, chosen so that their LDDs differ as much as possible (Appendix~\ref{app:exp-sphere}).
From independent samples, the LDD-LP recovers $R$ to within $5^\circ$ for both mixtures, while for $P_{\mathrm{unif}}$ and $P_{\mathrm{vMF}}$ the error stays large however many landmarks we use (Figure~\ref{fig:sphere}a).
Symmetry breaking thus decides recovery, as Proposition~\ref{prop:identifiable} states, even for the mixture whose margin is zero.
The margin test never rejects for the two symmetric measures (Figure~\ref{fig:sphere}c, Appendix~\ref{app:dkw}), but has limited power.

\textbf{With the pullback metric, models are coordinate charts (\ref{H2}).}
On MNIST and FashionMNIST, models with different priors differ by $10$ to $25$ nats in likelihood (Table~\ref{tab:equivalence}).
Yet their pullback distances correlate above $0.97$, while their Euclidean distances correlate at only $0.71$ and $0.61$ (Table~\ref{tab:invariance}).
Here, the alignment is highly nonrigid and every method that fits a rotation falls behind NGMA (Table~\ref{tab:alignment}).
The lung models are not rigidly aligned either, since a rigid map fitted on the true pairs stays far below a map fitted on them (Figure~\ref{fig:lung_main}).

\textbf{The permutation regime is easy with the shared metric (Proposition~\ref{prop:perm-exact}).}
With permuted samples, LDD-LP recovers $R$ exactly for every measure, including the uniform one, from $3$ landmarks on, as the proposition predicts.
When both models encode the same images, NGMA reaches R@1 of $0.990$--$0.999$ in every MNIST and FashionMNIST setting. In contrast, with a GW coupling on Euclidean distances, it reaches only $0.353$ and $0.679$ on the cross-prior pairs (Appendix~\ref{app:exp-overlap}).

\input{tables/main_results.tex}

\textbf{NGMA leads in every setting, from disjoint to fully shared samples.}
Our baselines without pairs or labels are Rosetta \citep{rosetta}, mini-vec2vec \citep{minivec2vec}, GENOT-Q \citep{klein2024genot} and LFM \citep{fumero2024latent}, and label-informed RelReps \citep{moschella2023relative} and RelGeo \citep{yu2024connecting}.
In the hardest regime, the two samples share no image, so no coupling is exact and only consistency remains (Section~\ref{sec:finiterecovery}).
Yet NGMA leads every baseline in all nine settings of Table~\ref{tab:alignment}, and keeps the edge when samples have images in common (Appendix~\ref{app:exp-overlap}, Table~\ref{tab:shared}).

\textbf{Identifiability is a property of the data-generating process, not the training data.}
Nothing in \ref{H1}--\ref{H2} requires the two models to share training points, so we align lung models trained on disjoint sets of cells, which share only the tissue (Figure~\ref{fig:lung_main}, Appendix~\ref{app:lung}).
Without pairs, our method retrieves the true partner of up to $38\%$ of ten thousand held-out cells, where chance is one in ten thousand, and the right cell type for nine in ten.
NGMA is above every baseline on every sample, RelGeo with cell-type labels included.
With Euclidean distances, it fails on most draws.
With the pullback metric of the scVI decoder alone, instead of the ensemble, it is lower on $17$ of $18$ draws.

\input{figures/lung_main.tex}

\textbf{Near-symmetries obstruct alignment and fused GW (Equation~\ref{eq:fused}) helps resolve them.}
On CelebA, a left-right flip nearly preserves the distribution of faces, so \ref{C1} nearly fails.
On every failed draw, with either coupling, the maps send faces to their mirror images (mirror R@1 of $0.87$--$0.99$, true R@1 at most $0.11$).
On every successful draw, they send faces to themselves (Appendix~\ref{app:mirror}).
The mirror alignment preserves distances and nearly preserves the distribution of representations, so without pairs it is almost indistinguishable from $A$.
Of the ten draws that fail with GW alone, NGMA's coupling rescues three and loses none, so it succeeds on $28$ of $35$ (Figure~\ref{fig:celeba-coupling}).
On $P_{\mathrm{mix}}$ on the sphere, GW alone leaves $4$ to $8$ of $10$ seeds more than $90^\circ$ from $R$ at every $k$ up to $256$.
NGMA's coupling has no such seed from $k=8$, thanks to its LDD-LP term, although the LDD-LP alone has the smaller median error for $k$ between $32$ and $256$ (Figure~\ref{fig:sphere}b).

%% file: figures/sphere.tex
\begin{figure}
  \centering
  \includegraphics[width=\textwidth]{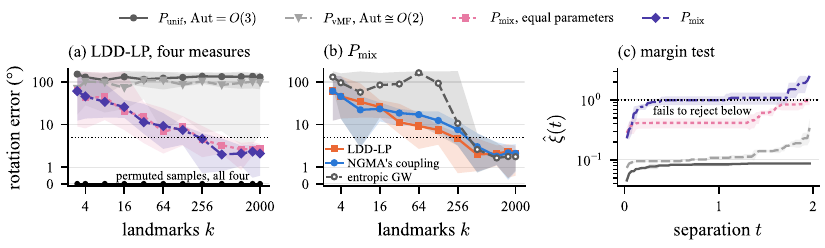}
  \vspace{-7mm}
  \caption{Data on the sphere with $A=R$. \textbf{(a)} Rotation error of the LDD-LP on independent samples against the number of landmarks $k$ (median and range over 10 seeds; at $k=2000$ every point is a landmark), and zero on permuted samples (black). \textbf{(b)} The LDD-LP, NGMA's coupling and entropic GW alone, annealed as in NGMA, on $P_{\mathrm{mix}}$. \textbf{(c)} The margin test: the estimated margin over its DKW accuracy, $\hat\xi(t)$, mean $\pm$ s.d.\@ over 3 seeds, which rejects equal LDDs where $\hat\xi(t)>1$.}
  \label{fig:sphere}
\end{figure}

%% file: tables/main_results.tex
\begin{table}[t]
\centering
\caption{R@1 among the $10{,}000$ test points ($19{,}962$ on CelebA) of VAEs aligned on disjoint samples ($n=3000$ per side, $7500$ on CelebA), as mean $\pm$ s.d.\ over model pairings and draws. The best unpaired baseline is Rosetta, or mini-vec2vec where marked \textsuperscript{\dag}. The best label-informed baseline is RelGeo, or RelReps where marked \textsuperscript{\ddag}. The paired ceilings fit NGMA's map and a rigid map on the true pairs. On CelebA, parentheses count draws reaching R@1 of $0.3$.}
\label{tab:alignment}
\vspace{-1mm}
\small
\setlength{\tabcolsep}{4pt}
\setlength{\aboverulesep}{0pt}
\setlength{\belowrulesep}{0pt}
\begin{tabular}{ll ccc|cc}
\toprule
 & & & \multicolumn{2}{c|}{Best baseline} & \multicolumn{2}{c}{Paired ceiling} \\
\cmidrule(lr){4-5}\cmidrule(lr){6-7}
 & Prior & NGMA & unpaired & label-informed & NGMA & Rigid map \\
\midrule
MNIST & Gaussian & \textbf{0.967}{\scriptsize$\pm$0.021} & 0.941{\scriptsize$\pm$0.028} & 0.525{\scriptsize$\pm$0.069}\textsuperscript{\ddag} & 0.998{\scriptsize$\pm$0.001} & 0.955{\scriptsize$\pm$0.021} \\
 & VampPrior & \textbf{0.989}{\scriptsize$\pm$0.012} & 0.930{\scriptsize$\pm$0.012} & 0.552{\scriptsize$\pm$0.235} & 1.000{\scriptsize$\pm$0.000} & 0.936{\scriptsize$\pm$0.008} \\
 & cross-prior & \textbf{0.955}{\scriptsize$\pm$0.008} & 0.571{\scriptsize$\pm$0.021} & 0.319{\scriptsize$\pm$0.152} & 0.999{\scriptsize$\pm$0.000} & 0.512{\scriptsize$\pm$0.021} \\
\midrule
Fashion- & Gaussian & \textbf{0.719}{\scriptsize$\pm$0.112} & 0.710{\scriptsize$\pm$0.040} & 0.159{\scriptsize$\pm$0.134} & 0.997{\scriptsize$\pm$0.001} & 0.782{\scriptsize$\pm$0.032} \\
MNIST & VampPrior & \textbf{0.979}{\scriptsize$\pm$0.027} & 0.656{\scriptsize$\pm$0.045} & 0.355{\scriptsize$\pm$0.131} & 1.000{\scriptsize$\pm$0.000} & 0.747{\scriptsize$\pm$0.029} \\
 & cross-prior & \textbf{0.871}{\scriptsize$\pm$0.049} & 0.065{\scriptsize$\pm$0.019} & 0.098{\scriptsize$\pm$0.033} & 0.998{\scriptsize$\pm$0.001} & 0.071{\scriptsize$\pm$0.004} \\
\midrule
CelebA & Gaussian & \textbf{0.804}{\scriptsize$\pm$0.375}\,{\scriptsize(12/15)} & 0.567{\scriptsize$\pm$0.453}\,{\scriptsize(8/15)} & 0.445{\scriptsize$\pm$0.033}\textsuperscript{\ddag} & 0.995{\scriptsize$\pm$0.000} & 0.993{\scriptsize$\pm$0.000} \\
 & VampPrior & \textbf{0.818}{\scriptsize$\pm$0.372}\,{\scriptsize(8/10)} & 0.616{\scriptsize$\pm$0.415}\textsuperscript{\dag}\,{\scriptsize(7/10)} & 0.387{\scriptsize$\pm$0.139} & 0.996{\scriptsize$\pm$0.000} & 0.995{\scriptsize$\pm$0.000} \\
 & cross-prior & \textbf{0.818}{\scriptsize$\pm$0.372}\,{\scriptsize(8/10)} & 0.757{\scriptsize$\pm$0.372}\,{\scriptsize(8/10)} & 0.220{\scriptsize$\pm$0.120} & 0.996{\scriptsize$\pm$0.000} & 0.990{\scriptsize$\pm$0.000} \\
\bottomrule
\end{tabular}
\end{table}

%% file: figures/lung_main.tex
\begin{figure}
  \centering
  \includegraphics[width=\textwidth]{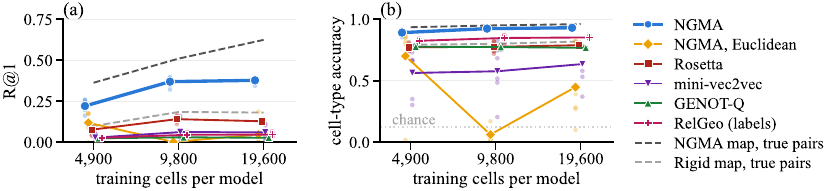}
  \vspace{-5mm}
  \caption{scVI models of lung cells trained on different cells, aligned on disjoint samples of $3000$ cells per side, against the number of training cells per model. \textbf{(a)} R@1 among $10{,}000$ held-out cells (chance $10^{-4}$). \textbf{(b)} Cell-type accuracy of the nearest cell. Lines are means over two model pairings and three draws, faint markers are single draws, and dashed lines are maps fitted on the true pairs.}
  \label{fig:lung_main}
\end{figure}

%% file: sections/related.tex
\section{Related work}
\label{sec:related}
\textbf{Model stitching} and relative representations align two models through pairs, anchors or class labels \citep{csiszarik2021similarity, moschella2023relative, maiorca2023latent, lahner2024direct}, and \citet{yu2024connecting} build them from pullback geodesic distances to anchors.
Without them, unsupervised word translation and its successors for embeddings fit an orthogonal map, which assumes Euclidean or cosine geometry \citep{conneau2018word, artetxe2018robust, grave2019unsupervised, rosetta, minivec2vec}.
vec2vec fits a map with adversarial and cycle losses and a loss that preserves pairwise inner products \citep{jha2025harnessing}.
These methods do not ask whether the recovered map is unique, and the recent ones rest on the Platonic representation hypothesis \citep{platonics}.
We formalise the manifold hypothesis, which gives both models a shared metric, the pullback metric, and makes the alignment map a measure-preserving isometry, identifiable under symmetry breaking (\ref{C1}).

\textbf{Gromov--Wasserstein matching} aligns two samples without correspondence \citep{memoli11, alvarez2018gromov, demetci2022scot, schnaus2025blind}.
\citet{klein2024genot} handle the non-uniqueness of GW couplings empirically, and we state when the optimum is unique (Proposition~\ref{prop:unique-gw-coupling}).
The LDD-LP is Mémoli's third lower bound on GW with the squared $W_2$ cost, whose entropic version \citet{sato2020fast} solve with Sinkhorn.
\citet{hur2025robust} match points with its $W_1$ cost, whose unique optimum is the true matching if no two points share a distance profile, a finite form of \ref{C2}.
\citet{piening2026novel} use the bound in place of GW in fused GW \citep{vayer2019optimal}, and \citet{raviv2012hierarchical} add a distance-histogram cost to GW over bijections.

\textbf{Identifiability theory} asks when the latent variables of one model are determined up to a set of indeterminacies \citep{hyvarinen1999nonlinear, khemakhem2020variational, xi2023}.
\citet{arvanitidis:iclr:2018} measure latent distances with the pullback metric of the decoder.
\citet{syrota2025} show that these distances are identifiable even when the coordinates are not.
We study the map between two models, which needs neither model to be identifiable.
\citet{Ebin1970Manifold} shows that a generic metric has no isometries other than the identity.
We fix the metric and show the same genericity over measures (Proposition~\ref{prop:trivial_automorphism}).

%% file: sections/outro.tex
\section{Discussion}
\label{sec:discuss}

We show that any two models of one data-generating process are related by a single alignment map, an isometry of their pullback geometries that carries one distribution of representations onto the other.
This holds whether a model is built from an encoder, a decoder or both, and whatever its architecture, loss, prior or training data.
The map is identifiable exactly when the data break all symmetries, which generic data do, and two samples of representations recover it exactly if both models encode the same observations and consistently if they do not.
NGMA uses this to align image and single-cell models without pairs, ahead of every baseline without true pairs, across priors and disjoint training data.

\textbf{Limitations.} The theory covers any models of one data-generating process, but our recovery guarantees hold only for global optima, which NGMA's solver may not reach.
NGMA's maps act on whole latent spaces of equal dimension, as in all our experiments.
The latent dimension is a modelling choice, since that of the data manifold is unknown, and two models can choose differently.

%% file: sections/appendix.tex
\input{sections/appendices/appendix_background}

\input{sections/appendices/appendix_proofs}

\input{sections/appendices/appendix_mixture_rigid} 
\input{sections/appendices/appendix_gap}
\input{sections/appendices/appendix_ngma}

\input{sections/appendices/appendix_experiments}

%% file: sections/appendices/appendix_background.tex
\section{Background}
\label{app:background}
This appendix collects the mathematical background required to follow the proofs in Appendix~\ref{app:proofs}.

\subsection{Riemannian manifolds}
Throughout, manifolds are compact, connected and of finite dimension $d$, and may have a boundary, as in \ref{H1}.
Since $\mathcal{M}$ is smooth, the boundary is that of a smooth manifold with boundary.
The proofs use this regularity in two places, for the inward-pointing vector field in the proof of Lemma~\ref{lem:distance-isometries} and for $\operatorname{dVol}_g(\partial\mathcal{M}) = 0$.
We write $\operatorname{int}\mathcal{M}$ for the interior, which is a connected Riemannian manifold without boundary that need not be complete.
Statements below about manifolds without boundary are applied to $\operatorname{int}\mathcal{M}$.
Proofs and a broader exposition are in \citet{docarmo1992riemannian, petersen2016riemannian, lee2018introduction}.
\begin{definition}[\citealp{docarmo1992riemannian, lee2018introduction}]
A \emph{Riemannian manifold} $(\mathcal{M}, g)$ is a smooth manifold $\mathcal{M}$ equipped with a smoothly varying family of inner products $g_x$ on each tangent space $T_x\mathcal{M}$.
The metric $g$ induces:
\begin{itemize}
    \item the \emph{geodesic distance} $d_g(x, y) = \inf_\gamma \int_0^1
\sqrt{g_{\gamma(t)}(\dot\gamma(t), \dot\gamma(t))}\,dt$, where the infimum is over piecewise smooth curves joining $x$ to $y$;
    \item the \emph{Riemannian volume measure} $\operatorname{dVol}_g$, the unique
measure that in any chart equals $\sqrt{\det(g_{ij})}\,dx^1\cdots dx^d$;
\end{itemize}
\end{definition}

\begin{definition}[\citealp{lee2018introduction}]
For each $x \in \mathcal{M}$ the \emph{exponential map} $\exp_x: T_x\mathcal{M} \to \mathcal{M}$ sends a tangent vector $v$ to the endpoint $\gamma(1)$ of the geodesic $\gamma$ with $\gamma(0) = x$ and $\dot\gamma(0) = v$.
\end{definition}
\begin{remark}[\citealp{docarmo1992riemannian}]
On a compact manifold without boundary $\exp_x$ is defined on all of $T_x\mathcal{M}$.
With a boundary it is defined near the origin of $T_x\mathcal{M}$ for interior $x$, which is all that Corollary~\ref{cor:pinning} uses.
\end{remark}

\begin{definition}[\citealp{kobayashi1963foundations, petersen2016riemannian}]
A \emph{Riemannian isometry} $T: (\mathcal{M}, g) \to (\mathcal{N}, h)$ is a diffeomorphism satisfying the pullback condition $T^* h = g$, i.e.\ $g_x(u, v) = h_{T(x)}(dT_x\, u,\, dT_x\, v)$ for all $x \in \mathcal{M}$ and $u, v \in T_x\mathcal{M}$.
Isometries preserve all metric quantities:
\begin{itemize}
    \item geodesic distances: $d_h(T(x), T(y)) = d_g(x, y)$;
    \item metric balls: $T\big(\bar B_{\mathcal{M}}(x, r)\big) = \bar B_{\mathcal{N}}(T(x), r)$ for every $x \in \mathcal{M}$ and $r \geq 0$, where $\bar B_{\mathcal{M}}(x, r)$ is the closed ball of radius $r$ around $x$;
    \item the differential at each point is an orthogonal linear map:
$dT_x \in O(T_x\mathcal{M}, g_x)$.
\end{itemize}
\end{definition}
\begin{theorem}[Myers--Steenrod; \citealp{myers1939group, palais1957differentiability}]
\label{thm:myers-steenrod}
A distance-preserving bijection of a connected Riemannian manifold without boundary, complete or not, is a smooth Riemannian isometry.
The isometry group $\operatorname{Iso}(\mathcal{M}, g)$ of such a manifold is a Lie group with respect to the compact-open topology, acting smoothly and effectively on $\mathcal{M}$.
If $\mathcal{M}$ is compact, then $\operatorname{Iso}(\mathcal{M}, g)$ is compact, and in particular the component group $\pi_0(\operatorname{Iso}(\mathcal{M},g))$ is finite.
\end{theorem}
\begin{proposition}[1-jet rigidity; \citealp{kobayashi1963foundations, myers1939group}]
A self-isometry of a connected Riemannian manifold, complete or not, is uniquely determined by its value and differential at a single point.
Concretely, if $\varphi: \mathcal{M} \to \mathcal{M}$ is an isometry with $\varphi(x_0) = x_0$ and $d_{x_0}\varphi = \operatorname{id}$, then $\varphi = \operatorname{id}_\mathcal{M}$.
\end{proposition}
\begin{corollary}[Pinning an isometry by $d+1$ points]
\label{cor:pinning}
Let $\varphi$ be an isometry of $\operatorname{int}\mathcal{M}$, let $x_1 \in \operatorname{int}\mathcal{M}$, and let $B = \exp_{x_1}(\{v: |v|_g < \varepsilon\}) \subset \operatorname{int}\mathcal{M}$ be a normal ball, so that $\exp_{x_1}$ is a diffeomorphism onto $B$ and each $y \in B$ is joined to $x_1$ by a unique geodesic of length less than $\varepsilon$.
If $\varphi$ fixes $x_1$ and points $x_2, \dots, x_{d+1} \in B$ whose log-map images $v_j = \exp_{x_1}^{-1}(x_j)$ are linearly independent, then $\varphi = \operatorname{id}$, and so is its extension to $\mathcal{M}$.
\end{corollary}
\begin{proof}
$\varphi$ fixes $x_1$ and maps the geodesic $t \mapsto \exp_{x_1}(t v_j)$, $t \in [0,1]$, to a geodesic from $x_1$ to $x_j$ of the same length, which is less than $\varepsilon$.
By uniqueness it is the same geodesic, so $d_{x_1}\varphi\, v_j = v_j$ for $j = 2, \dots, d{+}1$.
An orthogonal map fixing a basis of $T_{x_1}\mathcal{M}$ is the identity, so $d_{x_1}\varphi = \operatorname{id}$ and 1-jet rigidity on the connected manifold $\operatorname{int}\mathcal{M}$ gives $\varphi = \operatorname{id}$.
The extension to $\mathcal{M}$ is the identity by continuity, since $\operatorname{int}\mathcal{M}$ is dense.
\end{proof}
\begin{remark}
If $x_2, \dots, x_{d+1}$ are drawn independently from an absolutely continuous measure restricted to $B$, the log-map images are linearly independent almost surely.
Conditionally on $x_2, \dots, x_j$, the bad set for $x_{j+1}$ is the image under $\exp_{x_1}$ of a proper linear subspace, which is $\operatorname{dVol}_g$-null.
The condition is local, since it needs $d+1$ points in general position close together, not $d+1$ arbitrary points.
\end{remark}

\subsection{Metric measure spaces}
\begin{definition}[\citealp{gromov1999metric, memoli11}]
A \emph{metric measure space} (mm-space) $(\mathcal{M}, d_{\mathcal{M}}, \mu)$ is a compact metric space with a Borel probability measure.
Two mm-spaces are \emph{isomorphic} if a bijection $T\colon\mathcal{M}\to\mathcal{N}$ preserves distances and satisfies $T_\#\mu = \nu$; we write $\operatorname{Iso}\big((\mathcal{M}, d_{\mathcal{M}}, \mu), (\mathcal{N}, d_{\mathcal{N}}, \nu)\big)$ for the set of such maps.
The \emph{automorphism group} is
$$\operatorname{Aut}(\mathcal{M}, d_{\mathcal{M}}, \mu) \;=\; \operatorname{Iso}\!\big((\mathcal{M}, d_{\mathcal{M}}, \mu),\, (\mathcal{M}, d_{\mathcal{M}}, \mu)\big),$$
the distance-preserving bijections of $\mathcal{M}$ that preserve $\mu$.
It forms a group under composition.
\end{definition}
\begin{remark}[Notation]
In the appendices $(\mathcal{M}, g, \mu)$ denotes the mm-space $(\mathcal{M}, d_g, \mu)$, where $d_g$ is the geodesic distance of the Riemannian metric $g$ defined above.
The data mm-space $(\mathcal{M}, d_{\mathcal{M}}, P_{\mathcal{M}})$ and the latent mm-spaces $(\mathcal{Z}_a, d_a, P_a)$ of Section~\ref{sec:latentmmspaces} are of this form.
Statements proved for a general $(\mathcal{M}, g, \mu)$ apply to each of them, with \ref{C1} and \ref{C2} read as conditions on that space; the proof of Proposition~\ref{prop:identifiable-appendix} shows that their automorphism groups are conjugate.
\end{remark}
\begin{lemma}[Distance isometries are Riemannian isometries]
\label{lem:distance-isometries}
Let $(\mathcal{M}, g)$ be a compact connected Riemannian manifold, possibly with a boundary, with geodesic distance $d_g$.
Restriction to $\operatorname{int}\mathcal{M}$ is a bijection from the distance-preserving bijections of $(\mathcal{M}, d_g)$ onto the Riemannian isometries of $(\operatorname{int}\mathcal{M}, g)$, and this group is a compact Lie group acting smoothly and effectively on $\operatorname{int}\mathcal{M}$.
If $\mu = \rho\operatorname{dVol}_g$ with $\rho$ continuous, a distance-preserving bijection $\varphi$ satisfies $\varphi_\#\mu = \mu$ if and only if $\rho\circ\varphi = \rho$.
\end{lemma}
\begin{proof}
Let $d_\circ$ be the length metric of $\operatorname{int}\mathcal{M}$.
Clearly $d_g \le d_\circ$ on $\operatorname{int}\mathcal{M}$.
For the reverse inequality, let $X$ be a smooth vector field on $\mathcal{M}$ that points strictly inward along $\partial\mathcal{M}$, with flow $\theta_s$.
For small $s > 0$, $\theta_s$ maps $\mathcal{M}$ into $\operatorname{int}\mathcal{M}$, and $\theta_s \to \operatorname{id}$ in $C^1$ as $s \to 0$.
For a curve $\gamma$ in $\mathcal{M}$ between interior points $x$ and $y$, the curve that follows the flow line from $x$ to $\theta_s(x)$, then $\theta_s\circ\gamma$, then the flow line from $\theta_s(y)$ back to $y$, stays in $\operatorname{int}\mathcal{M}$ and has length at most $(1 + o(1))\,L(\gamma) + O(s)$.
Hence $d_\circ = d_g$ on $\operatorname{int}\mathcal{M}$.

A distance-preserving bijection $\varphi$ of $\mathcal{M}$ is a homeomorphism, so by invariance of domain it maps $\operatorname{int}\mathcal{M}$ onto itself.
Its restriction preserves $d_\circ$, and is therefore a smooth Riemannian isometry by the Myers--Steenrod theorem (Theorem~\ref{thm:myers-steenrod}), which does not require completeness.
Conversely, a Riemannian isometry of $\operatorname{int}\mathcal{M}$ preserves lengths of curves, hence $d_\circ = d_g$, and extends uniquely to a distance-preserving bijection of $\mathcal{M}$, since $\operatorname{int}\mathcal{M}$ is dense in the compact space $\mathcal{M}$.
The restriction map is a bijection and a group homomorphism.
The group of distance-preserving bijections of a compact metric space is compact in the uniform topology by the Arzel\`a--Ascoli theorem, and the restriction map is a continuous bijection from it onto $\operatorname{Iso}(\operatorname{int}\mathcal{M}, g)$ with the compact-open topology, hence a homeomorphism.
By Theorem~\ref{thm:myers-steenrod} the latter is a Lie group acting smoothly and effectively.

Finally, $\mu(\partial\mathcal{M}) = 0$, and an isometry preserves $\operatorname{dVol}_g$, so $\varphi_\#\mu = (\rho\circ\varphi^{-1})\operatorname{dVol}_g$.
This equals $\mu$ if and only if $\rho\circ\varphi^{-1} = \rho$ almost everywhere, hence everywhere by continuity, which is equivalent to $\rho\circ\varphi = \rho$.
\end{proof}
\begin{remark}
The following two identities, used repeatedly in the proofs, follow from the chain rule for pushforwards:
\begin{align*}
    (T \circ \phi)_{\#} \mu &= T_{\#}(\phi_{\#} \mu), \\
    (T^{-1})_{\#} \nu &= (T_{\#})^{-1} \nu.
\end{align*}
In particular, if $\phi \in \operatorname{Aut}(\mathcal{M},g,\mu)$ and $T \in \operatorname{Iso}\!\big((\mathcal{M},g,\mu),(\mathcal{N},h,\nu)\big)$, then $T \circ \phi \in \operatorname{Iso}\!\big((\mathcal{M},g,\mu),(\mathcal{N},h,\nu)\big)$ because $(T \circ \phi)_{\#} \mu = T_{\#}(\phi_{\#}\mu) = T_{\#}\mu = \nu$.
\end{remark}

For an mm-space $(\mathcal{M}, d_{\mathcal{M}}, \mu)$, a point $x \in \mathcal{M}$ and $r \geq 0$, the \emph{local distance distribution} (LDD) of $x$ is $\Phi_{\mathcal{M}}(x, r):= \mu\big(\bar B_{\mathcal{M}}(x,r)\big)$, as in Section~\ref{sec:ldd_sinkhorn}.
\begin{proposition}[\citealp{memoli11, memoli2022}]
LDDs are preserved by every isomorphism $T: (\mathcal{M},g,\mu) \to (\mathcal{N},h,\nu)$:
$$\Phi_\mathcal{N}(T(x), r) = \Phi_\mathcal{M}(x, r) \qquad \forall\, x \in \mathcal{M},\; r \geq 0.$$
\end{proposition}
\begin{remark}
This follows from two properties of $T$:
\begin{enumerate}
    \item \emph{Isometry maps balls to balls of the same radius:}
$T\big(\bar B_{\mathcal{M}}(x,r)\big) = \bar B_{\mathcal{N}}(T(x),r)$, since $T$ preserves geodesic distances.
    \item \emph{Measure preservation:}
$\nu(T(S)) = \mu(S)$ for every measurable $S \subseteq \mathcal{M}$, since $T_{\#}\mu = \nu$.
\end{enumerate}
Combining the two, $\Phi_\mathcal{N}(T(x),r) = \nu\big(\bar B_{\mathcal{N}}(T(x),r)\big) = \nu\big(T(\bar B_{\mathcal{M}}(x,r))\big) = \mu\big(\bar B_{\mathcal{M}}(x,r)\big) = \Phi_\mathcal{M}(x,r)$.
\end{remark}

\subsection{Analysis and optimal transport}
\begin{proposition}[Coarea formula; \citealp{chavel1984eigenvalues, evans2015measure}]
Let $f: \mathcal{M} \to \mathbb{R}$ be a Lipschitz function and $u: \mathcal{M} \to \mathbb{R}$ integrable.
Then
$$\int_\mathcal{M} u(x)\,|\nabla f(x)|\,\operatorname{dVol}_g \;=\; \int_{-\infty}^{\infty} \int_{f^{-1}(t)} u\,d\mathcal{H}^{d-1}\,dt,$$
where $\mathcal{H}^{d-1}$ is the $(d{-}1)$-dimensional Hausdorff measure.
\end{proposition}
\begin{remark}
\label{rem:null-spheres}
A consequence used in the proofs of Appendix~\ref{app:proofs} is that distance spheres have measure zero.
For any $p \in \mathcal{M}$ and $c > 0$, the distance function $z \mapsto d_g(z, p)$ has $|\nabla d_g(\cdot, p)| = 1$ almost everywhere \citep{docarmo1992riemannian}, so the coarea formula gives $\operatorname{Vol}_g\!\big(\{z: d_g(z,p) = c\}\big) = 0$.
The same holds with a boundary. The function $d_g(\cdot,p)$ is $1$-Lipschitz, and for interior $z \neq p$ a shortest curve from $p$ to $z$ exists by compactness and ends in a geodesic segment of $\operatorname{int}\mathcal{M}$, along which $d_g(\cdot,p)$ grows at unit rate, so $|\nabla d_g(\cdot,p)| = 1$ wherever it is differentiable, which is almost everywhere; the boundary is $\operatorname{Vol}_g$-null.
\end{remark}

Appendix~\ref{app:ldd-multi-scale} uses one more classical fact.
\begin{lemma}[Zero sets of real-analytic functions; \citealp{mityagin2020zero, krantz2002primer}]
\label{lem:analytic-null}
Let $\Omega \subseteq \mathbb{R}^n$ be open and connected and $f: \Omega \to \mathbb{R}$ real-analytic with $f \not\equiv 0$.
Then the zero set $\{z \in \Omega: f(z) = 0\}$ has Lebesgue measure zero.
The same holds, chartwise, on any connected real-analytic manifold.
\end{lemma}

We recall some standard facts from optimal transport \citep{villani2009optimal}.
\begin{definition}
Given probability measures $\mu$ on $\mathcal{M}$ and $\nu$ on $\mathcal{N}$, the set of \emph{couplings} is
$$\Pi(\mu, \nu) = \bigl\{ \eta \in \mathcal{P}(\mathcal{M} \times \mathcal{N}) \;\big|\; (\pi_\mathcal{M})_{\#}\eta = \mu,\; (\pi_\mathcal{N})_{\#}\eta = \nu \bigr\},$$
where $\pi_\mathcal{M}$ and $\pi_\mathcal{N}$ are the coordinate projections.
\end{definition}

\begin{proposition}[Couplings; \citealp{villani2009optimal}]
~
\begin{itemize}
     \item $\Pi(\mu,\nu)$ is tight (its members have the fixed
marginals $\mu$ and $\nu$ on compact spaces), so by Prokhorov's theorem every sequence of couplings has a weakly convergent subsequence.
    \item Transport cost functionals of the form
$\eta \mapsto \int c\,d\eta$ with continuous $c$ are weakly lower semicontinuous, so minimisers exist.
    \item A coupling $\eta$ is the \emph{graph measure} of a measurable map
$T: \mathcal{M} \to \mathcal{N}$ if $\eta = (\operatorname{id}_\mathcal{M} \times T)_{\#}\mu$; equivalently, $\eta$ is concentrated on the graph $\{(x, T(x)): x \in \mathcal{M}\}$.
In this case, $\eta$-almost every pair $(x, y)$ satisfies $y = T(x)$.
\end{itemize}
\end{proposition}

We also recall a fact from measure theory \citep{evans2015measure}.

\begin{proposition}
The support $\operatorname{supp}(\eta) \subseteq \mathcal{M} \times \mathcal{N}$ is the smallest closed set of full $\eta$-measure.
If $\mu$ and $\nu$ have full support (i.e.\ $\operatorname{supp}(\mu) = \mathcal{M}$ and $\operatorname{supp}(\nu) = \mathcal{N}$, which holds when $\mu = \rho\operatorname{dVol}_g$ with density $\rho > 0$ everywhere and likewise for $\nu$, as for $P_a$ and $P_b$), then for any $\eta \in \Pi(\mu,\nu)$:
$$\pi_\mathcal{M}\!\big(\operatorname{supp}(\eta)\big) = \mathcal{M}, \qquad \pi_\mathcal{N}\!\big(\operatorname{supp}(\eta)\big) = \mathcal{N}.$$
\end{proposition}

\begin{remark}
This surjectivity of the projections is used in the proof of Proposition~\ref{prop:unique-gw-coupling} to upgrade a zero-distortion condition on a subset $S \subseteq \mathcal{M} \times \mathcal{N}$ to a correspondence covering all of $\mathcal{M}$ and $\mathcal{N}$.
\end{remark}

The margin test of Section~\ref{sec:ldd_sinkhorn} rests on the following concentration inequality for empirical distribution functions.
\begin{theorem}[Dvoretzky--Kiefer--Wolfowitz inequality, with Massart's constant; \citealp{massart2}]
\label{thm:dkw}
Let $F$ be a distribution function on $\mathbb{R}$ and let $\widehat{F}_N$ be the empirical distribution function of $N$ i.i.d.\ draws from $F$.
For every $\varepsilon > 0$,
$$P\Bigl(\sup_{t \in \mathbb{R}}\, \bigl|\widehat{F}_N(t) - F(t)\bigr| > \varepsilon\Bigr) \;\leq\; 2e^{-2N\varepsilon^2}.$$
\end{theorem}

The genericity result of Appendix~\ref{app:proofs} uses the following topology on smooth functions.

\begin{definition}[Whitney $C^\infty$ topology]
Fix a finite atlas $(U_i, \varphi_i)$ of $\mathcal{M}$ and compact sets $K_i \subset U_i$ that cover $\mathcal{M}$.
The \emph{Whitney $C^\infty$ topology} on $C^\infty(\mathcal{M})$ is defined by the family of seminorms $\|\rho\|_{C^k} = \max_i \sup_{|\alpha| \leq k} \sup_{\varphi_i(K_i)} |\partial^\alpha (\rho \circ \varphi_i^{-1})|$ for $k = 0, 1, 2, \dots$ (any two choices of atlas and compact sets give equivalent seminorms on a compact manifold).
A sequence $\rho_n \to \rho$ iff all partial derivatives converge uniformly.
This makes $C^\infty(\mathcal{M})$ a Fr\'echet space (a complete, metrisable, locally convex topological vector space).
The subset $C^\infty_{>0}(\mathcal{M}) = \{\rho \in C^\infty(\mathcal{M}) \mid \rho > 0\}$ is open in this topology.
\end{definition}

\subsection{Lie groups and Lie algebras}
The following are general results and definitions from \citet{hall2015lie, lee2012smooth}.
\begin{definition}[Lie group]
A \emph{Lie group} is a smooth manifold $G$ that is also a group, such that the multiplication $(g, h) \mapsto gh$ and inversion $g \mapsto g^{-1}$ are both smooth maps.
\end{definition}

\begin{example}
The group $O(n) = \{Q \in \mathbb{R}^{n \times n} \mid Q^\top Q = I\}$ of orthogonal matrices is a compact Lie group of dimension $\tfrac{n(n-1)}{2}$.
The subgroup $SO(n) = \{Q \in O(n) \mid \det Q = 1\}$ of rotations is also a compact Lie group.
\end{example}

\begin{definition}[Lie algebra of a Lie group]
The \emph{Lie algebra} $\mathfrak{g}$ of a Lie group $G$ is the vector space $T_e G$ (the tangent space at the identity) equipped with a bracket $[\cdot, \cdot]: \mathfrak{g} \times \mathfrak{g} \to \mathfrak{g}$ that is bilinear, antisymmetric ($[X,Y] = -[Y,X]$), and satisfies the Jacobi identity.
Concretely, elements of $\mathfrak{g}$ are identified with left-invariant vector fields on $G$, and the bracket is the usual Lie bracket of vector fields.

The dimension of $\mathfrak{g}$ as a vector space equals the dimension of $G$ as a manifold.
\end{definition}

\begin{example}
The Lie algebra $\mathfrak{o}(3)$ of $O(3)$ is the $3$-dimensional space of skew-symmetric matrices $\{X \in \mathbb{R}^{3 \times 3} \mid X^\top = -X\}$, with bracket $[X,Y] = XY - YX$.
A standard basis is:
    $$
    X_1 = \begin{pmatrix} 0 & 0 & 0 \\ 0 & 0 & -1 \\ 0 & 1 & 0 \end{pmatrix}, \quad
    X_2 = \begin{pmatrix} 0 & 0 & 1 \\ 0 & 0 & 0 \\ -1 & 0 & 0 \end{pmatrix}, \quad
    X_3 = \begin{pmatrix} 0 & -1 & 0 \\ 1 & 0 & 0 \\ 0 & 0 & 0 \end{pmatrix},
    $$
generating rotations about the $x$-, $y$-, and $z$-axes respectively.
\end{example}

\begin{definition}[One-parameter subgroup]
A \emph{one-parameter subgroup} of a Lie group $G$ is a smooth homomorphism $\gamma: (\mathbb{R}, +) \to G$, i.e., a smooth curve satisfying $\gamma(s+t) = \gamma(s)\gamma(t)$ for all $s, t \in \mathbb{R}$ and $\gamma(0) = e$.
\end{definition}

\begin{definition}[Exponential map]
The \emph{exponential map} $\exp: \mathfrak{g} \to G$ sends each $X \in \mathfrak{g}$ to $\gamma_X(1)$, where $\gamma_X$ is the unique one-parameter subgroup with $\dot{\gamma}_X(0) = X$.
The full one-parameter subgroup is then $t \mapsto \exp(tX)$.
\end{definition}

\begin{example}
In $O(3)$, the one-parameter subgroup generated by $X_3$ is the family of rotation matrices:
    $$\exp(t X_3) = \begin{pmatrix} \cos t & -\sin t & 0 \\ \sin t & \cos t & 0 \\ 0 & 0 & 1 \end{pmatrix},$$
which rotates about the $z$-axis by angle $t$.
\end{example}

\begin{definition}[Smooth group action]
A \emph{smooth (left) action} of a Lie group $G$ on a smooth manifold $\mathcal{M}$ is a smooth map $\alpha: G \times \mathcal{M} \to \mathcal{M}$, written $\alpha(g, p) = g \cdot p$, satisfying $e \cdot p = p$ and $(gh) \cdot p = g \cdot (h \cdot p)$ for all $g, h \in G$ and $p \in \mathcal{M}$.
\end{definition}

\begin{definition}[Infinitesimal generator]
Given a smooth action of $G$ on $\mathcal{M}$ and an element $X \in \mathfrak{g}$, the \emph{infinitesimal generator} $X^\sharp$ is the vector field on $\mathcal{M}$ defined by
    $$X^\sharp_p = \left.\frac{d}{dt}\right|_{t=0} \exp(tX) \cdot p.$$
It is the velocity field on $\mathcal{M}$ induced by the flow $t \mapsto \exp(tX) \cdot p$.
\end{definition}

\begin{definition}[Effective action]
An action is \emph{effective} (or \emph{faithful}) if the only group element that fixes every point is the identity: $g \cdot p = p$ for all $p \in \mathcal{M}$ implies $g = e$.
\end{definition}

\begin{example}
The standard action of $O(3)$ on $S^2$ is effective.
The infinitesimal generator of $X_3$ (rotation about the $z$-axis) is the vector field $X_3^\sharp = \partial_\phi$ in spherical coordinates $(\theta, \phi)$. It points eastward at every point, with magnitude $\sin\theta$.
\end{example}

\begin{definition}[Lie derivative]
Let $X$ be a vector field on $\mathcal{M}$ with flow $\varphi_t$.
The \emph{Lie derivative} of a smooth function $\rho: \mathcal{M} \to \mathbb{R}$ along $X$ is:
    $$L_X \rho (p) = \left.\frac{d}{dt}\right|_{t=0} \rho(\varphi_t(p)) = d\rho_p(X_p).$$
In local coordinates, $L_X \rho = X^i \partial_i \rho$, the directional derivative of $\rho$ along $X$.
\end{definition}

The following fact converts a global symmetry condition into a differential equation.

\begin{lemma}[Invariance via Lie derivative]
Let $G$ act smoothly on $\mathcal{M}$, let $X \in \mathfrak{g}$, and let $\rho \in C^\infty(\mathcal{M})$.
Then $\rho$ is invariant under the one-parameter subgroup $\{\exp(tX)\}_{t \in \mathbb{R}}$, meaning $\rho(\exp(tX) \cdot p) = \rho(p)$ for all $t$ and all $p$, if and only if $L_{X^\sharp} \rho = 0$ everywhere on $\mathcal{M}$.
    \label{lem:invariance-lie-deriv}
\end{lemma}

\begin{proof}
Define $h(t) = \rho(\exp(tX) \cdot p)$.
Then $h'(t) = (L_{X^\sharp}\rho)(\exp(tX) \cdot p)$.
If $L_{X^\sharp}\rho = 0$, then $h' \equiv 0$, so $h(t) = h(0) = \rho(p)$.
Conversely, if $h(t) = h(0)$ for all $t$ and $p$, then $h'(0) = L_{X^\sharp}\rho(p) = 0$.
\end{proof}

\begin{example}
On $S^2$, the condition $L_{X_3^\sharp}\rho = \partial_\phi \rho = 0$ says that $\rho$ does not depend on longitude, that is, $\rho$ is rotationally symmetric about the $z$-axis.
Imposing all three conditions $L_{X_1^\sharp}\rho = L_{X_2^\sharp}\rho = L_{X_3^\sharp}\rho = 0$ forces $\rho$ to be constant on $S^2$.
\end{example}

\begin{definition}[Identity component]
The \emph{identity component} $G_0$ of a Lie group $G$ is the connected component containing the identity element $e$.
It is a closed normal subgroup of $G$ with $\dim G_0 = \dim G$ and the same Lie algebra $\mathfrak{g}$.
\end{definition}

\begin{definition}[Component group]
The quotient group $\pi_0(G):= G/G_0$ is called the \emph{component group}.
Its elements are the connected components of $G$, and it is a discrete group.
\end{definition}

\begin{example}
For $O(3)$, $G_0 = SO(3)$ (proper rotations), and $\pi_0(O(3)) \cong \mathbb{Z}/2\mathbb{Z}$ with the non-trivial element represented by any orientation-reversing isometry (e.g., the antipodal map $p \mapsto -p$).
\end{example}

%% file: sections/appendices/appendix_proofs.tex
\section{Proofs}
\label{app:proofs}

Throughout the appendix, $d:= \dim\mathcal{M}$.

\subsection{Proofs for Section~\ref{sec:latentmmspaces}}

\begin{remark}[The projection onto $\mathcal{M}$]
\label{rem:projection}
Here $\pi$ is the nearest-point map in the Euclidean distance of $\mathcal{X}$.
The distance to the closed set $\mathcal{M}$ is $1$-Lipschitz \citep[4.8(1)]{federer1959curvature}, hence differentiable almost everywhere by Rademacher's theorem, and every point where it is differentiable has a unique nearest point on $\mathcal{M}$ \citep[4.8(3)]{federer1959curvature}, so $\pi$ is defined Lebesgue-almost everywhere and is continuous on its domain \citep[4.8(4)]{federer1959curvature}.
Hence $\pi_{\#}P_{\mathcal{X}}$ is well defined whenever $P_{\mathcal{X}}$ has a density.
The noise condition of \ref{H1}, normal to $\mathcal{M}$ and shorter than the reach \citep{federer1959curvature}, is what makes $\pi(x)$ the point that $x$ was generated from, so that $\pi_{\#}P_{\mathcal{X}}$ is the distribution of the generating points and not a blurred version of it; within the reach $\pi$ is moreover Lipschitz \citep[4.8(8)]{federer1959curvature}.
The proofs use only that $P_{\mathcal{M}}$ has a smooth positive density on $\mathcal{M}$.
That assumption also rules out observations beyond the boundary of $\mathcal{M}$, which would project onto $\partial\mathcal{M}$, a set of volume zero; normal noise never produces them.
\end{remark}

\begin{lemma}[Sufficient conditions for \ref{H2}]
\label{lem:decoder-chart}
Let $\mathcal{M}\subseteq\mathcal{X}$ be a compact $d$-dimensional embedded submanifold, possibly with smooth boundary.
\emph{(i)} Let $\mathcal{W}$ be a manifold of dimension $q\geq d$, such as $\mathbb{R}^q$, and $\operatorname{dec}\colon\mathcal{W}\to\mathcal{X}$ smooth and injective with a Jacobian of rank $q$, with $\mathcal{M}\subseteq\operatorname{dec}(\mathcal{W})$ and $\mathcal{Z}:= \operatorname{dec}^{-1}(\mathcal{M})$ compact.
Then $\mathcal{Z}$ is a compact embedded submanifold of $\mathcal{W}$ of dimension $d$ and $\operatorname{dec}|_{\mathcal{Z}}\colon\mathcal{Z}\to\mathcal{M}$ is a diffeomorphism, so $\operatorname{dec}|_{\mathcal{Z}}$ satisfies \ref{H2}.
The preimage is closed, so it is compact whenever $\mathcal{W}$ is, and then $\operatorname{dec}$ is a smooth embedding \citep[Lemma~4.1]{syrota2025}; it is also compact whenever $\operatorname{dec}$ is a smooth embedding.
\emph{(ii)} Let $\operatorname{enc}\colon\mathcal{X}\to\mathbb{R}^q$ be smooth, injective on $\mathcal{M}$, with a Jacobian of rank $d$ on the tangent spaces of $\mathcal{M}$.
Then $\operatorname{enc}(\mathcal{M})$ is a compact embedded submanifold of $\mathbb{R}^q$ and $(\operatorname{enc}|_{\mathcal{M}})^{-1}\colon\operatorname{enc}(\mathcal{M})\to\mathcal{M}$ satisfies \ref{H2}.
\end{lemma}

\begin{proof}
\emph{(i)} Suppose first that $\operatorname{dec}$ is an embedding, so $\mathcal{N}:= \operatorname{dec}(\mathcal{W})$ is an embedded $q$-dimensional submanifold of $\mathcal{X}$ and $\operatorname{dec}\colon\mathcal{W}\to\mathcal{N}$ is a diffeomorphism.
The inclusion $\mathcal{M}\hookrightarrow\mathcal{X}$ has image in $\mathcal{N}$, so it is smooth as a map into $\mathcal{N}$ \citep{lee2018introduction}, and it is an injective immersion with compact domain, hence an embedding.
Composing with $\operatorname{dec}^{-1}\colon\mathcal{N}\to\mathcal{W}$ shows that $\operatorname{dec}^{-1}|_{\mathcal{M}}$ embeds $\mathcal{M}$ into $\mathcal{W}$ with image $\mathcal{Z}$, and $\operatorname{dec}|_{\mathcal{Z}}$ is its inverse.

In general, $\mathcal{Z}$ is compact, so it has an open neighbourhood $U$ in $\mathcal{W}$ whose closure $\bar U$ is compact.
On $\bar U$ the map $\operatorname{dec}$ is continuous and injective, hence a homeomorphism onto its image, and so is its restriction to $U$.
Thus $\operatorname{dec}|_U$ is an injective immersion and a homeomorphism onto its image, that is, a smooth embedding, with $\mathcal{M}\subseteq\operatorname{dec}(U)$ and $U\cap\operatorname{dec}^{-1}(\mathcal{M}) = \mathcal{Z}$.
The first paragraph applied to $\operatorname{dec}|_U$ concludes.

\emph{(ii)} The restriction $\operatorname{enc}|_{\mathcal{M}}$ is an injective immersion of a compact manifold, hence an embedding \citep{lee2018introduction}.
So $\operatorname{enc}(\mathcal{M})$ is a compact embedded submanifold of $\mathbb{R}^q$ of dimension $d$ and $\operatorname{enc}|_{\mathcal{M}}\colon\mathcal{M}\to\operatorname{enc}(\mathcal{M})$ is a diffeomorphism, whose inverse is a diffeomorphism onto $\mathcal{M}$.
\end{proof}

\begin{remark}[An injective immersion is not enough]
An injective immersion need not have a continuous inverse on its image.
The map $f\colon\mathbb{R}\to\mathbb{R}^2$ with $f(t) = (\cos\vartheta(t), \sin\vartheta(t))$, $\vartheta(t) = 2\pi t/(1+t)$ for $t \geq 0$, extended to $t < 0$ by a curve leaving the circle, is an injective immersion whose image contains the unit circle $\mathcal{M}$.
Yet $f^{-1}(\mathcal{M}) = [0,\infty)$, and points of $\mathcal{M}$ on either side of $(1,0)$ have preimages near $0$ and near $\infty$.
The pullback length metric on $f^{-1}(\mathcal{M})$ is then not the length metric of $\mathcal{M}$.
An injective decoder with full-rank Jacobian whose image contains $\mathcal{M}$ therefore need not satisfy \ref{H2} unless $f^{-1}(\mathcal{M})$ is compact, and Proposition~\ref{prop:alignment} can fail for it.
\end{remark}

\begin{proposition}[Proposition~\ref{prop:shared-metric}]
\label{prop:shared-metric-appendix}
Under \ref{H1}--\ref{H2}, metrics $d_a$ on $\mathcal{Z}_a$ and $d_b$ on $\mathcal{Z}_b$ satisfy $d_b(A(z),A(z')) = d_a(z,z')$ for all $z,z'\in\mathcal{Z}_a$ if and only if $d_a(f_a^{-1}(x), f_a^{-1}(y)) = d_b(f_b^{-1}(x), f_b^{-1}(y))$ for all $x,y\in\mathcal{M}$.
Euclidean metrics on two subsets of one $\mathbb{R}^q$ satisfy this if and only if $A$ extends to a rigid motion of $\mathbb{R}^q$, and geodesic metrics on two subsets of the unit sphere $S^{q-1}\subset\mathbb{R}^q$ if and only if $A$ extends to an orthogonal map of $\mathbb{R}^q$.
\end{proposition}

\begin{proof}
Since $f_a$ is a bijection onto $\mathcal{M}$, every pair $z,z'\in\mathcal{Z}_a$ is the pair of representations $z = f_a^{-1}(x)$, $z' = f_a^{-1}(y)$ of exactly one pair $x,y\in\mathcal{M}$, and Definition~\ref{def:alignment} gives $A(z) = f_b^{-1}(x)$ and $A(z') = f_b^{-1}(y)$.
Substituting turns one condition into the other.

For the Euclidean case, a rigid motion preserves Euclidean distances, so its restriction does.
Conversely, let $A$ preserve them, fix $z_0\in\mathcal{Z}_a$, and set $u:= z - z_0$ and $B(u):= A(z) - A(z_0)$.
Then $\|B(u)\| = \|u\|$ and $\|B(u) - B(u')\| = \|u - u'\|$, so the polarisation identity gives $\langle B(u), B(u')\rangle = \langle u, u'\rangle$ for all $u,u'$.
A map that preserves inner products on a set extends to a linear isometry of its span, which extends to an orthogonal map $Q$ of $\mathbb{R}^q$.
Hence $A(z) = Q(z - z_0) + A(z_0)$, a rigid motion.

For the spherical case, an orthogonal map preserves inner products, hence the geodesic distance $\arccos\langle z, z'\rangle$ on $S^{q-1}$.
Conversely, if $A$ preserves geodesic distances on the sphere, then $\langle A(z), A(z')\rangle = \cos d(A(z),A(z')) = \cos d(z,z') = \langle z, z'\rangle$, so $A$ preserves inner products and extends to an orthogonal map $Q$ of $\mathbb{R}^q$ as above.
The cosine distance is a monotone function of the geodesic distance, so the same holds for it.
\end{proof}

\begin{proposition}[Proposition~\ref{prop:alignment}]
\label{prop:alignment-appendix}
Under \ref{H1}--\ref{H2} the alignment map exists, is unique, and equals $A = f_b^{-1}\circ f_a$.
It satisfies $A^*g_b = g_a$ and $A_\#P_a = P_b$, so it is an isomorphism of $(\mathcal{Z}_a, d_a, P_a)$ and $(\mathcal{Z}_b, d_b, P_b)$.
\end{proposition}

\begin{proof}
By \ref{H2}, $f_a|_{\mathcal{Z}_a}$ and $f_b|_{\mathcal{Z}_b}$ are diffeomorphisms onto $\mathcal{M}$.
A map $A$ with $f_b\circ A = f_a$ must send $z$ to the unique preimage of $f_a(z)$ under $f_b$, so $A = f_b^{-1}\circ f_a$ is the only alignment map, and it is a diffeomorphism.
The chain rule gives
$$A^*g_b = A^*f_b^*g^{\mathcal{M}} = (f_b\circ A)^*g^{\mathcal{M}} = f_a^*g^{\mathcal{M}} = g_a,$$
as in \citet[Theorem~4.5]{syrota2025}.
Hence $A$ maps each piecewise smooth curve in $\mathcal{Z}_a$ to one in $\mathcal{Z}_b$ of equal length, and $A^{-1}$ does the same in reverse, so $d_b(A(z), A(z')) = d_a(z, z')$ for all $z, z'\in\mathcal{Z}_a$.
Finally, $A_\#P_a = (f_b^{-1}\circ f_a\circ f_a^{-1})_\#P_{\mathcal{M}} = (f_b^{-1})_\#P_{\mathcal{M}} = P_b$.
The same argument with $f_a$ in place of $A$ shows that $f_a$ is an isomorphism from $(\mathcal{Z}_a, d_a, P_a)$ onto $(\mathcal{M}, d_{\mathcal{M}}, P_{\mathcal{M}})$.
\end{proof}

\subsection{Proofs for Section~\ref{sec:identifiability}}

\begin{lemma}[Isomorphisms differ by automorphisms]
\label{lem:iso-aut}
Let $(\mathcal{M}, g, \mu)$ and $(\mathcal{N}, h, \nu)$ be mm-spaces and $T \in \operatorname{Iso}\big((\mathcal{M},g,\mu),(\mathcal{N},h,\nu)\big)$.
Then
$$\operatorname{Iso}\big((\mathcal{M},g,\mu),(\mathcal{N},h,\nu)\big) = \{T\circ\phi \;:\; \phi\in\operatorname{Aut}(\mathcal{M},g,\mu)\}.$$
In particular $T$ is the only isomorphism if and only if $\operatorname{Aut}(\mathcal{M},g,\mu) = \{\operatorname{id}_{\mathcal{M}}\}$, and then also $\operatorname{Aut}(\mathcal{N},h,\nu) = \{\operatorname{id}_{\mathcal{N}}\}$, since $\psi\mapsto T^{-1}\circ\psi\circ T$ is a group isomorphism between the two automorphism groups.
\end{lemma}

\begin{proof}
For $\phi\in\operatorname{Aut}(\mathcal{M},g,\mu)$, the composition $T\circ\phi$ preserves distances and $(T\circ\phi)_\#\mu = T_\#(\phi_\#\mu) = T_\#\mu = \nu$, so it is an isomorphism.
Conversely, for an isomorphism $T'$, the map $\phi:= T^{-1}\circ T'$ is a distance-preserving bijection of $\mathcal{M}$ with $\phi_\#\mu = (T^{-1})_\#\nu = \mu$, so $\phi\in\operatorname{Aut}(\mathcal{M},g,\mu)$ and $T' = T\circ\phi$.
The map $\phi\mapsto T\circ\phi$ is injective, which gives the second statement.
\end{proof}

\begin{proposition}[Proposition~\ref{prop:identifiable}]
\label{prop:identifiable-appendix}
Under \ref{H1}--\ref{H2}, the isomorphisms from $(\mathcal{Z}_a, d_a, P_a)$ to $(\mathcal{Z}_b, d_b, P_b)$ are exactly the maps $f_b^{-1}\circ S\circ f_a$ with $S\in\operatorname{Aut}(\mathcal{M}, d_{\mathcal{M}}, P_{\mathcal{M}})$.
Hence every such isomorphism equals the alignment $A$ if and only if \ref{C1} holds.
\end{proposition}

\begin{proof}
By Proposition~\ref{prop:alignment-appendix}, $A$ is an isomorphism of the latent mm-spaces and $f_a$ is an isomorphism of $(\mathcal{Z}_a, d_a, P_a)$ onto the data mm-space.
Conjugation by $f_a$, $S\mapsto f_a^{-1}\circ S\circ f_a$, is therefore a group isomorphism from $\operatorname{Aut}(\mathcal{M}, d_{\mathcal{M}}, P_{\mathcal{M}})$ onto $\operatorname{Aut}(\mathcal{Z}_a, d_a, P_a)$.
Lemma~\ref{lem:iso-aut} applied to $T = A$ shows that the isomorphisms are the maps $A\circ f_a^{-1}\circ S\circ f_a$, and $A\circ f_a^{-1} = f_b^{-1}$ by Definition~\ref{def:alignment}, which gives the first statement.
The second follows because the latent automorphism group is trivial exactly when the data automorphism group is.
\end{proof}

\begin{proposition}[Density generically trivialises the automorphism group; Proposition~\ref{prop:trivial_automorphism}]
Let $(\mathcal{M}, g)$ be a compact connected Riemannian manifold, possibly with smooth boundary, and let $G$ be its group of distance-preserving bijections, identified with $\operatorname{Iso}(\operatorname{int}\mathcal{M}, g)$ by Lemma~\ref{lem:distance-isometries}.
The set of smooth positive densities for which the automorphism group is trivial,
    $$\mathcal{D}_{\mathrm{rigid}}:= \left\{ \rho \in C^\infty_{>0}(\mathcal{M}) \;\middle|\; \operatorname{Aut}(\mathcal{M}, g, \rho\, \operatorname{dVol}_g) = \{\operatorname{id}\} \right\},$$
is open and dense in $C^\infty_{>0}(\mathcal{M})$ equipped with the Whitney $C^\infty$ topology.
    \label{prop:trivial_automorphism-appendix}
\end{proposition}

Proposition~\ref{prop:trivial_automorphism} is the case $g = g^{\mathcal{M}}$, with $\rho$ the density of $P_{\mathcal{M}}$.
By Lemma~\ref{lem:distance-isometries}, a density $\rho$ lies in $\mathcal{D}_{\mathrm{rigid}}$ precisely when no non-trivial isometry $\varphi \in G \setminus \{\operatorname{id}\}$ leaves $\rho$ invariant, i.e., there is no $\varphi \neq \operatorname{id}$ with $\rho \circ \varphi = \rho$ everywhere.
The proof shows that the densities with a non-trivial stabiliser form a closed set with empty interior.
Open and dense is the topological notion of genericity.
No canonical measure exists on the infinite-dimensional space $C^\infty_{>0}(\mathcal{M})$, so we do not claim that almost every density satisfies \ref{C1}.
\begin{proof}
All points, neighbourhoods and Lie derivatives below are taken in $\operatorname{int}\mathcal{M}$, on which $G$ acts smoothly and effectively; this is possible because a non-trivial element of $G$ moves some interior point, and invariance $\rho\circ\varphi = \rho$ on the dense set $\operatorname{int}\mathcal{M}$ is invariance on $\mathcal{M}$.
By Lemma~\ref{lem:distance-isometries}, $G$ is a compact Lie group with Lie algebra $\mathfrak{g}$\footnote{Here $\mathfrak{g}$ is the Lie algebra of $G$, not the Riemannian metric $g$.} of dimension $m$ and finite component group $\pi_0(G) = G/G_0$.
Fix an inner product on $\mathfrak{g}$, with norm $\|\cdot\|$.

For each $\rho \in C^\infty_{>0}(\mathcal{M})$, define its \emph{stabiliser} $\operatorname{Stab}(\rho):= \{\varphi \in G \mid \rho \circ \varphi = \rho\}$ as the closed Lie subgroup of $G$ whose action leaves $\rho$ invariant.
Its Lie algebra is
    $$K(\rho):= \operatorname{Lie}(\operatorname{Stab}(\rho)) = \{Y \in \mathfrak{g} \mid L_{Y^\sharp}\rho = 0\},$$
by Lemma~\ref{lem:invariance-lie-deriv}.
Let $k(\rho):= \dim K(\rho)$, and let $T_\rho: \mathfrak{g} \to C^\infty(\mathcal{M})$ be the linear map $Y \mapsto L_{Y^\sharp}\rho$, so $k(\rho) = \dim \ker T_\rho$.
Note that $\mathcal{D}_{\mathrm{rigid}} = \{\rho \mid \operatorname{Stab}(\rho) = \{\operatorname{id}\}\}$.

\textbf{Openness.}
We show $C^\infty_{>0}(\mathcal{M}) \setminus \mathcal{D}_{\mathrm{rigid}}$ is closed.
Suppose $\rho_n \to \rho$ in $C^\infty$ with each $\operatorname{Stab}(\rho_n) \neq \{\operatorname{id}\}$.
Pick $\varphi_n \in \operatorname{Stab}(\rho_n) \setminus \{\operatorname{id}\}$.
By compactness of $G$, we can pass to a subsequence (not relabelled) with $\varphi_n \to \varphi \in G$.
Then $\rho_n\circ\varphi_n\to\rho\circ\varphi$ uniformly, since $|\rho_n(\varphi_n(p)) - \rho(\varphi(p))|\le\|\rho_n-\rho\|_\infty + |\rho(\varphi_n(p))-\rho(\varphi(p))|$, where the first term tends to zero and the second does so uniformly in $p$ because $\rho$ is uniformly continuous and $\varphi_n\to\varphi$ uniformly.
As $\rho_n\circ\varphi_n = \rho_n\to\rho$, this gives $\rho\circ\varphi = \rho$, so $\varphi\in\operatorname{Stab}(\rho)$.

If $\varphi \neq \operatorname{id}$, we are done.
If $\varphi = \operatorname{id}$, then for large $n$, $\varphi_n$ lies in a neighbourhood of $\operatorname{id}$ on which $\exp$ is a diffeomorphism; write $\varphi_n = \exp(t_n Y_n)$ with $\|Y_n\| = 1$, $t_n \to 0^+$.
Pass to a further subsequence with $Y_n \to Y$, $\|Y\| = 1$.
By the fundamental theorem of calculus applied to $s \mapsto \rho_n(\exp(sY_n)\cdot p)$ for any $p\in \mathcal{M}$,
    $$0 = \rho_n(\exp(t_n Y_n)\cdot p) - \rho_n(p) = \int_0^{t_n} (L_{Y_n^\sharp}\rho_n)(\exp(sY_n)\cdot p)\, ds.$$
Dividing by $t_n$ expresses $0$ as the average of the integrand over $s \in [0, t_n]$.
As $n \to \infty$, $\rho_n \to \rho$ in $C^1$, $Y_n \to Y$, and $\exp(sY_n)\cdot p \to p$ uniformly in $s \in [0, t_n]$; so the average tends to $(L_{Y^\sharp}\rho)(p)$.
Hence $L_{Y^\sharp}\rho \equiv 0$, so $Y \in K(\rho)$ with $\|Y\| = 1$, giving $\exp(sY) \in \operatorname{Stab}(\rho)$ for all $s$, with $\exp(sY) \neq \operatorname{id}$ for small $s \neq 0$ since $\exp$ is a local diffeomorphism at $0$.

\textbf{Density.}
We show every $\rho_0 \in C^\infty_{>0}(\mathcal{M})$ lies in the closure of $\mathcal{D}_{\mathrm{rigid}}$.
The argument has two stages.
The first reduces $k(\rho)$ to $0$, so that $\operatorname{Stab}(\rho)$ becomes finite, and the second removes the remaining finite symmetries.

    \medskip

\emph{Step 1: Reduce $k$ to $0$.}
We prove the following lemma: \emph{if $k(\rho) = \ell \geq 1$, then every $C^\infty$-neighbourhood of $\rho$ contains some $\rho' \in C^\infty_{>0}(\mathcal{M})$ with $k(\rho') \leq \ell-1$.}
Iterating at most $m$ times yields $\rho_1 \in C^\infty_{>0}(\mathcal{M})$ arbitrarily close to $\rho_0$ with $k(\rho_1) = 0$.

Let $V:= C^\infty(\mathcal{M})/\operatorname{Im}(T_\rho)$.
For $\psi \in C^\infty(\mathcal{M})$, define
    $$\overline{T}_\psi: K(\rho) \to V, \qquad Y \mapsto [L_{Y^\sharp}\psi].$$
\emph{Claim 1: for every nonzero $Y \in K(\rho)$ there exists $\psi \in C_c^\infty(\mathcal{M})$ with $\overline{T}_\psi(Y) \neq 0$.}

Since the action is effective and $Y \neq 0$, there is $p \in \mathcal{M}$ with $Y^\sharp|_p \neq 0$.
Choose local coordinates $(x^1, \dots, x^d)$ on a neighbourhood $U$ of $p$ in which $Y^\sharp = \partial_{x^1}$ (straightening theorem).
Consider the linear map
$$
\Lambda: C_c^\infty(U) \to C^\infty(\mathcal{M}),
\qquad \psi \mapsto L_{Y^\sharp}\psi = \partial_{x^1}\psi,
$$
where functions on $U$ are extended by zero to $\mathcal{M}$.
The image $\operatorname{Im}(\Lambda)$ is infinite-dimensional.
For instance, the functions $\partial_{x^1}(\chi \cdot (x^1)^j) = \chi \cdot j(x^1)^{j-1} + (\partial_{x^1}\chi)\cdot(x^1)^j$, for $j = 1, 2, 3, \dots$ and a fixed bump $\chi \in C_c^\infty(U)$ with $\chi \equiv 1$ near $p$, are linearly independent in $C^\infty(\mathcal{M})$ (their Taylor expansions at $p$ have distinct leading orders).

On the other hand, $\operatorname{Im}(T_\rho) \subseteq C^\infty(\mathcal{M})$ is finite-dimensional, of dimension at most $m - \ell$.
An infinite-dimensional subspace cannot be contained in a finite-dimensional subspace, so there exists $\psi \in C_c^\infty(U)$ with $L_{Y^\sharp}\psi \notin \operatorname{Im}(T_\rho)$, i.e., $\overline{T}_\psi(Y) \neq 0$.
This establishes Claim~1.

Fix a nonzero $Y_0 \in K(\rho)$ (possible since $\ell \geq 1$) and let $\psi$ be as in Claim~1 for this $Y_0$.
Fix any norm on the finite-dimensional subspace $\operatorname{Im}(T_\rho) + T_\psi(\mathfrak{g}) \subset C^\infty(\mathcal{M})$, in which all vectors below live.
For $|s|$ small, $\rho + s\psi$ is positive, because $\rho > 0$ on the compact $\mathcal{M}$ and $\psi$ is bounded, and $\rho + s\psi \to \rho$ in $C^\infty$ as $s \to 0$.

\emph{Claim 2: $k(\rho + s\psi) \leq \ell - 1$ for all sufficiently small $s \neq 0$.}

Suppose not.
Then there is a sequence $s_n \to 0$, $s_n \neq 0$, with $k(\rho + s_n\psi) \geq \ell$.
Choose $\ell$-dimensional subspaces $L_n \subseteq \ker T_{\rho + s_n\psi}$; by compactness of the Grassmannian of $\ell$-planes in $\mathfrak{g}$, pass to a subsequence with $L_n \to L$, $\dim L = \ell$.
We show $L \subseteq K(\rho)$ with $\overline{T}_\psi|_{L} \equiv 0$; since $\dim L = \ell = \dim K(\rho)$ this forces $L = K(\rho) \ni Y_0$, contradicting $\overline{T}_\psi(Y_0) \neq 0$.

Fix a unit $W \in L$ and choose unit $W_n \in L_n$ with $W_n \to W$.
Decompose $W_n = W_n^\parallel + Z_n$ with $W_n^\parallel \in K(\rho)$, $Z_n \in K(\rho)^\perp$.
Since $T_\rho|_{K(\rho)^\perp}$ is injective between finite-dimensional spaces, there is $C$ with $\|Z\| \leq C\|T_\rho(Z)\|$ on $K(\rho)^\perp$.
From $0 = T_{\rho+s_n\psi}(W_n) = T_\rho(Z_n) + s_n T_\psi(W_n)$ we get $\|Z_n\| \leq C|s_n|\,\|T_\psi(W_n)\| \leq CM|s_n| \to 0$, where $M:= \sup_{\|V\|=1}\|T_\psi(V)\|$; hence $W = \lim W_n^\parallel \in K(\rho)$.
Moreover $Z_n/s_n$ is bounded; pass to a further subsequence with $Z_n/s_n \to W' \in K(\rho)^\perp$.
Dividing the displayed identity by $s_n$ and letting $n \to \infty$ gives $T_\rho(W') = -T_\psi(W)$, so $T_\psi(W) \in \operatorname{Im}(T_\rho)$, i.e.\ $\overline{T}_\psi(W) = 0$.
As $W \in L$ was an arbitrary unit vector, the claim follows.

Claims~1 and~2 prove the lemma, and iterating it completes Step~1.
We obtain $\rho_1 \in C^\infty_{>0}(\mathcal{M})$ arbitrarily close to $\rho_0$ with $k(\rho_1) = 0$.
Since $\operatorname{Stab}(\rho_1)$ is a closed subgroup of the compact group $G$ with trivial Lie algebra, it is discrete, hence finite.

    \medskip

\emph{Step 2: Break the remaining finite symmetries.}
Let $\operatorname{Stab}(\rho_1) = \{\operatorname{id}, \varphi_1, \dots, \varphi_N\}$.
Since $k(\rho_1) = 0$, the linear map $T_{\rho_1}$ is injective, so $\operatorname{Im}(T_{\rho_1})$ is a subspace of dimension $m$.

\emph{Choice of the bumps.}
For each $j$, choose $p_j$ with $\varphi_j(p_j) \neq p_j$, which exists because $\varphi_j \neq \operatorname{id}$ moves some interior point, and an open set $U_j \ni p_j$ with $U_j \cap \varphi_j(U_j) = \emptyset$, which exists because $\varphi_j$ is continuous.
A bump $\psi_j \in C_c^\infty(U_j)$ then sits at points that $\varphi_j$ moves, but not at their images, as in Section~\ref{sec:identifiability}.
For bumps $\Psi = (\psi_1, \dots, \psi_N)$ with $\psi_i \in C_c^\infty(U_i)$, put
    $$\Delta_j(\Psi):= \sum_{i=1}^N \bigl(\psi_i\circ\varphi_j - \psi_i\bigr), \qquad j = 1, \dots, N.$$
We choose $\Psi$ with $\Delta_j(\Psi) \notin \operatorname{Im}(T_{\rho_1})$ for every $j$, which is possible for the following reason.
Each $\Delta_j$ is linear in $\Psi$, so $B_j:= \{\Psi : \Delta_j(\Psi) \in \operatorname{Im}(T_{\rho_1})\}$ is a linear subspace of $E:= \prod_{i=1}^N C_c^\infty(U_i)$.
It is a proper subspace.
On the $j$-th factor alone, $\Delta_j$ is the map $\psi_j \mapsto \psi_j\circ\varphi_j - \psi_j$, and this map is injective on $C_c^\infty(U_j)$, because $\psi_j\circ\varphi_j$ vanishes on $U_j$ when $U_j \cap \varphi_j(U_j) = \emptyset$, so $\psi_j\circ\varphi_j = \psi_j$ forces $\psi_j = 0$ on $U_j$ and hence everywhere.
Its image is therefore infinite-dimensional and not contained in the $m$-dimensional space $\operatorname{Im}(T_{\rho_1})$, so some element of $E$ lies outside $B_j$.
A real vector space is not a finite union of proper subspaces, so some $\Psi \in E$ lies outside $B_1 \cup \dots \cup B_N$.
Fix such a $\Psi$ and write $\Delta_j:= \Delta_j(\Psi)$.

For $\varepsilon > 0$ let
    $$\rho_2^\varepsilon:= \rho_1 + \varepsilon \sum_{j=1}^N \psi_j.$$
It is positive for small $\varepsilon$, because $\rho_1 > 0$ on the compact $\mathcal{M}$ and the $\psi_j$ are bounded, and $\rho_2^\varepsilon \to \rho_1$ in $C^\infty$ as $\varepsilon \to 0$.
Since $\rho_1\circ\varphi_j = \rho_1$,
\begin{equation}\label{eq:bump-defect}
    \rho_2^\varepsilon\circ\varphi_j - \rho_2^\varepsilon = \varepsilon\,\Delta_j \qquad \text{for } j = 1, \dots, N,
\end{equation}
and $\Delta_j \neq 0$ because $0 \in \operatorname{Im}(T_{\rho_1})$, so $\varphi_j \notin \operatorname{Stab}(\rho_2^\varepsilon)$ for every $\varepsilon > 0$.
This is the bump that breaks $\varphi_j$.
It remains to show that no new symmetry appears, that is, $\operatorname{Stab}(\rho_2^\varepsilon) = \{\operatorname{id}\}$ for all sufficiently small $\varepsilon > 0$.
Then $\rho_2^\varepsilon \in \mathcal{D}_{\mathrm{rigid}}$ is $C^\infty$-close to $\rho_1$, hence to $\rho_0$, which proves density.

\emph{No new symmetry appears.}
Suppose not.
Then there are $\varepsilon_n \to 0$ and $\varphi_n \in \operatorname{Stab}(\rho_2^{\varepsilon_n}) \setminus \{\operatorname{id}\}$.
Write $\rho_n:= \rho_2^{\varepsilon_n}$, so $\rho_n \to \rho_1$ in $C^\infty$.
By compactness of $G$, pass to a subsequence with $\varphi_n \to \varphi \in G$.
The first paragraph of the openness argument, applied to $\rho_n \to \rho_1$ and $\varphi_n \to \varphi$, gives $\varphi \in \operatorname{Stab}(\rho_1)$, so $\varphi = \operatorname{id}$ or $\varphi = \varphi_j$ for some $j$.

\emph{The limit is the identity.}
The second paragraph of the openness argument, applied to the same sequences, produces $Y \in K(\rho_1)$ with $\|Y\| = 1$.
Then $k(\rho_1) \geq 1$, which contradicts $k(\rho_1) = 0$.

\emph{The limit is $\varphi_j$.}
Put $\chi_n:= \varphi_j^{-1}\circ\varphi_n$, so $\chi_n \to \operatorname{id}$ and, for large $n$, $\chi_n = \exp(Y_n)$ with $Y_n \to 0$ in $\mathfrak{g}$.
Since $\rho_1\circ\varphi_j = \rho_1$, the identity $\rho_n\circ\varphi_n = \rho_n$ reads
\begin{equation}\label{eq:near-phi-j}
    \rho_1\circ\chi_n - \rho_1 = \varepsilon_n\Bigl(\sum_i\psi_i - \sum_i\psi_i\circ\varphi_j\circ\chi_n\Bigr).
\end{equation}
If $Y_n = 0$ for infinitely many $n$, then $\varphi_n = \varphi_j$ along them, and Equation~\ref{eq:bump-defect} gives $\varepsilon_n\Delta_j = 0$, contradicting $\Delta_j \neq 0$.
So $Y_n \neq 0$ for large $n$, and we write $Y_n = t_n\hat Y_n$ with $t_n:= \|Y_n\| \to 0^+$ and $\|\hat Y_n\| = 1$.
Pass to a subsequence along which $\hat Y_n \to \hat Y$ with $\|\hat Y\| = 1$ and $t_n/\varepsilon_n \to \lambda \in [0, \infty]$.
Fix $p \in \operatorname{int}\mathcal{M}$ and evaluate Equation~\ref{eq:near-phi-j} at $p$.
As in the openness argument, the fundamental theorem of calculus gives
    $$\rho_1(\chi_n(p)) - \rho_1(p) = \int_0^{t_n}(L_{\hat Y_n^\sharp}\rho_1)(\exp(s\hat Y_n)\cdot p)\,ds = t_n\bigl[(L_{\hat Y^\sharp}\rho_1)(p) + o(1)\bigr],$$
because $(Y, q) \mapsto (L_{Y^\sharp}\rho_1)(q)$ is continuous and $\exp(s\hat Y_n)\cdot p \to p$ uniformly in $s \in [0, t_n]$.
On the right-hand side, $\chi_n(p) \to p$ and the $\psi_i$ and $\varphi_j$ are continuous, so
    $$\varepsilon_n\Bigl(\sum_i\psi_i(p) - \sum_i\psi_i(\varphi_j(\chi_n(p)))\Bigr) = \varepsilon_n\bigl[-\Delta_j(p) + o(1)\bigr].$$
Hence
\begin{equation}\label{eq:bump-ratio}
    t_n\bigl[(L_{\hat Y^\sharp}\rho_1)(p) + o(1)\bigr] = \varepsilon_n\bigl[-\Delta_j(p) + o(1)\bigr].
\end{equation}
If $\lambda = \infty$, divide Equation~\ref{eq:bump-ratio} by $t_n$ and let $n \to \infty$.
This gives $(L_{\hat Y^\sharp}\rho_1)(p) = 0$, and as $p$ was arbitrary, $\hat Y \in K(\rho_1)$ with $\|\hat Y\| = 1$, which contradicts $k(\rho_1) = 0$.
If $\lambda < \infty$, divide by $\varepsilon_n$ instead.
This gives $\lambda\,(L_{\hat Y^\sharp}\rho_1)(p) = -\Delta_j(p)$, and as $p$ was arbitrary, $\Delta_j = T_{\rho_1}(-\lambda\hat Y) \in \operatorname{Im}(T_{\rho_1})$, which contradicts the choice of $\Psi$.

Both limits are impossible, so $\operatorname{Stab}(\rho_2^\varepsilon) = \{\operatorname{id}\}$ for all sufficiently small $\varepsilon > 0$.
This completes the density argument.
\end{proof}

\subsection{Proofs for Section~\ref{sec:recovery}}

\begin{proposition}[Proposition~\ref{prop:unique-gw-coupling}]\label{prop:unique-gw-coupling-appendix}
Under \ref{H1}--\ref{H2}, the optimal couplings of Equation~\ref{eq:GWdist} are exactly the couplings $(\operatorname{id}\times T)_{\#}P_a$ induced by isomorphisms $T$ of $(\mathcal{Z}_a, d_a, P_a)$ and $(\mathcal{Z}_b, d_b, P_b)$.
Under \ref{C1} the optimal coupling is unique and equals $\eta^\star = (\operatorname{id}\times A)_{\#}P_a$.
\end{proposition}

\begin{proof}
The objective $J_{\mathrm{GW}}$ of Equation~\ref{eq:GWdist} is nonnegative.

\emph{Isomorphisms give optimal couplings.}
Let $T$ be an isomorphism.
Since $T_{\#}P_a = P_b$, the coupling $(\operatorname{id}\times T)_{\#}P_a$ lies in $\Pi(P_a,P_b)$, and since $d_b(T(z),T(z')) = d_a(z,z')$ for all $z,z'$, its distortion vanishes.
The alignment $A$ is an isomorphism by Proposition~\ref{prop:alignment}, so the minimum is $0$ and the optimal couplings are exactly those with $J_{\mathrm{GW}}(\eta) = 0$.

\emph{Optimal couplings come from isomorphisms.}
Let $\eta\in\Pi(P_a,P_b)$ with $J_{\mathrm{GW}}(\eta) = 0$, and let $S:= \operatorname{supp}(\eta) \subseteq \mathcal{Z}_a\times\mathcal{Z}_b$.
The integrand $|d_a(z,z') - d_b(w,w')|^2$ is continuous and nonnegative, and $\operatorname{supp}(\eta\otimes\eta) = S\times S$, so
\begin{equation}\label{eq:pointwise-zero-app}
    d_a(z,z') = d_b(w,w') \qquad \text{for all } (z,w),(z',w')\in S.
\end{equation}
The projection of the compact set $S$ to $\mathcal{Z}_a$ is compact and carries full $P_a$-mass, so it contains $\operatorname{supp}(P_a) = \mathcal{Z}_a$, using that $P_a$ has full support (Section~\ref{sec:latentmmspaces}).
Likewise the projection to $\mathcal{Z}_b$ is $\mathcal{Z}_b$.
If $(z,w_1)$ and $(z,w_2)$ lie in $S$, Equation~\ref{eq:pointwise-zero-app} applied to these two pairs gives $d_b(w_1,w_2) = d_a(z,z) = 0$, so $w_1 = w_2$, and the same argument with the roles exchanged shows that $(z_1,w),(z_2,w)\in S$ forces $z_1=z_2$.
Hence $S$ is the graph of a bijection $T\colon\mathcal{Z}_a\to\mathcal{Z}_b$, and Equation~\ref{eq:pointwise-zero-app} says that $T$ preserves distances, so $T$ is continuous.
A coupling with first marginal $P_a$ that is supported on the graph of the continuous map $T$ equals $(\operatorname{id}\times T)_{\#}P_a$, and its second marginal gives $T_{\#}P_a = P_b$.
So $T$ is an isomorphism and $\eta$ is the coupling it induces.

\emph{Uniqueness under \ref{C1}.}
By Proposition~\ref{prop:identifiable}, \ref{C1} makes $A$ the only isomorphism, so $\eta^\star$ is the only optimal coupling.
\end{proof}

\begin{proposition}[Proposition~\ref{prop:convex-alignment}]
\label{prop:convex-alignment-appendix}
Under \ref{H1}--\ref{H2} and \ref{C2}, $\eta^\star = (\operatorname{id}\times A)_{\#}P_a$ is the unique optimal coupling of the LDD-LP in Equation~\ref{eq:sinkhorn_problem}.
\end{proposition}

\begin{proof}
Write $c(z,w):= W_2^2\big(\Phi_a(z,\cdot),\Phi_b(w,\cdot)\big)$ for the integrand of Equation~\ref{eq:sinkhorn_problem}, so that $J_{\mathrm{LDD}}(\eta)=\int c\,\mathrm{d}\eta$.
Since $A$ is an isomorphism (Proposition~\ref{prop:alignment-appendix}), $\Phi_b(A(z),\cdot) = \Phi_a(z,\cdot)$ for every $z\in\mathcal{Z}_a$ (Section~\ref{sec:ldd_sinkhorn}), so $\eta^\star\in\Pi(P_a,P_b)$ has cost $\int c\,\mathrm{d}\eta^\star = 0$.
As $c\ge 0$, the minimum is zero and $\eta^\star$ attains it.
Let $\eta\in\Pi(P_a,P_b)$ be any minimiser.

Under \ref{H1}, distance spheres are $P_{\mathcal{M}}$-null (Remark~\ref{rem:null-spheres}), so every LDD $\Phi(x,\cdot)$ is continuous in $r$, and $x\mapsto\Phi(x,\cdot)$ is continuous in the supremum norm (Lemma~\ref{lem:sup-continuity}).
Hence the set $\{(x,x'): x\neq x',\ \Phi(x,\cdot) = \Phi(x',\cdot)\}$ is the intersection of a closed set with an open set in the compact space $\mathcal{M}\times\mathcal{M}$, hence $\sigma$-compact, and the collision set $\mathcal{C}$ of \ref{C2}, its projection onto the first factor, is $\sigma$-compact and in particular Borel.
Let $\mathcal{C}_a:= \{z: \Phi_a(z',\cdot) = \Phi_a(z,\cdot)\text{ for some }z'\neq z\}$ be the collision set of $\mathcal{Z}_a$ and $\mathcal{C}_b$ that of $\mathcal{Z}_b$.
Since $f_a$ and $f_b$ are isomorphisms onto the data mm-space, $\mathcal{C}_a = f_a^{-1}(\mathcal{C})$ and $\mathcal{C}_b = f_b^{-1}(\mathcal{C}) = A(\mathcal{C}_a)$, so $P_a(\mathcal{C}_a) = P_b(\mathcal{C}_b) = P_{\mathcal{M}}(\mathcal{C}) = 0$.

The cost $c$ is continuous, because $W_2\big(\Phi_a(z,\cdot),\Phi_a(z',\cdot)\big)\le d_a(z,z')$ by coupling the two distance distributions through one random data point, and likewise for $b$, and it is nonnegative, so $\int c\,\mathrm{d}\eta = 0$ forces $c(z,w) = 0$ for $\eta$-almost every $(z,w)$.
For such a pair, $\Phi_a(z,\cdot)$ and $\Phi_b(w,\cdot)$ agree for almost every $r$, and since both are continuous in $r$ they agree for every $r$.
Also $\eta(\mathcal{C}_a\times\mathcal{Z}_b) = P_a(\mathcal{C}_a) = 0$.
Disintegrate $\eta = \int\delta_z\otimes\eta_z\,\mathrm{d}P_a(z)$.
For $P_a$-almost every $z$, then, $z\notin\mathcal{C}_a$ and $\eta_z$ is supported on $\{w: \Phi_b(w,\cdot) = \Phi_a(z,\cdot)\}$.
This set contains $A(z)$ and no other point.
A second $w\neq A(z)$ with $\Phi_b(w,\cdot) = \Phi_b(A(z),\cdot)$ would put $A(z)$ in $\mathcal{C}_b$, that is, $z\in\mathcal{C}_a$.
Hence $\eta_z = \delta_{A(z)}$ for $P_a$-almost every $z$, and $\eta = (\operatorname{id}\times A)_{\#}P_a = \eta^\star$.
\end{proof}

The proof uses \ref{C2} directly and not \ref{C1}.
That \ref{C2} implies \ref{C1}, as stated in Section~\ref{sec:ldd_sinkhorn}, is Lemma~\ref{lem:a2-implies-a1}.

\begin{lemma}[Generic samples]
\label{lem:multiset}\label{lem:unique-permutation}
Let $\rz_1,\dots,\rz_n$, $n\geq 3$, be i.i.d.\ from $P_a$.
Almost surely the sample points are distinct, the pairwise distances $d_a(\rz_i,\rz_j)$, $i<j$, are distinct, the rows of the distance matrix are distinct as multisets, and no non-trivial permutation of the sample preserves all pairwise distances.
\end{lemma}

\begin{proof}
The distribution of representations $P_a = (f_a^{-1})_{\#}P_{\mathcal{M}}$ has a density with respect to the volume of $g_a$, because $P_{\mathcal{M}}$ has one with respect to the volume of $\mathcal{M}$ and $f_a$ is an isometry onto $\mathcal{M}$.
In particular $P_a$ has no atoms, so the sample points are almost surely distinct.
Distance spheres have volume zero (Remark~\ref{rem:null-spheres}), so they have $P_a$-measure zero.

All $\binom{n}{2}$ pairwise distances are almost surely distinct.
If two distinct index pairs satisfied $d_a(\rz_p,\rz_q) = d_a(\rz_r,\rz_s)$, some index would lie in exactly one of the pairs.
Conditionally on all other sample points, that point is forced onto a distance sphere, which has probability zero.
Integrating over the conditioning and taking a union over the finitely many pairs of index pairs gives the claim.

Let $R_i:= \{d_a(\rz_i, \rz_l): l\neq i\}$ be the multiset of entries of row $i$.
For $i\neq j$, the multisets $R_i$ and $R_j$ share the entry $d_a(\rz_i,\rz_j)$, and their remaining entries are distances between distinct index pairs, hence pairwise distinct.
Since $n\geq 3$, each row has at least one remaining entry, so $R_i\neq R_j$.

Finally, suppose a permutation $\tau\neq\operatorname{id}$ preserves all pairwise distances, and pick $i$ with $\tau(i)\neq i$.
Then $R_{\tau(i)} = \{d_a(\rz_{\tau(i)}, \rz_{\tau(l)}): l\neq i\} = \{d_a(\rz_i,\rz_l): l\neq i\} = R_i$, which contradicts the distinctness of rows.
\end{proof}

\begin{proposition}[Proposition~\ref{prop:perm-exact}]
\label{prop:perm-exact-appendix}
In the permutation regime, under \ref{H1}--\ref{H2} and with $n\geq 3$, the empirical $J_\beta$ almost surely has the unique optimal coupling $\eta_\sigma:= \frac1n\sum_i\delta_{(\rz^a_i,\,\rz^b_{\sigma(i)})}$ for every $\beta\in[0,1]$.
This holds whether or not \ref{C1} holds.
\end{proposition}

\begin{proof}
Write $\eta\in\Pi(\hat P_a,\hat P_b)$ as an $n\times n$ matrix with row and column sums $1/n$, the empirical GW objective as $J_{\mathrm{GW}}(\eta) = \sum_{i,j,k,l}\big|C^a_{ik} - C^b_{jl}\big|^2\eta_{ij}\eta_{kl}$, and the empirical LDD-LP cost as $c_{ij} = W_2^2\big(\hat\Phi_a(\rz^a_i,\cdot), \hat\Phi_b(\rz^b_j,\cdot)\big)$, where $\hat\Phi_a(\rz^a_i,r) = \frac1n\#\{k: C^a_{ik}\le r\}$ is the empirical distribution function of row $i$ of $C^a$, and likewise for $b$.
Both objectives are nonnegative.
We work on the almost-sure event of Lemma~\ref{lem:multiset}.

\emph{The coupling along $\sigma$ is optimal for both problems.}
Since $A$ is an isometry, $C^b_{\sigma(i)\sigma(k)} = d_b(A(\rz^a_i),A(\rz^a_k)) = d_a(\rz^a_i,\rz^a_k) = C^a_{ik}$, so $J_{\mathrm{GW}}(\eta_\sigma) = 0$.
The same identity shows that row $\sigma(i)$ of $C^b$ and row $i$ of $C^a$ are equal as multisets.
Their empirical distribution functions therefore coincide, $c_{i\sigma(i)} = 0$, and $\eta_\sigma$ has zero LDD cost.

\emph{GW has no other optimum.}
Let $J_{\mathrm{GW}}(\eta) = 0$, and let $S:= \{(i,j): \eta_{ij} > 0\}$.
Every term of $J_{\mathrm{GW}}$ with $(i,j),(k,l)\in S$ vanishes, so $C^a_{ik} = C^b_{jl}$ for all such pairs.
If $(i,j)$ and $(i,j')$ lie in $S$, this gives $C^b_{jj'} = C^a_{ii} = 0$, so $j=j'$ because the sample points are distinct, and symmetrically $(i,j),(i',j)\in S$ forces $i=i'$.
The marginal constraints put every $i$ and every $j$ in at least one pair of $S$, so $S$ is the graph of a permutation $\sigma'$ with $C^b_{\sigma'(i)\sigma'(k)} = C^a_{ik}$.
The permutation $\tau:= \sigma^{-1}\circ\sigma'$ then satisfies $C^a_{\tau(i)\tau(k)} = C^b_{\sigma'(i)\sigma'(k)} = C^a_{ik}$, so it preserves all pairwise distances of the $a$-sample, and Lemma~\ref{lem:multiset} gives $\tau = \operatorname{id}$.
Hence $\sigma' = \sigma$, and the marginals force $\eta = \eta_\sigma$.

\emph{The LDD-LP has no other optimum.}
Two distributions on the line are at $W_2$ distance zero only if they are equal, and an empirical distribution function on a finite multiset determines the multiset.
So $c_{ij} = 0$ exactly when row $i$ of $C^a$ and row $j$ of $C^b$ are equal as multisets.
Row $j$ of $C^b$ equals row $\sigma^{-1}(j)$ of $C^a$, and the rows of $C^a$ are distinct, so $c_{ij} = 0$ exactly when $j = \sigma(i)$.
A coupling with zero cost is therefore supported on the graph of $\sigma$, and the marginals force it to be $\eta_\sigma$.

\emph{The fused problem has no other optimum.}
The empirical objective $J_\beta(\eta) = (1-\beta)J_{\mathrm{GW}}(\eta) + \beta\sum_{i,j}c_{ij}\eta_{ij}$ is GW at $\beta=0$ and the LDD-LP at $\beta=1$.
For $\beta\in(0,1)$ it is nonnegative and vanishes at $\eta_\sigma$ by the first part, and its zeros are zeros of $J_{\mathrm{GW}}$, so $\eta_\sigma$ is its unique optimum by the second part.

None of the arguments uses \ref{C1} or \ref{C2}.
\end{proof}

\begin{lemma}\label{lem:perm-pinning}
In the permutation regime under \ref{H1}--\ref{H2}, almost surely for all sufficiently large $n$, the alignment $A$ is the only distance-preserving bijection $T\colon(\mathcal{Z}_a,d_a)\to(\mathcal{Z}_b,d_b)$ with $T(\rz^a_i) = \rz^b_{\sigma(i)}$ for all $i$.
\end{lemma}
\begin{proof}
Let $T$ be such a bijection.
Since $A(\rz^a_i) = \rz^b_{\sigma(i)}$, the map $\varphi:= A^{-1}\circ T$ is a distance-preserving bijection of $(\mathcal{Z}_a,d_a)$ that fixes every $\rz^a_i$.
By Lemma~\ref{lem:distance-isometries} its restriction to $\operatorname{int}\mathcal{Z}_a$ is a Riemannian isometry.
It remains to find sample points $x_1,\dots,x_{d+1}$ as in Corollary~\ref{cor:pinning}.
Fix $p\in\operatorname{int}\mathcal{Z}_a$.
It has a totally normal neighbourhood $W\subset\operatorname{int}\mathcal{Z}_a$, so for some $\delta>0$ and every $x\in W$, $\exp_x$ is a diffeomorphism from the $\delta$-ball of $T_x\mathcal{Z}_a$ onto a set containing $W$ \citep[Chapter~3, Theorem~3.7]{docarmo1992riemannian}.
Since $P_a$ has full support, $P_a(W)>0$, and by the strong law of large numbers almost surely at least $d+1$ of the $\rz^a_i$ lie in $W$ for all large $n$.
Let $x_1$ be the first of them.
The others lie in a normal ball around $x_1$, and given $x_1$ they are independent draws from $P_a$ conditioned on $W$, which is absolutely continuous.
By the remark following Corollary~\ref{cor:pinning}, their log-map images at $x_1$ are then linearly independent almost surely.
Corollary~\ref{cor:pinning} gives $\varphi=\operatorname{id}$, so $T=A$.
\end{proof}

\begin{proposition}[Consistency in the independent regime]
\label{prop:consistency}
In the independent regime under \ref{H1}--\ref{H2}, let $n\to\infty$.
\emph{(i)} Under \ref{C1}, every sequence of global minimisers $\hat\eta_n$ of the empirical GW problem, or of the empirical $J_\beta$ for a fixed $\beta\in(0,1)$, converges weakly to $\eta^\star$ almost surely.
\emph{(ii)} Under \ref{C2}, the same holds for minimisers of the empirical LDD-LP.
\emph{(iii)} In each case, let $\hat T_n$ minimise $\int d_b\big(T(z),w\big)^2\,\mathrm{d}\hat\eta_n(z,w)$ over a class $\mathcal{T}$ of continuous maps $\mathcal{Z}_a\to\mathcal{Z}_b$ that is compact in the uniform topology and contains $A$.
Then $\hat T_n\to A$ uniformly on $\mathcal{Z}_a$ almost surely.
\end{proposition}

\begin{proof}
All spaces are compact metric spaces, so weak convergence of probability measures on them is metrised by the Wasserstein distance $W_1$ \citep[Theorem~6.9]{villani2009optimal}, and the space of probability measures on $\mathcal{Z}_a\times\mathcal{Z}_b$ is compact.
The empirical measures $\hat P_a^n$ and $\hat P_b^n$ converge weakly to $P_a$ and $P_b$ almost surely \citep{billingsley1999convergence}.
We fix an outcome in this event and in the almost-sure events used below.

\emph{Approximating couplings.}
For every $n$ there is $\eta_n\in\Pi(\hat P_a^n,\hat P_b^n)$ with $\eta_n\to\eta^\star$ weakly.
To construct it, take optimal couplings $\alpha_n\in\Pi(\hat P_a^n,P_a)$ and $\beta_n\in\Pi(P_b,\hat P_b^n)$ for $W_1$, glue $\alpha_n$, $\eta^\star$ and $\beta_n$ along their shared marginals \citep[Chapter~1]{villani2009optimal}, and let $\eta_n$ be the marginal of the glued measure on the first and last factors.
Then $\eta_n\in\Pi(\hat P_a^n,\hat P_b^n)$ and $W_1(\eta_n,\eta^\star)\le W_1(\hat P_a^n,P_a) + W_1(P_b,\hat P_b^n)\to 0$, for the sum metric on the product.

\emph{(i)} The objective $J_{\mathrm{GW}}$ of Equation~\ref{eq:GWdist} is defined for every probability measure $\eta$ on $\mathcal{Z}_a\times\mathcal{Z}_b$.
Its integrand is continuous and bounded, and $\eta_n\to\eta$ weakly implies $\eta_n\otimes\eta_n\to\eta\otimes\eta$ weakly, so $J_{\mathrm{GW}}$ is weakly continuous.
The empirical GW problem minimises $J_{\mathrm{GW}}$ over $\Pi(\hat P_a^n,\hat P_b^n)$, so $J_{\mathrm{GW}}(\hat\eta_n)\le J_{\mathrm{GW}}(\eta_n)\to J_{\mathrm{GW}}(\eta^\star) = 0$.
Let $\bar\eta$ be the weak limit of a subsequence of $(\hat\eta_n)$, which exists by compactness.
Its marginals are the limits of the empirical marginals, so $\bar\eta\in\Pi(P_a,P_b)$, and $J_{\mathrm{GW}}(\bar\eta) = 0$ by continuity.
Under \ref{C1}, Proposition~\ref{prop:unique-gw-coupling} gives $\bar\eta = \eta^\star$.
Every subsequence thus has a further subsequence converging to $\eta^\star$, so the whole sequence does.

\emph{(ii)} Write $\Phi_a(z,r):= P_a(\bar B_a(z,r))$ and $\hat\Phi_a^n(z,r):= \hat P_a^n(\bar B_a(z,r))$, and likewise for $b$, for $r$ in $[0,R]$ with $R$ the diameter of $\mathcal{M}$, which $\mathcal{Z}_a$ and $\mathcal{Z}_b$ share.
We first show $\sup_{z,r}|\hat\Phi_a^n(z,r) - \Phi_a(z,r)|\to 0$ almost surely.
The map $z\mapsto\Phi_a(z,\cdot)$ is continuous in the supremum norm (Lemma~\ref{lem:sup-continuity}), so its image is compact in the continuous functions on $[0,R]$, hence equicontinuous, and $\Phi_a$ is uniformly continuous on $\mathcal{Z}_a\times[0,R]$.
Given $\varepsilon>0$, choose $\delta>0$ with $|\Phi_a(z,r) - \Phi_a(z',r')| < \varepsilon$ whenever $d_a(z,z')\le\delta$ and $|r-r'|\le\delta$, and a finite $\delta$-net $z_1,\dots,z_K$ of $\mathcal{Z}_a$.
For $d_a(z,z_k)\le\delta$ the triangle inequality gives $\bar B_a(z_k,r-\delta)\subseteq\bar B_a(z,r)\subseteq\bar B_a(z_k,r+\delta)$, hence $\hat\Phi_a^n(z_k,r-\delta)\le\hat\Phi_a^n(z,r)\le\hat\Phi_a^n(z_k,r+\delta)$.
For each of the finitely many $z_k$, $r\mapsto\hat\Phi_a^n(z_k,r)$ is the empirical distribution function of $d_a(z_k,\cdot)$ under $\hat P_a^n$, which converges uniformly in $r$ almost surely by the Glivenko--Cantelli theorem.
Combining the sandwich with uniform continuity gives $\limsup_n\sup_{z,r}|\hat\Phi_a^n(z,r) - \Phi_a(z,r)|\le\varepsilon$, and letting $\varepsilon\to 0$ along a sequence proves the claim.
The empirical cost $c_n(z,w):= W_2^2\big(\hat\Phi_a^n(z,\cdot),\hat\Phi_b^n(w,\cdot)\big)$ therefore converges uniformly to the population cost $c$, because $W_2^2\le R\,W_1\le R^2\|F-F'\|_\infty$ for two distributions on $[0,R]$ with distribution functions $F$ and $F'$, and $|W_2^2(\mu,\nu) - W_2^2(\mu',\nu')|\le 2R\big(W_2(\mu,\mu') + W_2(\nu,\nu')\big)$ for distributions on $[0,R]$.
The population cost is continuous because $W_2\big(\Phi_a(z,\cdot),\Phi_a(z',\cdot)\big)\le d_a(z,z')$, by coupling the two distance distributions through one random data point.
Then
$$\int c\,\mathrm{d}\hat\eta_n \le \int c_n\,\mathrm{d}\hat\eta_n + \|c_n - c\|_\infty \le \int c_n\,\mathrm{d}\eta_n + \|c_n - c\|_\infty \le \int c\,\mathrm{d}\eta_n + 2\|c_n - c\|_\infty \to \int c\,\mathrm{d}\eta^\star = 0.$$
Any weak limit point $\bar\eta$ of $(\hat\eta_n)$ lies in $\Pi(P_a,P_b)$ and has $\int c\,\mathrm{d}\bar\eta = 0$, so it is optimal for the LDD-LP, and under \ref{C2} Proposition~\ref{prop:convex-alignment} gives $\bar\eta = \eta^\star$.
The subsequence argument of (i) concludes.

\emph{The fused problem.}
For $\beta\in(0,1)$, let $\hat\eta_n$ minimise $J^n_\beta(\eta):=(1-\beta)J_{\mathrm{GW}}(\eta)+\beta\int c_n\,\mathrm{d}\eta$ over $\Pi(\hat P_a^n,\hat P_b^n)$.
Then $(1-\beta)J_{\mathrm{GW}}(\hat\eta_n)\le J^n_\beta(\hat\eta_n)\le J^n_\beta(\eta_n)\to 0$, by the continuity of $J_{\mathrm{GW}}$ in (i) and the uniform convergence of $c_n$ in (ii), so every limit point $\bar\eta$ has $J_{\mathrm{GW}}(\bar\eta)=0$, and under \ref{C1} it equals $\eta^\star$ by Proposition~\ref{prop:unique-gw-coupling}.
The subsequence argument of (i) concludes.

\emph{(iii)} Let $F_\eta(T):= \int d_b(T(z),w)^2\,\mathrm{d}\eta(z,w)$, and let $D$ be the diameter of $\mathcal{Z}_b$.
For maps $T,T'$, $|F_\eta(T) - F_\eta(T')|\le 2D\sup_z d_b(T(z),T'(z))$ for every $\eta$, so each $F_\eta$ is continuous on $\mathcal{T}$ and minimisers exist.
Given $\varepsilon>0$, take a finite $\varepsilon$-net $T_1,\dots,T_K$ of $\mathcal{T}$ in the uniform metric.
Each $(z,w)\mapsto d_b(T_k(z),w)^2$ is continuous and bounded, so $F_{\hat\eta_n}(T_k)\to F_{\eta^\star}(T_k)$ by (i) or (ii), and
$$\sup_{T\in\mathcal{T}}\big|F_{\hat\eta_n}(T) - F_{\eta^\star}(T)\big| \le 4D\varepsilon + \max_k\big|F_{\hat\eta_n}(T_k) - F_{\eta^\star}(T_k)\big|.$$
Hence $F_{\hat\eta_n}\to F_{\eta^\star}$ uniformly on $\mathcal{T}$.
The limit is $F_{\eta^\star}(T) = \int d_b(T(z),A(z))^2\,\mathrm{d}P_a(z)$, which vanishes exactly when $T = A$ $P_a$-almost everywhere, and hence everywhere since $T$ and $A$ are continuous and $P_a$ has full support.
Let $\bar T$ be the uniform limit of a subsequence of $(\hat T_n)$, which exists by compactness.
Then $F_{\eta^\star}(\bar T) = \lim F_{\hat\eta_{n_k}}(\hat T_{n_k})\le\lim F_{\hat\eta_{n_k}}(A) = F_{\eta^\star}(A) = 0$, so $\bar T = A$, and the subsequence argument concludes.
\end{proof}

The next result makes the margin quantitative for the LDD-LP.
It needs one hypothesis beyond \ref{H1}--\ref{H2}, a uniform bound on how fast ball masses grow,
\begin{equation}\label{eq:ldd-lipschitz}
\big|\Phi(x,r) - \Phi(x,r')\big| \le K\,|r-r'| \qquad \text{for all } x\in\mathcal{M} \text{ and } r,r'\ge 0,
\end{equation}
that is, the distance from a point to a random data point has a density bounded by $K$, uniformly in the point.
On a closed manifold this follows from \ref{H1}.
Let $\|\rho\|_\infty$ bound the density of $P_{\mathcal{M}}$ and let $S(x,s)$ be the distance sphere of radius $s$ around $x$.
For $r' < r$, the coarea formula of Appendix~\ref{app:background}, with $|\nabla d_{\mathcal{M}}(x,\cdot)| = 1$ almost everywhere, gives
$$\Phi(x,r) - \Phi(x,r') = P_{\mathcal{M}}\big(\{r' < d_{\mathcal{M}}(x,\cdot)\le r\}\big) \le \|\rho\|_\infty\int_{r'}^{r}\mathcal{H}^{d-1}\big(S(x,s)\big)\,\mathrm{d}s.$$
The sphere $S(x,s)$ is contained in $\exp_x(\{v\in T_x\mathcal{M}: |v| = s\})$, whose $(d{-}1)$-dimensional area is at most $s^{d-1}$ times the integral over the unit sphere of $T_x\mathcal{M}$ of $\|\mathrm{d}(\exp_x)_{sv}\|^{d-1}$ by the area formula.
This is bounded uniformly in $x$ and $s\le R$, because $\exp$ is smooth on the compact set $\{w\in T\mathcal{M}: |w|\le R\}$, so Equation~\ref{eq:ldd-lipschitz} holds with $K$ equal to $\|\rho\|_\infty$ times that bound.
With a boundary we state it as a hypothesis.
Throughout, $R$ is the diameter of $\mathcal{M}$, which is also the diameter of $\mathcal{Z}_a$ and $\mathcal{Z}_b$, LDDs are functions on $[0,R]$, the empirical LDDs are $\hat\Phi_a(z,r):= \hat P_a(\bar B_a(z,r))$ and likewise for $b$, $c_n$ is the empirical LDD-LP cost built from them, and
$$\delta_n:= \sup_{z}W_2\big(\hat\Phi_a(z,\cdot),\Phi_a(z,\cdot)\big) + \sup_{w}W_2\big(\hat\Phi_b(w,\cdot),\Phi_b(w,\cdot)\big)$$
is the estimation error of the LDDs, with $W_2$ between the distributions on $[0,R]$ that have these distribution functions.
For such distributions $W_2^2\le R\,W_1\le R^2\|F-F'\|_\infty$, so $\delta_n$ is at most $R\sqrt2$ times the square root of the estimation error in the supremum norm over the point and the radius.
The bounds below hold for any two samples.
The sampling regime enters only through the rates at which $W_1(\hat P_a,P_a)$, $W_1(\hat P_b,P_b)$ and $\delta_n$ vanish, and in the independent regime all three tend to zero almost surely by the proof of Proposition~\ref{prop:consistency} and the inequality above.

\begin{proposition}[Stability of the LDD-LP]\label{prop:ldd-stability}
Assume \ref{H1}--\ref{H2} and Equation~\ref{eq:ldd-lipschitz} with constant $K$, and let $R$ be the diameter of $\mathcal{M}$.
\emph{(i)} For every probability measure $\eta$ on $\mathcal{Z}_a\times\mathcal{Z}_b$ and every $t$ with $m(t)>0$,
\begin{equation}\label{eq:ldd-stability}
\eta\big(\{(z,w): d_b(w,A(z))\ge t\}\big) \;\le\; \frac{K^2}{m(t)^3}\int c\,\mathrm{d}\eta.
\end{equation}
\emph{(ii)} For any two samples, a minimiser $\hat\eta_n$ of the empirical LDD-LP satisfies
\begin{equation}\label{eq:ldd-cost}
\int c\,\mathrm{d}\hat\eta_n \;\le\; 2R\Big[W_1(\hat P_a,P_a)+W_1(\hat P_b,P_b)+2\,\delta_n\Big],
\end{equation}
with $W_1$ taken in $d_a$ and $d_b$.
\emph{(iii)} Let $B_n(t)$ be $K^2/m(t)^3$ times the right-hand side of Equation~\ref{eq:ldd-cost}.
In the metric $\max(d_a,d_b)$ on $\mathcal{Z}_a\times\mathcal{Z}_b$,
$$d_P(\hat\eta_n,\eta^\star) \;\le\; \inf_{t>0}\,\max\big\{t,\ B_n(t)\big\} + d_P(\hat P_a,P_a).$$
\end{proposition}

\begin{proof}
Write $c(z,w) = W_2^2\big(\Phi_a(z,\cdot),\Phi_b(w,\cdot)\big)$ for the population cost and recall $\Phi_a(z,\cdot) = \Phi(f_a(z),\cdot)$ and $\Phi_b(w,\cdot) = \Phi(f_b(w),\cdot)$.
For two distributions on the line, the squared Wasserstein distance is the squared $L^2$ distance between their quantile functions $F^{-1}(u):= \inf\{r: F(r)\ge u\}$ \citep{villani2009optimal}, as in Section~\ref{sec:ldd_sinkhorn}.

\emph{LDDs are Lipschitz in the point.}
If $d_{\mathcal{M}}(x,x') = u$, the triangle inequality gives $\bar B_{\mathcal{M}}(x',r-u)\subseteq\bar B_{\mathcal{M}}(x,r)\subseteq\bar B_{\mathcal{M}}(x',r+u)$, and Equation~\ref{eq:ldd-lipschitz} gives $\|\Phi(x,\cdot) - \Phi(x',\cdot)\|_\infty\le Ku$.
Since $f_a$ and $f_b$ are isometries onto $\mathcal{M}$, the same holds for $\Phi_a$ in $d_a$ and for $\Phi_b$ in $d_b$.

\emph{(i) Mismatched pairs have large cost.}
Let $(z,w)$ satisfy $d_b(w,A(z))\ge t$, and put $x:= f_a(z) = f_b(A(z))$ and $y:= f_b(w)$.
Then $d_{\mathcal{M}}(x,y) = d_b(A(z),w)\ge t$, so $F:= \Phi(x,\cdot)$ and $G:= \Phi(y,\cdot)$ satisfy $\delta:= \|F - G\|_\infty\ge m(t)$.
By Equation~\ref{eq:ldd-lipschitz}, both are $K$-Lipschitz, so the supremum is attained at some $r_0$, and after exchanging the two we may assume $F(r_0) = G(r_0) + \delta$.
Let $u\in(G(r_0), F(r_0))$, an interval of length $\delta$.
Since $F$ is continuous, $F(F^{-1}(u)) = u$, and $F(r_0)\ge u$ gives $F^{-1}(u)\le r_0$; the Lipschitz bound $F(r_0) - u\le K\big(r_0 - F^{-1}(u)\big)$ then gives $F^{-1}(u)\le r_0 - (F(r_0) - u)/K$.
In the same way, $G(r_0) < u$ gives $G^{-1}(u) > r_0$ and $u - G(r_0)\le K\big(G^{-1}(u) - r_0\big)$, so $G^{-1}(u)\ge r_0 + (u - G(r_0))/K$.
Hence $G^{-1}(u) - F^{-1}(u)\ge\delta/K$ on an interval of length $\delta$, and
$$c(z,w) = \int_0^1\big(F^{-1}(u) - G^{-1}(u)\big)^2\,\mathrm{d}u \ge \delta\Big(\frac{\delta}{K}\Big)^2 = \frac{\delta^3}{K^2} \ge \frac{m(t)^3}{K^2}.$$
Markov's inequality gives Equation~\ref{eq:ldd-stability} for every probability measure $\eta$.

\emph{The cost is Lipschitz, and close to its empirical version.}
Let $x,x'\in\mathcal{M}$ with $d_{\mathcal{M}}(x,x') = u$ and let $Y\sim P_{\mathcal{M}}$.
The pair $\big(d_{\mathcal{M}}(x,Y), d_{\mathcal{M}}(x',Y)\big)$ couples the two distance distributions, and $|d_{\mathcal{M}}(x,Y) - d_{\mathcal{M}}(x',Y)|\le u$ by the triangle inequality, so $W_2\big(\Phi(x,\cdot),\Phi(x',\cdot)\big)\le u$.
All distances lie in $[0,R]$, so every $W_2$ between two LDDs is at most $R$, and $|a^2 - b^2|\le 2R|a - b|$ for $a,b\in[0,R]$.
The triangle inequality for $W_2$ then gives $|c(z,w) - c(z',w')|\le 2R\big(d_a(z,z') + d_b(w,w')\big)$, and, with $c_n(z,w) = W_2^2\big(\hat\Phi_a(z,\cdot),\hat\Phi_b(w,\cdot)\big)$, $\varepsilon_n:= \sup_{z,w}|c_n(z,w) - c(z,w)|\le 2R\,\delta_n$.

\emph{(ii) The computed coupling has small population cost.}
Let $\alpha\in\Pi(\hat P_a,P_a)$ and $\beta\in\Pi(\hat P_b,P_b)$ be $W_1$-optimal couplings in $d_a$ and $d_b$.
Gluing $\alpha$, $\eta^\star$ and $\beta$ along their common marginals gives a probability measure on quadruples $(\hat z, z, w, \hat w)$ with $(\hat z,z)\sim\alpha$, $(z,w)\sim\eta^\star$ and $(w,\hat w)\sim\beta$.
Its $(\hat z,\hat w)$-marginal $\eta_n$ lies in $\Pi(\hat P_a,\hat P_b)$, and the quadruple couples $\eta_n$ with $\eta^\star$ at expected displacement $\mathbb{E}[d_a(\hat z,z) + d_b(\hat w,w)] = W_1(\hat P_a,P_a) + W_1(\hat P_b,P_b)$.
Since $\int c\,\mathrm{d}\eta^\star = 0$, the Lipschitz bound on $c$ gives $\int c\,\mathrm{d}\eta_n\le 2R\big(W_1(\hat P_a,P_a) + W_1(\hat P_b,P_b)\big)$.
As $\hat\eta_n$ minimises $\int c_n$ over $\Pi(\hat P_a,\hat P_b)$ and $|c - c_n|\le\varepsilon_n$,
$$\int c\,\mathrm{d}\hat\eta_n \le \int c_n\,\mathrm{d}\hat\eta_n + \varepsilon_n \le \int c_n\,\mathrm{d}\eta_n + \varepsilon_n \le \int c\,\mathrm{d}\eta_n + 2\varepsilon_n \le 2R\Big[W_1(\hat P_a,P_a) + W_1(\hat P_b,P_b) + 2\,\delta_n\Big],$$
which is Equation~\ref{eq:ldd-cost}.

\emph{(iii) Prokhorov distance.}
Let $S_t:= \{(z,w): d_b(w,A(z))\ge t\}$, so that $\hat\eta_n(S_t)\le B_n(t)$ by (i) and (ii), and let $\tau(z,w):= (z, A(z))$, so that $\tau_{\#}\hat\eta_n = (\operatorname{id}\times A)_{\#}\hat P_a$.
Off $S_t$, the point $(z,w)$ lies at distance less than $t$ from $\tau(z,w)$.
For a Borel set $B$ and its open $s$-neighbourhood $B^s$, with $s:= \max\{t, B_n(t)\}$,
$$\hat\eta_n(B)\le\hat\eta_n(B\setminus S_t) + \hat\eta_n(S_t)\le(\tau_{\#}\hat\eta_n)(B^{s}) + s,$$
so $d_P\big(\hat\eta_n, (\operatorname{id}\times A)_{\#}\hat P_a\big)\le s$ \citep{billingsley1999convergence}.
The map $\operatorname{id}\times A$ is an isometric embedding of $(\mathcal{Z}_a, d_a)$ into $(\mathcal{Z}_a\times\mathcal{Z}_b,\max(d_a,d_b))$, because $A$ is an isometry, so $d_P\big((\operatorname{id}\times A)_{\#}\hat P_a,\eta^\star\big)\le d_P(\hat P_a,P_a)$.
The triangle inequality and the infimum over $t$ conclude.
\end{proof}

The bound does not involve $A$ on its right-hand side, and its margin is the supremum-norm margin of Definition~\ref{def:margin} that the test of Section~\ref{sec:ldd_sinkhorn} estimates.
If in addition $m(t)\ge a\,t^{\alpha}$ for small $t$, choosing $t$ to balance the two terms of the Prokhorov bound turns it into a rate in the sampling and estimation errors.
Of these, the $W_1$ terms decay as $n^{-1/d}$ for $d\ge 3$, the rate of \citet{fournier2015rate} for measures on $\mathbb{R}^d$, which transfers to $(\mathcal{Z}_a, d_a)$ and $(\mathcal{Z}_b, d_b)$ through finitely many bi-Lipschitz charts of the compact $d$-dimensional $\mathcal{M}$.
We do not evaluate either bound, since $K$, $a$ and $\alpha$ are unknown for learned representations, and at practical sample sizes the constants make the bound uninformative.
We give no analogous bound for GW, since \ref{C1} carries no margin.

\subsubsection{Fused GW}
\label{app:fused}

This subsection proves the uniqueness of the optimum of the fused objective $J_\beta$ of Equation~\ref{eq:fused} stated in Section~\ref{sec:ldd_sinkhorn}, for $\beta\in[0,1)$ and with $c$ the cost of Equation~\ref{eq:sinkhorn_problem}.
For a coupling $\eta$, let $L_\eta(z,w):= \iint|d_a(z,z')-d_b(w,w')|^2\,\mathrm{d}\eta(z',w')$, so that the GW objective at $\eta$ is $\langle L_\eta,\eta\rangle$ and the derivative of $J_\beta$ at $\eta$ in the direction $\eta'-\eta$ is $\langle 2(1-\beta)L_\eta+\beta c,\ \eta'-\eta\rangle$.

\begin{lemma}[The LDD-LP term is a pointwise lower bound]\label{lem:fused-lower}
For every $\eta\in\Pi(P_a,P_b)$ and all $(z,w)$, $c(z,w)\le L_\eta(z,w)$.
In particular, the LDD-LP objective is at most the GW objective at every coupling.
\end{lemma}

\begin{proof}
Pushing $\eta$ through $(z',w')\mapsto\big(d_a(z,z'),d_b(w,w')\big)$ gives a coupling of the distance distributions of $z$ and $w$, because the marginals of $\eta$ are $P_a$ and $P_b$.
Its cost is $L_\eta(z,w)$, and $c(z,w)=W_2^2$ is the least cost of such a coupling.
\end{proof}

The lemma puts the two terms of $J_\beta$ in the same units, so the weight $\beta$ compares like with like and neither term needs rescaling.

\begin{proposition}[Unique optimum]\label{prop:fused-unique}
Under \ref{H1}--\ref{H2} and \ref{C1}, $\eta^\star$ is the unique minimiser of $J_\beta$ for every $\beta\in[0,1)$.
\end{proposition}

\begin{proof}
Both terms are nonnegative.
Both vanish at $\eta^\star$, the GW term because $A$ is an isometry and the LDD-LP term because $A$ preserves LDDs (Section~\ref{sec:ldd_sinkhorn}).
Since $1-\beta>0$, a zero of $J_\beta$ is a zero of the GW objective, and under \ref{C1} the only one is $\eta^\star$ (Proposition~\ref{prop:unique-gw-coupling}).
\end{proof}

$J_\beta$ is not convex for $\beta<1$, since its second derivative is $(1-\beta)$ times that of the GW objective, which is negative in some directions, for example when two points exchange the points they are coupled to.
We prove no guarantee for its stationary points.
What the linear term adds is visible at the independent coupling $P_a\otimes P_b$, where solvers usually start.

\begin{proposition}[Linearisation at the independent coupling]\label{prop:fused-indep}
Let $e_a(z)$ and $v_a(z)$ be the mean and variance of $d_a(z,Z')$ for $Z'\sim P_a$, and $e_b(w)$ and $v_b(w)$ those of $d_b(w,W')$ for $W'\sim P_b$.
Then $L_{P_a\otimes P_b}(z,w)=\big(e_a(z)-e_b(w)\big)^2+v_a(z)+v_b(w)$.
\end{proposition}

\begin{proof}
Under $P_a\otimes P_b$, the distances $s=d_a(z,Z')$ and $s'=d_b(w,W')$ are independent, so $\mathbb{E}(s-s')^2=\mathbb{E}s^2+\mathbb{E}s'^2-2\,\mathbb{E}s\,\mathbb{E}s'=\big(e_a(z)-e_b(w)\big)^2+v_a(z)+v_b(w)$.
\end{proof}

The terms $v_a(z)$ and $v_b(w)$ add the same constant to the cost of every coupling, so a transport step from $P_a\otimes P_b$ with this cost compares points only by their mean distance to all others.
The step for $J_\beta$ adds $\beta c$ to the cost and compares their whole LDDs.

%% file: sections/appendices/appendix_mixture_rigid.tex
\subsection{Symmetry breaking in vMF mixtures}
\label{app:mixture-rigidity}\label{app:vmf-rigidity}

Section~\ref{sec:identifiability} builds $P_{\mathrm{mix}}$ from three vMF components whose placement breaks every symmetry of the sphere (Figure~\ref{fig:sphere-cases}, right).
This subsection proves that it does, for vMF mixtures on $S^d$.
For $c\in S^d\subset\mathbb{R}^{d+1}$ and $\kappa>0$, the vMF density with respect to the round volume is $p_{c,\kappa}(x) = C_d(\kappa)\,e^{\kappa\langle c,x\rangle}$, with $C_d(\kappa)$ the normalising constant.
Every isometry of the round sphere is the restriction of an orthogonal map $Q\in O(d+1)$ \citep{docarmo1992riemannian}.
$Q$ preserves the volume, and $\langle c,Q^{-1}x\rangle = \langle Qc,x\rangle$, so it maps one component onto another, $Q_{\#}(p_{c,\kappa}\operatorname{dVol}) = p_{Qc,\kappa}\operatorname{dVol}$.
A symmetry of a mixture therefore permutes its components, because distinct components are linearly independent:

\begin{lemma}[Linear independence of spherical exponentials]
\label{lem:vmf-independence}
Let $\theta_1, \dots, \theta_K \in \mathbb{R}^{d+1}$ ($d \geq 1$) be pairwise distinct.
Then the functions $x \mapsto e^{\langle \theta_m, x\rangle}$, $m = 1, \dots, K$, are linearly independent on $S^d$.
Consequently, vMF components with pairwise distinct parameters $(c_m, \kappa_m)$ are linearly independent, since $(c, \kappa) \mapsto \kappa c$ is injective on $S^d \times (0,\infty)$.
\end{lemma}

\begin{proof}
Suppose $\sum_m a_m e^{\langle \theta_m, x\rangle} = 0$ for all $x \in S^d$.
Choose $u \in S^d$ avoiding the finitely many proper hyperplanes $\{u: \langle \theta_m - \theta_l, u\rangle = 0\}$, $m \neq l$, so that the values $\alpha_m:= \langle \theta_m, u\rangle$ are pairwise distinct, and fix any unit $v \perp u$, setting $\beta_m:= \langle \theta_m, v\rangle$.
Restricting to the great circle $\gamma(t) = \cos t \cdot u + \sin t \cdot v$ gives
$$h(t):= \sum_{m=1}^{K} a_m\, e^{\alpha_m \cos t + \beta_m \sin t} = 0 \qquad \forall\, t \in \mathbb{R}.$$
The function $h$ extends to an entire function of $t \in \mathbb{C}$ and vanishes on $\mathbb{R}$, hence vanishes identically on $\mathbb{C}$.
Evaluate at $t = -iy$ with $y > 0$.
Since $\cos(-iy) = \cosh y$ and $\sin(-iy) = -i \sinh y$, the $m$-th term has modulus $|a_m|\, e^{\alpha_m \cosh y}$.
Let $m^\star$ be the index of the (unique) largest $\alpha_m$ among those with $a_m \neq 0$; if any coefficient is nonzero, then dividing $h(-iy) = 0$ by $e^{\alpha_{m^\star} \cosh y}$ and letting $y \to \infty$ leaves $|a_{m^\star}|$ on one side and $0$ on the other, a contradiction.
Hence all $a_m = 0$.
\end{proof}

\begin{proposition}[Rigidity of vMF mixtures]
\label{prop:vmf-rigidity-appendix}
Let $d \geq 2$ and let $\mu = \sum_{m=1}^{K} w_m\, p_{c_m, \kappa_m} \operatorname{dVol}$ on $(S^d, g_{S^d})$ with $K \geq d+1$, parameters $(c_m, \kappa_m)$ pairwise distinct, $w_m > 0$, $\sum_m w_m = 1$.
Assume
\begin{enumerate}
    \item[\emph{(i)}] some $d+1$ of the centres $c_1, \dots, c_K$ are
linearly independent in $\mathbb{R}^{d+1}$; and
    \item[\emph{(ii)}] at least one of: \emph{(a)} the weights $w_m$ are
pairwise distinct; \emph{(b)} the concentrations $\kappa_m$ are pairwise distinct; \emph{(c)} the pairwise geodesic distances $d(c_i, c_j)$, $i < j$, are pairwise distinct.
\end{enumerate}
Then $\operatorname{Aut}(S^d, g_{S^d}, \mu) = \{\operatorname{id}\}$, so \ref{C1} holds.
\end{proposition}

\begin{proof}
Let $Q\in\operatorname{Aut}(S^d, g_{S^d}, \mu)\subseteq O(d+1)$, so that $\sum_m w_m\,p_{Qc_m,\kappa_m} = \sum_m w_m\,p_{c_m,\kappa_m}$.
\emph{Step 0: $Q$ permutes the components.}
The parameters $(Qc_m,\kappa_m)$ are pairwise distinct, as are the $(c_m,\kappa_m)$.
Collect both sides over the union $T$ of the two parameter sets, $\sum_{\tau\in T}a_\tau\,p_\tau = 0$, where $a_\tau$ is the weight of $\tau$ on the left minus its weight on the right, each at most one $w_m$.
By Lemma~\ref{lem:vmf-independence}, every $a_\tau = 0$.
For $\tau = (Qc_m,\kappa_m)$ the left contributes $w_m>0$, so some unique $\pi(m)$ has $(c_{\pi(m)},\kappa_{\pi(m)}) = (Qc_m,\kappa_m)$ and $w_{\pi(m)} = w_m$.
The map $\pi$ is injective, since $Q$ is, so it is a permutation.

\emph{Step 1: $\pi = \operatorname{id}$.}
Under (ii)(a), $\pi$ preserves the pairwise distinct weights, so $\pi = \operatorname{id}$; under (ii)(b), the same argument applies to the concentrations.
Under (ii)(c), since $Q$ is an isometry, $d(c_{\pi(i)}, c_{\pi(j)}) = d(Qc_i, Qc_j) = d(c_i, c_j)$ for all $i < j$, and as all pairwise distances are distinct, $\pi$ maps each unordered pair $\{i,j\}$ to itself.
If $\pi \neq \operatorname{id}$, pick $i$ with $\pi(i) = j \neq i$ and any $l \notin \{i, j\}$, which exists since $K \geq d+1 \geq 3$.
Pair preservation gives $\pi(i) \in \{i, l\}$, yet $\pi(i) = j \notin \{i, l\}$, a contradiction.
Hence $\pi = \operatorname{id}$.

\emph{Step 2: $Q = I$.}
From $\pi = \operatorname{id}$ we get $Qc_m = c_m$ for all $m$.
By (i), $Q$ fixes $d+1$ linearly independent vectors in $\mathbb{R}^{d+1}$, that is, a basis, so $Q = I$.
\end{proof}

Condition (i) cannot be dropped, since the reflection in a hyperplane that contains every centre fixes each of them and preserves the mixture.
$P_{\mathrm{mix}}$ satisfies (i) and all three branches of (ii) (Appendix~\ref{app:exp-sphere}), so \ref{C1} holds for it.

%% file: sections/appendices/appendix_gap.tex
\section{The gap between \texorpdfstring{\ref{C1}}{C1} and \texorpdfstring{\ref{C2}}{C2}}
\label{app:ldd-multi-scale}

The two problems of Section~\ref{sec:ldd_sinkhorn} need different conditions on the data mm-space.
GW compares all pairwise distances, and under symmetry breaking (\ref{C1}) its optimal coupling is unique (Proposition~\ref{prop:unique-gw-coupling}).
The LDD-LP compares points only through their LDDs, and its optimal coupling is unique under distinguishable points (\ref{C2}) (Proposition~\ref{prop:convex-alignment}).
The second condition is the stronger one.
A symmetry preserves distances and $P_{\mathcal{M}}$, so it preserves every LDD, and under \ref{C2} it cannot move any point:

\begin{lemma}\label{lem:a2-implies-a1}
Under \ref{H1}, \ref{C2} implies \ref{C1}.
\end{lemma}

\begin{proof}
Let $S$ be a symmetry.
It maps every closed ball onto the closed ball of the same radius around the image of the centre, and it preserves $P_{\mathcal{M}}$, so $\Phi(S(x),\cdot) = \Phi(x,\cdot)$ for every $x$.
If $S(x)\neq x$, then $x$ shares its LDD with the distinct point $S(x)$, so $x\in\mathcal{C}$.
Hence $S = \operatorname{id}$ on $\mathcal{M}\setminus\mathcal{C}$, which has full measure by \ref{C2} and is therefore dense, since the positive density of \ref{H1} gives every open set positive mass.
$S$ is continuous, so $S = \operatorname{id}$ everywhere.
\end{proof}

\subsection{Collisions without symmetries}

The converse fails, because two points can share an LDD although no symmetry exchanges them.
Small balls show how much such a collision requires.
For $x$ in the interior of $\mathcal{M}$ and small $r$, the mass of a ball expands as
$$\Phi(x,r) = \omega_d r^d\rho(x)\big[1 + c_2(x)\,r^2 + O(r^4)\big],\qquad c_2(x) = -\frac{\operatorname{Scal}(x)}{6(d+2)} + \frac{\Delta\rho(x)}{2(d+2)\,\rho(x)},$$
where $d = \dim\mathcal{M}$, $\omega_d$ is the volume of the unit ball of $\mathbb{R}^d$, $\rho$ is the density of $P_{\mathcal{M}}$, and $\operatorname{Scal}$ and $\Delta$ are the scalar curvature and the Laplace--Beltrami operator of $g^{\mathcal{M}}$.
\citet{Gray1973} derives the expansion for $\rho\equiv 1$, and expanding $\rho$ in normal coordinates around $x$ adds the term with $\Delta\rho$.
Higher orders add coefficients $c_4, c_6, \dots$ built from the curvature, $\rho$ and their derivatives at $x$, and each of them can be read off $\Phi(x,\cdot)$ as $r\to 0$.
Two points with the same LDD therefore agree in $\rho$, in every coefficient and in the mass of every larger ball.
These are infinitely many equations on the $2d$ coordinates of the pair, so for a density without special structure we expect collisions to be rare and \ref{C2} to hold.
Unlike for \ref{C1} (Proposition~\ref{prop:trivial_automorphism}), we do not prove this.

A map that preserves every LDD, however, produces a collision at every point it moves.
A symmetry is such a map, and \ref{C1} rules it out.
A map that preserves $P_{\mathcal{M}}$ and every LDD, but not distances, is another, and on the following torus it moves almost every point:

\begin{proposition}\label{prop:torus}
Let $\mathcal{M}$ be the flat torus $\mathbb{R}^2/2\pi\mathbb{Z}^2$, embedded in $\mathbb{R}^4$ by $x\mapsto(\cos x_1,\sin x_1,\cos x_2,\sin x_2)$, and let $P_{\mathcal{M}}$ have the density
$$\rho(x) = \frac{1}{4\pi^2}\Big(1 + \tfrac14\cos x_1 + \tfrac14\cos(x_1+x_2) + \tfrac14\sin 2x_1\Big)$$
with respect to its area.
This data mm-space satisfies \ref{H1} and \ref{C1}.
The map $\tau(x) = (x_1,\,-2x_1-x_2)$ satisfies $\Phi(\tau(x),\cdot) = \Phi(x,\cdot)$ for every $x$, and $\tau(x)\neq x$ unless $x_1+x_2\in\pi\mathbb{Z}$, so the collision set has full measure and \ref{C2} fails.
\end{proposition}

\begin{proof}
The embedding has orthonormal partial derivatives, so it induces the flat metric $\mathrm{d}x_1^2+\mathrm{d}x_2^2$, its area is the Lebesgue measure on $[0,2\pi)^2$, and $d_{\mathcal{M}}$ is the distance of the flat torus.
$\mathcal{M}$ is a smooth, compact and connected surface in $\mathbb{R}^4$.
The density $\rho$ is smooth, positive because the three trigonometric terms add up to at most $\frac34$ in absolute value, and integrates to one because each of them integrates to zero, so \ref{H1} holds.

\emph{LDDs.}
Translations of the torus are isometries, so $d_{\mathcal{M}}(x,y) = \|y-x\|_{\mathcal{M}}$ with $\|v\|_{\mathcal{M}}:= d_{\mathcal{M}}(0,v)$, and $\|-v\|_{\mathcal{M}} = \|v\|_{\mathcal{M}}$.
Hence $\Phi(x,r) = \int\rho(x+v)\,\chi_r(v)\,\mathrm{d}v$ with $\chi_r(v):= \mathbf{1}\{\|v\|_{\mathcal{M}}\le r\}$, which is even in $v$.
For $k\in\mathbb{Z}^2$, evenness gives $\int\sin(k\cdot v)\,\chi_r(v)\,\mathrm{d}v = 0$, so with $\hat\chi_r(k):= \int\cos(k\cdot v)\,\chi_r(v)\,\mathrm{d}v$ the addition formulas give
$$\Phi(x,r) = \frac{1}{4\pi^2}\Big(\hat\chi_r(0) + \tfrac14\hat\chi_r(1,0)\cos x_1 + \tfrac14\hat\chi_r(1,1)\cos(x_1+x_2) + \tfrac14\hat\chi_r(2,0)\sin 2x_1\Big).$$
The map $\tau$ is $x\mapsto Bx$ with $B = \big(\begin{smallmatrix}1&0\\-2&-1\end{smallmatrix}\big)$.
$B$ has integer entries and determinant $-1$, so $\tau$ is a well-defined bijection of the torus, and $\tau\circ\tau = \operatorname{id}$.
It leaves $x_1$ and $2x_1$ unchanged and maps $x_1+x_2$ to $-(x_1+x_2)$, so the formula above gives $\Phi(\tau(x),r) = \Phi(x,r)$ for every $r$.
On the torus, $\tau(x) = x$ if and only if $2(x_1+x_2)\in 2\pi\mathbb{Z}$, which holds on two closed curves, a null set.
Every other point shares its LDD with the distinct point $\tau(x)$, so $\mathcal{C}$ has full measure.

\emph{Symmetry breaking.}
Let $S$ be a symmetry.
Since $\mathbb{R}^2$ is simply connected, $S$ lifts through the quotient map $\mathbb{R}^2\to\mathcal{M}$ to a continuous map $\tilde S$ of $\mathbb{R}^2$.
The quotient map preserves distances between points less than $\pi$ apart, so $\tilde S$ preserves Euclidean distances on a neighbourhood of every point and agrees there with a rigid motion.
Two rigid motions that agree on an open set are equal, and $\mathbb{R}^2$ is connected, so $\tilde S(x) = Ox+b$ with $O\in O(2)$ and $b\in\mathbb{R}^2$.
$\tilde S(x+2\pi n)$ and $\tilde S(x)$ project to the same point for every $n\in\mathbb{Z}^2$, so $On\in\mathbb{Z}^2$, the columns of $O$ are integer unit vectors, and $O$ is a signed permutation matrix.
$S$ preserves the area, so $S_{\#}P_{\mathcal{M}} = P_{\mathcal{M}}$ and the continuity of $\rho$ give $\rho\circ S = \rho$.
Each term $\cos(k\cdot x)$ or $\sin(k\cdot x)$ of $\rho$ becomes a term of frequency $O^{\top}k$ in $\rho\circ S$, since $k\cdot(Ox+b) = (O^{\top}k)\cdot x + k\cdot b$.
The functions $e^{\mathrm{i}k\cdot x}$, $k\in\mathbb{Z}^2$, are orthogonal, so $\rho\circ S = \rho$ forces $O^{\top}$ to map the frequencies $\pm(1,0)$, $\pm(1,1)$, $\pm(2,0)$ of $\rho$ onto themselves.
Their lengths $1$, $\sqrt2$ and $2$ differ, and $O^{\top}$ preserves lengths, so $O^{\top}(1,0) = \pm(1,0)$ and $O^{\top}(1,1) = \pm(1,1)$, which leaves $O = \pm I$.
If $O = I$, the terms of frequency $(1,0)$ in $\rho(x+b) = \rho(x)$ give $\cos(x_1+b_1) = \cos x_1$, so $b_1\in 2\pi\mathbb{Z}$, and those of frequency $(1,1)$ give $b_1+b_2\in 2\pi\mathbb{Z}$, so $S = \operatorname{id}$.
If $O = -I$, the terms of frequency $(1,0)$ in $\rho(b-x) = \rho(x)$ give $\cos(b_1-x_1) = \cos x_1$, so $b_1\in 2\pi\mathbb{Z}$.
The terms of frequency $(2,0)$ then require $\sin(2b_1-2x_1)$, which equals $-\sin 2x_1$, to equal $\sin 2x_1$ for every $x_1$, which fails.
Hence $S = \operatorname{id}$ and \ref{C1} holds.
\end{proof}

The map $\tau$ preserves the area, since $|\det B| = 1$, and $\rho\circ\tau = \rho$, so it preserves $P_{\mathcal{M}}$ and every LDD.
It does not preserve distances, since for small $s>0$ it maps the point $(s,0)$, at distance $s$ from $0 = \tau(0)$, to $(s,-2s)$, at distance $\sqrt5\,s$.

The sphere example of Section~\ref{sec:identifiability} also has collisions without symmetries.
There, three vMF components with distinct weights and concentrations break every symmetry of the sphere, which gives $P_{\mathrm{mix}}$ (Figure~\ref{fig:sphere-cases}, right).
The experiments of Section~\ref{sec:experiments} add the same mixture with equal weights and equal concentrations.
It still satisfies \ref{C1}, because its centres $c_1,c_2,c_3$ are linearly independent with distinct pairwise angles (Proposition~\ref{prop:vmf-rigidity-appendix}, (i) and (ii)(c)), but its LDDs collide on curves.
The mass of a cap under one component depends only on the angle between the centre of the cap and that of the component.
With equal weights and concentrations, $\Phi(x,\cdot)$ therefore depends only on the multiset of inner products $\{\langle x,c_m\rangle\}_{m=1}^3$.
Let $V$ be the matrix with rows $c_1,c_2,c_3$, and let $P\neq I$ be a permutation matrix.
The point $y = V^{-1}PVx$ has the inner products of $x$ in another order, so $\Phi(y,\cdot) = \Phi(x,\cdot)$ whenever $\|y\| = 1$.
These $x$ form the zero set on $S^2$ of the quadratic form $\|V^{-1}PVx\|^2 - \|x\|^2$.
The form is not zero, because an orthogonal $V^{-1}PV$ would have a transpose that permutes the centres and preserves their pairwise angles, which their distinct angles rule out for $P\neq I$.
Its zero set on $S^2$ is therefore null (Lemma~\ref{lem:analytic-null}).
Three of the colliding pairs are antipodal.
For $\{i,j,k\} = \{1,2,3\}$, the point $x_k\propto(c_i+c_j)\times c_k$ has inner products $a$, $-a$ and $0$ with $c_i$, $c_j$ and $c_k$, with $a\neq 0$ because the centres are linearly independent, and $-x_k$ has the same multiset.

\subsection{The margin}

Since LDDs are estimated from samples, Section~\ref{sec:ldd_sinkhorn} requires them to be well separated and measures their separation by the margin $m(t)$ of Definition~\ref{def:margin}.
The margin fixes a separation $t$, because nearby points have nearby LDDs, so over all pairs of distinct points the smallest distance between LDDs is zero, with or without collisions:

\begin{lemma}\label{lem:sup-continuity}
Under \ref{H1}, the map $x\mapsto\Phi(x,\cdot)$ is continuous in the supremum norm.
\end{lemma}

\begin{proof}
Under \ref{H1}, distance spheres are $P_{\mathcal{M}}$-null (Remark~\ref{rem:null-spheres}), so each $\Phi(x,\cdot)$ is continuous, and it equals $1$ from the diameter $R$ of $\mathcal{M}$ on.
Fix $x$ and $x'$ and let $u = d_{\mathcal{M}}(x,x')$.
By the triangle inequality, the closed balls of radius $r$ around $x$ and $x'$ differ only within the shell $\{z: |d_{\mathcal{M}}(x,z) - r|\le u\}$, so $|\Phi(x,r) - \Phi(x',r)|$ is at most the mass of this shell.
As $u\to 0$, the shells decrease to a distance sphere, which is null, so $\Phi(x',r)\to\Phi(x,r)$ for every $r$.
For uniformity, let $\epsilon>0$ and choose radii $0 = r_0 < \dots < r_L = R$ with $\Phi(x,r_{l+1}) - \Phi(x,r_l)\le\epsilon$.
For $x'$ close enough to $x$, $|\Phi(x',r_l) - \Phi(x,r_l)|\le\epsilon$ at every $r_l$.
For $r\in[r_l,r_{l+1}]$, monotonicity then gives $\Phi(x',r)\le\Phi(x',r_{l+1})\le\Phi(x,r)+2\epsilon$ and likewise $\Phi(x',r)\ge\Phi(x,r)-2\epsilon$, so $\|\Phi(x',\cdot) - \Phi(x,\cdot)\|_\infty\le 2\epsilon$.
\end{proof}

At a fixed separation, a positive margin rules out every collision, and by compactness the converse holds as well:

\begin{proposition}\label{prop:margin-form}
Under \ref{H1}, the collision set $\mathcal{C}$ is empty if and only if $m(t)>0$ for every $t\in(0,R]$.
\end{proposition}

\begin{proof}
If $m(t)>0$ for every $t$, two points $x\neq y$ are admissible in the infimum of $m(d_{\mathcal{M}}(x,y))$, so their LDDs differ.
Conversely, let $\mathcal{C}$ be empty and suppose $m(t) = 0$ for some $t$.
Choose pairs $(x_n,y_n)$ with $d_{\mathcal{M}}(x_n,y_n)\ge t$ and $\|\Phi(x_n,\cdot) - \Phi(y_n,\cdot)\|_\infty\to 0$.
By compactness, a subsequence converges to a pair $(x,y)$ with $d_{\mathcal{M}}(x,y)\ge t$, so $x\neq y$, and by Lemma~\ref{lem:sup-continuity} $\Phi(x,\cdot) = \Phi(y,\cdot)$, which contradicts $\mathcal{C} = \emptyset$.
\end{proof}

A positive margin at every separation is therefore stronger than \ref{C2}, which allows a null collision set.
One pair of colliding points at separation $s$ makes $m(t) = 0$ for every $t\le s$.
In the sphere example, the antipodal pairs of the equal-parameter mixture make its margin zero at every separation, and so do the colliding points of $P_{\mathrm{unif}}$ and $P_{\mathrm{vMF}}$, which lie up to $\pi$ apart.

The separation is also the resolution at which $A$ is recovered.
If the estimation error of the LDDs is small against $m(t)$, the LDD-LP puts little mass on pairs $(z,w)$ with $w$ at least $t$ from $A(z)$ (Proposition~\ref{prop:ldd-stability}).
Since $m$ is non-decreasing in $t$, a finer resolution comes with a smaller margin.

\subsection{The margin test}
\label{app:dkw}

From samples we cannot compute the margin, but we can test whether two query points at least $t$ apart share an LDD.
The test compares the smallest distance between their empirical LDDs with the accuracy that the DKW inequality guarantees for each of them:

\begin{lemma}\label{lem:dkw-test}
Let $s_1,\dots,s_N$ and $q_1,\dots,q_k$ be independent samples from $P_{\mathcal{M}}$, let $\hat\Phi_i(r) = \frac1N\#\{j: d_{\mathcal{M}}(q_i,s_j)\le r\}$ be the empirical LDD of $q_i$, and let $\varepsilon_N = \sqrt{\ln(2k/\delta)/2N}$.
For $t>0$, let $\hat m(t)$ be the smallest $\|\hat\Phi_i - \hat\Phi_j\|_\infty$ and $m_k(t)$ the smallest $\|\Phi(q_i,\cdot) - \Phi(q_j,\cdot)\|_\infty$, both over pairs with $d_{\mathcal{M}}(q_i,q_j)\ge t$.
With probability at least $1-\delta$, simultaneously for all $t$,
\emph{(i)} every pair with $d_{\mathcal{M}}(q_i,q_j)\ge t$ satisfies $\|\Phi(q_i,\cdot) - \Phi(q_j,\cdot)\|_\infty\ge\hat m(t) - 2\varepsilon_N$,
\emph{(ii)} $|\hat m(t) - m_k(t)|\le 2\varepsilon_N$, and
\emph{(iii)} $\hat m(t)\le 2\varepsilon_N$ implies $m(t)\le 4\varepsilon_N$.
\end{lemma}

\begin{proof}
Given $q_i$, $\hat\Phi_i$ is the empirical distribution function of the $N$ i.i.d.\ distances $d_{\mathcal{M}}(q_i,s_j)$, whose distribution function is $\Phi(q_i,\cdot)$.
The DKW inequality with Massart's constant (Theorem~\ref{thm:dkw}; \citealp{DKW, massart2}) and a union bound over $i$ give the event $E$ that $\|\hat\Phi_i - \Phi(q_i,\cdot)\|_\infty\le\varepsilon_N$ for all $i$, with probability at least $1-\delta$.
On $E$, the triangle inequality moves every pairwise distance between LDDs by at most $2\varepsilon_N$, which gives (i).
The minima $\hat m(t)$ and $m_k(t)$ run over the same pairs, so they also differ by at most $2\varepsilon_N$, and both equal $+\infty$ when no pair is $t$ apart, which gives (ii).
If $\hat m(t)\le 2\varepsilon_N$, the pair that attains it is at least $t$ apart and has LDDs at most $4\varepsilon_N$ apart, which gives (iii).
The event $E$ does not depend on $t$.
\end{proof}

By (i), the probability that $\hat m(t)>2\varepsilon_N$ while two query points at least $t$ apart share an LDD is at most $\delta$, so rejecting when $\hat\xi(t):= \hat m(t)/2\varepsilon_N > 1$ is a level-$\delta$ test.
By (ii), the test detects a margin $m$ among the query points once $4\varepsilon_N<m$, that is, with more than $8\ln(2k/\delta)/m^2$ reference points.

\begin{remark}[Estimation and coverage]
\label{rem:coverage}
Lemma~\ref{lem:dkw-test} controls the estimation of each LDD, but not how well the query points cover $\mathcal{M}$.
Their margin $m_k(t)$ is an infimum over fewer pairs than $m(t)$, so $m_k(t)\ge m(t)$, and the two meet only as the queries become dense.
Whether the test finds a collision therefore depends on how many points share an LDD.
Under $P_{\mathrm{unif}}$ and $P_{\mathrm{vMF}}$, every point shares its LDD with a continuum of others, random queries land near one of them, and the test does not reject.
In the equal-parameter mixture, the collisions above lie on curves and a random pair of queries rarely lies on one, so the test can reject although $m(t) = 0$, which two of the three seeds do at separations from $1.67$ on, where few query pairs remain, while the mean in Figure~\ref{fig:sphere}c stays below $1$.
A rejection thus bounds the LDD distances of the queried pairs only.
\end{remark}

In low dimension, the queries can cover $\mathcal{M}$, and the estimate then bounds $m(t)$ itself.
Under Equation~\ref{eq:ldd-lipschitz}, LDDs are Lipschitz in the point, $\|\Phi(x,\cdot) - \Phi(x',\cdot)\|_\infty\le K\,d_{\mathcal{M}}(x,x')$ (proof of Proposition~\ref{prop:ldd-stability}).
If every point of $\mathcal{M}$ lies within $\gamma$ of a query point, every pair at least $t$ apart lies within $\gamma$ of a query pair at least $t-2\gamma$ apart, so on $E$
$$m(t)\ \ge\ \hat m(t-2\gamma) - 2\varepsilon_N - 2K\gamma.$$
The number of query points such a cover needs grows exponentially with the dimension of $\mathcal{M}$.

The experiments evaluate every LDD on a grid of radii (Appendix~\ref{app:exp-sphere}).
A supremum over the grid is at most the full supremum, and the DKW bound is uniform in $r$, so everything above holds with the grid norm in place of $\|\cdot\|_\infty$, except that (iii) needs $4\varepsilon_N + Kh$ in place of $4\varepsilon_N$ for radii $h$ apart, since the difference of two LDDs is $2K$-Lipschitz in $r$.

%% file: sections/appendices/appendix_ngma.tex
\section{The NGMA procedure}
\label{app:ngma}

NGMA turns the two stages of Section~\ref{sec:recovery} into an estimator.
Three things separate it from the idealised procedure analysed there.
The latent distances are not given and have to be estimated from the decoders.
The coupling comes from a non-convex solver.
And the map is a neural network fitted without a single pair, so model selection has to work without pairs too.
This appendix gives the details of Section~\ref{sec:ngma} in the same order, and Algorithm~\ref{alg:ngma} summarises the procedure.

\textbf{The input} is computed from the decoders.
Following \citet{syrota2024decoder}, each model has an ensemble of $K$ decoders $f_{a,1},\dots,f_{a,K}$ on one encoder, and the metric tensor at $z$ is the expected pullback metric
\begin{equation}\label{eq:ensemble-metric}
G_a(z) = \frac1K\sum_{k=1}^{K} J_{f_{a,k}}(z)^{\top} J_{f_{a,k}}(z),
\end{equation}
which grows where the decoders disagree, that is, away from the data.
Geodesic distances require a boundary value problem for every pair, which is out of reach for $n^2$ pairs.
We instead measure each pair by the length of the straight segment between them under Equation~\ref{eq:ensemble-metric}, discretised into $T$ steps ($T=32$ for the VAEs and $T=10$ for the scVI models of Appendix~\ref{app:lung}),
\begin{equation}\label{eq:chord}
C^a_{ij} = \sum_{t=1}^{T} \Big(\frac1K\sum_{k=1}^{K} \big\|f_{a,k}(z_t) - f_{a,k}(z_{t-1})\big\|^2\Big)^{1/2},
\qquad z_t = \rz^a_i + \tfrac{t}{T}\big(\rz^a_j - \rz^a_i\big),
\end{equation}
and likewise for $C^b$.
Both the ensemble metric and the segment length approximate the pullback geometry of Section~\ref{sec:latentmmspaces}, and the results of Section~\ref{sec:finiterecovery} do not cover either approximation.
In Appendix~\ref{app:experiments}, pullback distances are these segment lengths, except where Appendix~\ref{app:geodesic-segments} compares them with geodesics.

\textbf{The coupling} is computed by entropic GW \citep{peyre2016gromov} on $J_\beta$, with $C^a$ and $C^b$ both divided by the mean of their two mean entries, which fixes the overall scale and keeps the relative scale of the two spaces, and with the cost $c$ of Equation~\ref{eq:sinkhorn_problem} computed from the same normalised distances.
Four stages lower $\beta$ through $1$, $0.8$, $0.5$ and $0$, and the entropy weight through $\varepsilon_0$, $0.1\varepsilon_0^{2/3}$, $0.01\varepsilon_0^{1/3}$ and $10^{-3}$, a geometric interpolation between the LDD-LP's weight $\varepsilon_0=0.3\max c$ and GW's schedule.
Each stage starts from the solution of the previous one, and the first from the independent coupling.
Each iteration replaces the coupling $\eta$ by the entropic optimal coupling for the cost $\big(2(1-\beta)L_\eta+\beta c\big)/(2-\beta)$, the derivative of $J_\beta$ (Appendix~\ref{app:fused}) divided by $2-\beta$, which for $\beta=0$ is the update of entropic GW \citep{peyre2016gromov}.
Each stage runs at most $500$ iterations, stopping when the coupling changes by less than $10^{-9}$ in relative terms, and each iteration solves its transport problem by log-domain Sinkhorn iterations to a change of $10^{-9}$, in double precision.
The result is a soft coupling $\hat\eta\in\mathbb{R}^{n\times n}$ with uniform marginals, and every later step reads only $\hat\eta$, the two samples and their metric tensors.
Every stage with $\beta<1$ has the minimiser $\eta^\star$ under \ref{C1} (Proposition~\ref{prop:fused-unique}), so the anneal changes the path but not the minimiser.
At the independent coupling, the linearised GW objective compares points only by their mean distances (Proposition~\ref{prop:fused-indep}), whereas the first stage compares their whole LDDs.
The anneal helps the solver avoid poor stationary points, but it does not guarantee the global one.

\textbf{Standardisation and targets.}
Before the maps are fitted, each cloud is centred and divided by its mean norm $s_a$, and the metric tensors are rescaled to $s_a^2\,G_a$ to match the new coordinates, in which the maps act.
Standardising removes the offset and scale between the two clouds from what the maps have to learn, and it puts every squared distance in the loss in the same units.
The target of $\rz^a_i$ is the point that $\hat\eta$ most likely pairs it with, $\tau^a_i:= \rz^b_{j^\star(i)}$ with $j^\star(i) = \arg\max_j \hat\eta_{ij}$, and symmetrically $\tau^b_j:= \rz^a_{i^\star(j)}$.
This replaces the barycentric projection of the coupling, with which it coincides at $\eta^\star$.

\textbf{The two maps} of Section~\ref{sec:ngma} are
\begin{equation}\label{eq:ngma-maps}
\phi(z) = Wz + v + h(z), \qquad \psi(w) = W'w + v' + h'(w),
\end{equation}
where $h$ and $h'$ are multilayer perceptrons with two hidden layers of width $128$ and GELU activations \citep{hendrycks2016gelu}, whose last layers are initialised at zero \citep{zhang2019fixup}.
The affine part starts at the best orthogonal map and translation under the coupling,
\begin{equation}\label{eq:procrustes}
(Q_0, v_0) = \arg\min_{Q\in O(q),\, v\in\mathbb{R}^q}\ \sum_{i,j}\hat\eta_{ij}\,\big\|Q \rz^a_i + v - \rz^b_j\big\|^2,
\end{equation}
which has a closed form through one singular value decomposition of the $\hat\eta$-weighted cross-covariance \citep{schonemann1966procrustes}.
So $W = Q_0$, $v = v_0$, $W' = Q_0^{\top}$ and $v' = -Q_0^{\top}v_0$ at the start, and all parameters are trained, with none shared between the two maps.
The coupling is typically right about regions and wrong about individual representations.
A rigid map can be fitted from regional information and still extrapolate to every representation, since it has too few parameters to follow the coupling's local mistakes, and everything after this step corrects it.

\textbf{The objective} is Equation~\ref{eq:ngma-objective}, with the average over the $n_a$ training points replaced by an average over a batch of $256$ of them, and likewise for $\ell_{ba}$.
Since $G_b$ is known only at the samples, the isometry term reads it at the point $\hat w$ of $\mathbf{z}^b$ nearest to $\phi(\rz^a_i)$, and no gradient passes through this choice.
The isometry term is the pointwise form of $A^{*}g_b = g_a$ (Proposition~\ref{prop:alignment}), and it is the only term that reads the geometry.
It is also small for maps that send representations to the wrong points with the right local metric, which the fit term penalises by keeping each map close to the coupling's regional structure.
The cycle term pushes the two maps to undo each other, as $A$ and $A^{-1}$ do.
Measure preservation, the remaining half of an mm-space isomorphism, enters through the uniform marginals of $\hat\eta$ and through the selection rule below, which counts one-to-one pairs.
Both maps are trained jointly with Adam \citep{kingma2015adam} at learning rate $10^{-3}$ for $3000$ steps.

\textbf{Training and selection without pairs.}
Before training, we hold out $n_0=500$ random points of $\mathbf{z}^a$ and every point of $\mathbf{z}^b$ whose largest entry of $\hat\eta$ falls on one of them, so that no held-out point enters the loss.
Let $\mathrm{mNN}(\phi;\mathbf{z}^a\to\mathbf{z}^b)$ be the fraction of held-out points $\rz^a_i$ for which $\phi(\rz^a_i)$ and some $\rz^b_j$ are each other's nearest neighbours, and define $\mathrm{mNN}(\psi;\mathbf{z}^b\to\mathbf{z}^a)$ likewise.
At the start and every $250$ steps, the pair is scored by
\begin{equation}\label{eq:mnn-score}
s(\phi,\psi) = \tfrac12\Big[\mathrm{mNN}\big(\phi;\ \mathbf{z}^a\to\mathbf{z}^b\big) + \mathrm{mNN}\big(\psi;\ \mathbf{z}^b\to\mathbf{z}^a\big)\Big],
\end{equation}
and the checkpoint with the largest score is kept, the rigid start included.
A measure-preserving map between two samples of one distribution induces a near-bijection between them, and $\mathrm{mNN}$ counts how much of it a map realises.
It reads no pair, and averaging over held-out points keeps it from rewarding a map that memorised its training points.
On disjoint samples no map reaches one, so $s$ only compares candidates.

\textbf{Refinement and ranking.}
The first fit learns from targets that are wrong for many individual representations, while the pairs that both maps agree on are a smaller and more reliable set.
As in unsupervised word translation \citep{conneau2018word, artetxe2018robust}, we refit on those pairs, the mutual nearest neighbours under the combined distance
\begin{equation}\label{eq:combined}
D(i,j) = \omega_a\,\frac{\|\phi(\rz^a_i) - \rz^b_j\|}{\operatorname{med}_{i',j'}\|\phi(\rz^a_{i'}) - \rz^b_{j'}\|}
+ \omega_b\,\frac{\|\psi(\rz^b_j) - \rz^a_i\|}{\operatorname{med}_{i',j'}\|\psi(\rz^b_{j'}) - \rz^a_{i'}\|}.
\end{equation}
Each weight is proportional to the squared fraction of points whose image under its map is a mutual nearest neighbour of a point of the other sample, and the two sum to one, so a map that pairs few points contributes little.
These mutual nearest neighbours become anchors.
For anchors on training points the targets are replaced, $\tau^a_i \leftarrow \rz^b_j$ and $\tau^b_j \leftarrow \rz^a_i$, half of every batch is drawn from the anchors, and both maps are trained for $1000$ steps.
We run up to three such rounds, stop early when the number of anchors grows by less than $2\%$, and keep the round with the largest held-out mutual-nearest-neighbour fraction under $D$, the state before the first round included.
The kept $\phi$ aligns any new representation of model $a$, $\psi$ aligns in the other direction, and $D$ ranks candidate pairs between two given samples, which is how test candidates are ranked.

\begin{algorithm}[t]
\caption{NGMA.
The input is two samples of representations and the decoder ensembles of both models.
No pair is read.}
\label{alg:ngma}
\begin{algorithmic}[1]
\State $C^a, C^b \gets$ segment lengths, Equation~\ref{eq:chord}; $G_a, G_b \gets$ Equation~\ref{eq:ensemble-metric} at every sample point
\State $\hat\eta \gets$ entropic GW on $J_\beta$, with $\beta$ lowered from $1$ to $0$ and the regularisation from $\varepsilon_0$ to $10^{-3}$
\State standardise both clouds; $\tau^a, \tau^b \gets$ the points at the row and column maxima of $\hat\eta$
\State hold out $n_0$ points of $\mathbf{z}^a$ and the points of $\mathbf{z}^b$ whose largest entry of $\hat\eta$ falls on them
\State $(Q_0, v_0) \gets$ Equation~\ref{eq:procrustes}; $\phi \gets$ the rigid map, $\psi \gets$ its inverse
\State train $(\phi,\psi)$ on Equation~\ref{eq:ngma-objective}; keep the checkpoint with the largest $s$, Equation~\ref{eq:mnn-score}
\For{up to three rounds}
  \State anchors $\gets$ mutual nearest neighbours under $D$, Equation~\ref{eq:combined}; replace their targets
  \State train $(\phi,\psi)$ on Equation~\ref{eq:ngma-objective} with half of each batch from the anchors
\EndFor
\State \Return the round with the largest held-out mutual-nearest-neighbour fraction under $D$
\end{algorithmic}
\end{algorithm}

\textbf{Relation to existing procedures.}
Procrustes initialisation and refinement on mutual nearest neighbours come from unsupervised word translation \citep{conneau2018word, artetxe2018robust}.
\citet{conneau2018word} also select models without pairs, by the mean similarity of the translations a map induces, where we count mutual pairs on held-out points.
Fitting both directions with a cycle term follows \citet{zhu2017unpaired}.
GENOT \citep{klein2024genot} also learns a map from GW couplings; other neural GW methods train the map with a Gromov--Monge gap \citep{sotiropoulou2024strongly, uscidda2024disentangled} or estimate the entropic GW coupling with neural networks \citep{wang2024neural}.
NGMA differs from them in what the map has to be, an isometry of the two pullback geometries, which is what the isometry term reads and what Section~\ref{sec:latentmmspaces} says the target is.

%% file: sections/appendices/appendix_experiments.tex
\section{Experimental details}
\label{app:experiments}

\subsection{Data on the sphere}
\label{app:exp-sphere}

\paragraph{Measures.}
The uniform measure, a vMF measure with concentration $\kappa=8$, and the mixture $P_{\mathrm{mix}}$ of Figure~\ref{fig:sphere-cases}: three vMF components with centres $c_1=(0.556,-0.148,0.818)$, $c_2=(0.168,0.594,0.787)$, $c_3=(0.451,0.804,-0.387)$, concentrations $8$, $12$, $16$ and weights $0.42$, $0.33$, $0.25$.
The centres are linearly independent and $49.5^\circ$, $75.6^\circ$ and $100.7^\circ$ apart, so $P_{\mathrm{mix}}$ satisfies \ref{C1} by Proposition~\ref{prop:vmf-rigidity-appendix}.
The fourth measure is $P_{\mathrm{mix}}$ with equal weights and concentration $12$.
It still satisfies \ref{C1}, because its centres are linearly independent with distinct pairwise angles, but its LDDs collide along curves that contain the three antipodal pairs $\pm x_k$ with $x_k \propto (c_i+c_j)\times c_k$ (Appendix~\ref{app:ldd-multi-scale}), so $m(t)=0$ at every $t$.
We check both numerically, computing LDDs by one-dimensional quadrature, since the mass of a cap under a vMF component depends only on the angle between the cap and the component's centre.
On the curves the LDDs of the equal-parameter mixture agree to within $10^{-10}$, while at the antipodal pairs those of $P_{\mathrm{mix}}$ differ by $0.03$ to $0.16$ in sup norm.

\paragraph{Samples and regimes.}
The alignment is the rotation $R$ by $60^\circ$ about $(0,1,1)/\sqrt2$.
Each side has $n=2000$ points, over ten seeds.
With independent samples, $\mathbf{z}^a$ and $\mathbf{z}^b$ are drawn independently from $P$ and $R_{\#}P$.
With permuted samples, $\rz^b_{\sigma(i)}=R\rz^a_i$ for a random permutation $\sigma$, and $C^b$ is $C^a$ with rows and columns permuted by $\sigma$.

\paragraph{Couplings and landmarks.}
The LDD-LP and GW are solved on $k$ landmarks per side, chosen greedily by D-optimal design on the LDDs \citep{Pukelsheim1993}.
LDDs are evaluated on $B=256$ radii spread evenly over $[0,\pi]$, the same on both sides.
Each step of the design adds the point whose LDD most increases $\log\det G_S$, the Gram matrix of the selected LDDs $S$.
Since $\det G_{S\cup\{i\}} = \det G_S\,\|h_i - P_S h_i\|^2$, with $h_i$ the LDD of point $i$ on the grid of radii and $P_S$ the projection onto the span of the selected LDDs, each step adds the LDD farthest from that span.
The LDDs span at most $B$ dimensions, so the design orders at most $B$ landmarks.
We stop it there, or earlier once the largest increment falls below $10^{-7}$, and continue by farthest-point sampling in the $L^2$ distance between LDDs.
Each $k$ takes a prefix of one ordering per side, so in the permutation regime the two sides choose the same points.
On learned representations, NGMA solves the coupling on all sample points.
The LDD-LP uses the cost of Equation~\ref{eq:sinkhorn_problem} on the geodesic distances from each landmark to all points of its sample, and we solve it exactly by network simplex \citep{flamary2021pot}.
In Figure~\ref{fig:sphere}b, NGMA's coupling is the annealing of Appendix~\ref{app:ngma} on the same landmarks, with the LDD-LP's cost as its linear term, and entropic GW anneals $\varepsilon$ over the same four stages without that term.
Both cap the inner Sinkhorn at $1000$ iterations per step.
We run them on the two mixtures only, and NGMA's coupling on independent samples only, since under $P_{\mathrm{unif}}$ and $P_{\mathrm{vMF}}$ no coupling identifies $R$.
All $960$ linear programs reached their optimum, and no stage of the $480$ GW solves reached its iteration limit.
In one GW coupling ($P_{\mathrm{mix}}$, $k=128$) the inner Sinkhorn stopped at its iteration limit with column marginals off by up to $10\%$.
Of the $240$ solves of NGMA's coupling, $91$ reached the limit of $500$ iterations in their last one or two stages, mostly at $k=16$--$128$; all keep their marginals to within $3\times10^{-7}$.

\paragraph{Map and error.}
We fit the rotation in closed form, as the $\hat R\in SO(3)$ minimising $\sum_{ij}\eta_{ij}\|\hat R \rz^a_i - \rz^b_j\|^2$ over the landmark coupling $\eta$, and report the angle of $\hat R^\top R$.

\paragraph{Margin test.}
Figure~\ref{fig:sphere}c runs the test of Section~\ref{sec:ldd_sinkhorn} with $N=512{,}000$ reference points, $k=2000$ query points, $B=2000$ radii and $\delta=0.05$, over three seeds, at $129$ separations in $[0,\pi]$.
It stops at the largest $t$ at which every measure keeps $1000$ query pairs.

\subsection{Learned representations: setup}
\label{app:exp-setup}

\paragraph{Models.}
The VAEs on MNIST and FashionMNIST have a four-layer convolutional encoder and residual upsampling decoders, with a $16$-dimensional latent space.
On CelebA ($64\times64$ images) the latent space has $32$ dimensions.
Each model has one encoder and an ensemble of eight decoders.
The loss is the negative ELBO with a KL weight of $16$, a Gaussian likelihood with one learned scale and free bits of $0.25$, and the KL weight is warmed up over $100$ epochs ($30$ on CelebA).
The prior is a standard Gaussian or a VampPrior \citep{pmlr-v84-tomczak18a} with $500$ pseudo-inputs ($200$ on CelebA).
We train with Adamax at learning rate $10^{-3}$, batch size $128$ ($256$ on CelebA), for $500$ epochs ($150$ on CelebA), and keep the checkpoint with the best validation ELBO.

\paragraph{Pairings and draws.}
Each dataset has four models per prior.
A same-prior setting aligns the six pairs among them, and the cross-prior setting four pairs of a VampPrior and a Gaussian model.
On CelebA there are three Gaussian, two VampPrior and two cross-prior pairings.
NGMA, Rosetta and mini-vec2vec run five draws per pairing, the other methods at least two.

\paragraph{Samples.}
Both samples come from the validation split ($6000$ images on MNIST and FashionMNIST, $15{,}000$ on CelebA).
Disjoint samples split it in two halves of $n=3000$ ($7500$ on CelebA).
The overlap sweep of Figure~\ref{fig:coupling-overlap-all} shares $0$, $5$, $10$, $15$, $33$, $66$ or $100\%$ of the images, with two draws for every MNIST and FashionMNIST pairing.

\paragraph{Latent geometry.}
The metric of a model is the ensemble pullback metric of Equation~\ref{eq:ensemble-metric} with $K=8$ decoders and exact Jacobians in pixel space \citep{syrota2024decoder}, and latent distances are the segment lengths of Equation~\ref{eq:chord} with $T=32$.

\paragraph{Couplings.}
NGMA's coupling is $\hat\eta$, computed as in Appendix~\ref{app:ngma}.
The LDD-LP coupling of Figure~\ref{fig:coupling-overlap-all} uses the cost of Equation~\ref{eq:sinkhorn_problem} on the empirical LDDs, divided by its maximum, with Sinkhorn iterations at entropy weight $3\times10^{-4}$.

\paragraph{Evaluation.}
Each method maps the test representations of one model onto the other, and we score retrieval among all test points, $10{,}000$ on MNIST and FashionMNIST and $19{,}962$ on CelebA.
R@1 is the fraction of queries whose true partner is the nearest candidate, with ties averaged.
NGMA ranks the candidates of a query by the combined distance of Equation~\ref{eq:combined}, which weights both maps by the squared fraction of mutual nearest neighbours each one reaches on the samples it was fitted to, so the ranking reads no test pairs.
We score every method by this rule.
For an orthogonal map (the rigid map, Rosetta, mini-vec2vec and the final rotation of LFM), the inverse preserves distances, and RelReps and RelGeo compare both sides in one shared space.
In both cases the $b\to a$ distances are the transpose of the $a\to b$ ones, so the rule gives their $a\to b$ ranking.
GENOT-Q's flow has no inverse, and we score it $a\to b$.
The NGMA ceiling reports the mean R@1 of its two directions.
CSLS@1, R@1 after cross-domain similarity local scaling \citep{conneau2018word}, uses $10$ neighbours on each side and the negative Euclidean distance as similarity.
For the baselines on CelebA, it is computed on a random subsample of $10{,}000$ test points.
Means are over model pairings of each pairing's mean, and the $\pm$ is the s.d.\ over all draws.
A draw with R@1 below $0.3$ counts as failed.

\paragraph{Baselines.}
The two ceilings are fitted on $3000$ true pairs ($7500$ on CelebA) from the validation split.
The rigid map is an orthogonal map with a translation, fitted to the pairs by Procrustes \citep{schonemann1966procrustes}.
The NGMA ceiling is NGMA's map class trained by least squares on the pairs, without NGMA's other loss terms.
RelReps \citep{moschella2023relative} represents each point by its cosine similarities to class-centroid anchors, and RelGeo \citep{yu2024connecting} by its distances to them, measured as straight-segment lengths under the decoder ensemble.
On CelebA each of the $40$ attributes gives an anchor at the mean representation of the images that have it.
RelReps-Centroid also adds an anchor for the images that lack it.
Latent functional maps (LFM) \citep{fumero2024latent} build a $100$-nearest-neighbour graph on each sample, fit a functional map between their first $50$ Laplacian eigenvectors from heat-kernel signatures, class labels or $50$ anchor pairs, and end with a rotation fitted on the matched points.
Rosetta \citep{rosetta} and mini-vec2vec \citep{minivec2vec} run with their authors' code and defaults on centred, unit-normalised representations.
GENOT-Q \citep{klein2024genot} is our implementation: entropic GW with $\varepsilon=10^{-2}$ on the full samples, with the Euclidean or the pullback cost, followed by a conditional flow that maps each test point to the mean of eight samples.
We do not report vec2vec \citep{jha2025harnessing}.
Its run and checkpoint were selected on the test pairs, and it stayed near chance in every setting.

\subsection{Learned representations: results}
\label{app:exp-full}

\paragraph{All methods.}
Tables~\ref{tab:full-mnist}--\ref{tab:full-celeba} list every method and NGMA variant behind Table~\ref{tab:alignment}, with CSLS@1 beside R@1.
Figure~\ref{fig:celeba-coupling} compares NGMA's coupling with entropic GW alone on every CelebA draw.

\input{tables/full-results/mnist_results}
\input{tables/full-results/fmnist_results}
\input{tables/full-results/celeba_results}
\input{figures/celeba_coupling}

\paragraph{Equivalence and isometry.}
\label{app:exp-equivalence}
The two models of a cross-prior pair are far from equivalent (Table~\ref{tab:equivalence}), yet their pullback distances agree under the true correspondence, while their Euclidean distances do not (Table~\ref{tab:invariance}).
The table compares the distance between two images in one model with their distance in the other, and counts how often an image has the same nearest neighbour in both models.
This is the situation of Proposition~\ref{prop:alignment}, which needs a shared data manifold, not equivalent models.

\input{tables/equivalence}
\input{tables/invariance}

\paragraph{Shared images.}
\label{app:exp-overlap}
Figure~\ref{fig:coupling-overlap-all} shows NGMA's R@1 against the share of images the two samples have in common, in every MNIST and FashionMNIST setting.
With all images shared, NGMA reaches $0.990$--$0.999$, above the rigid map fitted on the true pairs in every setting.
At every share, NGMA has the highest R@1 of the five methods in every setting (Table~\ref{tab:shared}), including the label-informed RelReps and RelGeo.
Rosetta and mini-vec2vec centre and normalise the codes before fitting their rotation, so they can exceed that rigid map; Rosetta does in MNIST cross-prior at every share ($0.571$--$0.633$ against $0.512$).

\input{figures/coupling_overlap_all}
\input{tables/shared_baselines}

\subsection{Straight segments against geodesics}
\label{app:geodesic-segments}

NGMA measures latent distances by the lengths of straight segments (Equation~\ref{eq:chord}) instead of geodesic distances, which need an optimisation for every pair.
Optimising a path for each of the $n^2$ pairs of a sample is out of reach, so for this comparison we compute geodesics only between landmarks, $500$ points per model chosen by the D-optimal design of Appendix~\ref{app:exp-sphere}, which picks points whose LDDs differ as much as possible.
Each geodesic starts from the straight segment between its endpoints, and we optimise its path under Equation~\ref{eq:ensemble-metric}.
Within each model of a Gaussian and a VampPrior pair of FashionMNIST models, the segments are longer than the geodesics by a median of $11.5$--$15.6\%$ and keep their order, with Spearman correlations of $0.974$--$0.984$.

Comparing the two models of a pair needs distances between the same images in both.
Both models of a cross-prior pair encode the same sample of images, but each chooses its landmarks from its own LDDs, so the two choices differ and share only $67$ images.
On these images and under the true correspondence, the distances of the two models correlate at $0.994$ with geodesics and at $0.981$ with segments.

\subsection{The CelebA failures are mirror alignments}
\label{app:mirror}

Seven CelebA draws fail with NGMA's coupling and ten with GW alone (Table~\ref{tab:alignment}, Figure~\ref{fig:celeba-coupling}).
A left-right flip permutes pixels, so it preserves distances in the observation space.
Since it maps faces to faces, its restriction to the data manifold preserves $d_{\mathcal{M}}$, and since a face and its mirror image are about equally likely, it nearly preserves $P_{\mathcal{M}}$.
The flip is then nearly a symmetry of the data mm-space, $f_b^{-1}\circ\mathrm{flip}\circ f_a$ is nearly an isomorphism of the latent mm-spaces, and \ref{C1} nearly fails.

We test this directly.
Every test image $x$ is flipped into $x'$, and both models encode the flipped images.
If a fitted map follows the mirror alignment, it sends the representation of $x$ in one model close to the representation of $x'$ in the other.
True R@1 scores each map against the representations of the unflipped images, and mirror R@1 against those of the flipped images with $x'$ as the partner, in each direction separately.
The self-mirror rate of a model is the fraction of images whose own mirror is the nearest flipped representation, and it measures how far the model keeps a face and its mirror apart.
We score every reported CelebA draw with both couplings, and as controls the Gaussian pairing $ab$ of MNIST, whose digits are not mirror-symmetric, and of FashionMNIST, whose garments often are.

On every failed draw, with either coupling and in either direction, mirror R@1 is $0.870$--$0.993$ and true R@1 at most $0.113$, and on every successful draw the reverse holds (Table~\ref{tab:mirror}).
The models keep a face and its mirror apart, since a face's own mirror is the nearest flipped representation for only $10$--$12\%$ of faces.
A high mirror R@1 therefore means that the map sends faces to their mirrors, not that the models confuse the two.
On MNIST, mirror R@1 stays below $0.01$, and on FashionMNIST it lies in between.
The mirror alignment also explains why the objectives cannot tell failed draws from successful ones.
The flip preserves distances and nearly preserves LDDs, so GW and the LDD-LP score the mirror alignment almost as well as $A$, and on the GW couplings the GW objective is $0.0103$--$0.0108$ on failed draws against $0.0104$--$0.0106$ on successful ones.

\input{tables/mirror_test}

\subsection{scVI models of lung cells}
\label{app:lung}

\paragraph{Data.}
We use the $62{,}954$ cells of the Tabula Sapiens lung measured with 10x 3$'$ chemistry \citep{tabulasapiens2022}, from four donors with $34$ annotated cell types, on the $3000$ most variable genes, as raw counts.
The cells are split at random into $10{,}000$ held-out evaluation cells, a pool of $9000$ cells from which the samples are drawn, and two disjoint training halves, each with nested subsets of $4900$, $9800$ and $19{,}600$ cells.
Every donor appears in every part.

\paragraph{Models and geometry.}
Each model is scVI \citep{scVI} with the recipe of the published Tabula Sapiens lung model: a $20$-dimensional latent space, a negative-binomial likelihood with one dispersion $\theta_g$ per gene, and $400$ epochs.
We fit one model per training subset and seed.
Models on different halves see different cells.
We keep each encoder and add seven decoders, trained on the model's own posterior representations with its likelihood.
The data metric is the Fisher metric of the negative-binomial mean, which is Euclidean in the coordinates $y_g = 2\sqrt{\theta_g}\operatorname{asinh}\big(\sqrt{l\rho_g/\theta_g}\big)$ at the median library size $l$, and latent distances are the segment lengths of Equation~\ref{eq:chord} with $T=10$ under the eight decoders.

\paragraph{Pairs, samples and evaluation.}
For each training size we align two pairs of models trained on different cells and two trained on the same cells, on disjoint samples of $3000$ cells per side and three draws per pair, with the settings of Appendix~\ref{app:ngma}.
Both models encode the same $10{,}000$ evaluation cells, which gives the true pairs.
We report R@1 among these cells (chance $10^{-4}$) and the cell-type accuracy of the nearest cell, both scored by the rule of Appendix~\ref{app:exp-setup}.

\paragraph{Results.}
NGMA is above every baseline on all $36$ samples, by $0.109$--$0.298$ in R@1, and above the rigid map fitted on the true pairs on all of them (Table~\ref{tab:lung_baselines}).
With Euclidean geometry, R@1 stays below $0.01$ on $10$ of the $18$ draws between models trained on different cells, and with the pullback metric on none.

Single cells are only partly determined by these models.
The map fitted on the true pairs reaches $0.36$--$0.69$, and NGMA $59$--$72\%$ of it.

\paragraph{Limits.}
Both models read the same $3000$ genes, so any cell can be encoded by both and pairs are not scarce.
The experiment tests whether the two models alone identify the alignment, not a use in biology.
The models have no batch covariate, so their representations carry donor structure, which both sides share under the random split.

\input{tables/full-results/scvi_lung}

\subsection{Compute}

\label{app:compute}
Every GPU job ran on one graphics compute die (GCD), with 64\,GB of device memory and 40\,GB of host memory, and
Table~\ref{tab:compute} gives the time of each stage for the image experiments.
The pullback distances take nearly all of it.
They decode the segment between every pair of points under eight decoders, so their cost grows as $n^2$; at CelebA's $n=15000$ they take about 810 GPU-hours per model.
The sphere experiments run on a CPU.

\begin{table}[h]
\centering
\caption{Compute per stage, in hours on one GPU: one GCD of an AMD Instinct MI250X, which has two
and which the software sees as two GPUs. Distances are pullback distances under the 8-decoder
ensemble between all $n$ images a model encodes ($n=6000$ for MNIST and FashionMNIST, $n=15000$ for
CelebA, of which each draw uses 7500); jobs split by rows compute each pair twice, the upper end of
the MNIST/FashionMNIST range. The whole project, development included, used about 50\,000 GPU-hours.}
\label{tab:compute}
\begin{tabular}{lrr}
\toprule
Stage & MNIST, FashionMNIST & CelebA \\
\midrule
VAE training, per model           & 3          & 16--25 \\
Pullback distances, per model     & 9--18      & 810 \\
Coupling, per draw                & 0.02--0.2  & 0.05--2.3 \\
Map fit and evaluation, per draw  & 0.1--0.25  & 0.35--0.75 \\
\midrule
One pair of models, one draw      & 25--43     & 1\,700 \\
\bottomrule
\end{tabular}
\end{table}

%% file: tables/full-results/mnist_results.tex
\begin{table}[t]
\centering
\caption{Instance retrieval on MNIST from disjoint samples ($n=3{,}000$ per side), every method, scored as in Table~\ref{tab:alignment}. Mean $\pm$ s.d.\ is the mean over model pairings (six for Gaussian and VampPrior, four for cross-prior) of each pairing's mean, with the s.d.\ over all draws. NGMA, Rosetta and mini-vec2vec use five draws per pairing, the other rows at least two. CSLS@1 is top-1 retrieval under the hubness correction of \citet{conneau2018word}. RelReps uses class-centroid anchors, so RelReps-Centroid is the same computation as RelReps $a\to b$ here and is not repeated. NGMA is ahead of every baseline not given the true pairs in all three columns: 0.967 against 0.941 (Rosetta), 0.989 against 0.930 (Rosetta), 0.955 against 0.571 (Rosetta). With the cross-prior pair, NGMA from GW on the Euclidean cost reaches 0.146 and from the LDD-LP coupling 0.894.}
\label{tab:full-mnist}
\vspace{4pt}
\scriptsize
\resizebox{\textwidth}{!}{%
\begin{tabular}{lcccccc}
\toprule
 & \multicolumn{2}{c}{Gaussian} & \multicolumn{2}{c}{VampPrior} & \multicolumn{2}{c}{cross-prior} \\
\cmidrule(lr){2-3} \cmidrule(lr){4-5} \cmidrule(lr){6-7}
Method & R@1 & CSLS@1 & R@1 & CSLS@1 & R@1 & CSLS@1 \\
\midrule
\multicolumn{7}{l}{\emph{Given the true pairs}} \\
NGMA, fitted on the true pairs & 0.998$\pm$0.001 & 0.999$\pm$0.000 & 1.000$\pm$0.000 & 1.000$\pm$0.000 & 0.999$\pm$0.000 & 1.000$\pm$0.000 \\
Rigid map, fitted on the true pairs & 0.955$\pm$0.021 & 0.975$\pm$0.014 & 0.936$\pm$0.008 & 0.954$\pm$0.012 & 0.512$\pm$0.021 & 0.441$\pm$0.032 \\
\midrule
\multicolumn{7}{l}{\emph{Label-informed (class labels or anchors)}} \\
RelReps & 0.525$\pm$0.069 & 0.585$\pm$0.069 & 0.297$\pm$0.082 & 0.338$\pm$0.096 & 0.004$\pm$0.001 & 0.007$\pm$0.001 \\
RelGeo & 0.512$\pm$0.124 & 0.564$\pm$0.131 & 0.552$\pm$0.235 & 0.603$\pm$0.233 & 0.319$\pm$0.152 & 0.370$\pm$0.169 \\
LFM, class labels & 0.024$\pm$0.011 & 0.028$\pm$0.013 & 0.100$\pm$0.037 & 0.156$\pm$0.054 & 0.004$\pm$0.003 & 0.007$\pm$0.003 \\
LFM, 50 paired anchors & 0.040$\pm$0.026 & 0.050$\pm$0.035 & 0.064$\pm$0.082 & 0.100$\pm$0.122 & 0.004$\pm$0.002 & 0.005$\pm$0.004 \\
\midrule
\multicolumn{7}{l}{\emph{Given no pairs}} \\
LFM, unsupervised & 0.000$\pm$0.000 & 0.000$\pm$0.000 & 0.004$\pm$0.005 & 0.006$\pm$0.008 & 0.000$\pm$0.000 & 0.000$\pm$0.000 \\
Rosetta & 0.941$\pm$0.028 & 0.962$\pm$0.021 & 0.930$\pm$0.012 & 0.952$\pm$0.015 & 0.571$\pm$0.021 & 0.793$\pm$0.013 \\
mini-vec2vec & 0.849$\pm$0.193 & 0.874$\pm$0.186 & 0.922$\pm$0.011 & 0.945$\pm$0.013 & 0.393$\pm$0.180 & 0.586$\pm$0.248 \\
GENOT-Q, Euclidean cost, $a\to b$ & 0.002$\pm$0.003 & -- & 0.076$\pm$0.005 & -- & 0.001$\pm$0.001 & -- \\
GENOT-Q, pullback cost, $a\to b$ & 0.033$\pm$0.004 & -- & 0.088$\pm$0.006 & -- & 0.027$\pm$0.002 & -- \\
\textbf{NGMA} & 0.967$\pm$0.021 & 0.986$\pm$0.012 & 0.989$\pm$0.012 & 0.994$\pm$0.008 & 0.955$\pm$0.008 & 0.988$\pm$0.005 \\
NGMA, GW coupling & 0.963$\pm$0.025 & 0.984$\pm$0.015 & 0.989$\pm$0.012 & 0.994$\pm$0.008 & 0.955$\pm$0.006 & 0.988$\pm$0.004 \\
NGMA, LDD-LP coupling & 0.741$\pm$0.155 & 0.829$\pm$0.150 & 0.981$\pm$0.012 & 0.991$\pm$0.008 & 0.894$\pm$0.019 & 0.967$\pm$0.009 \\
NGMA, GW on the Euclidean cost & 0.972$\pm$0.023 & 0.987$\pm$0.015 & 0.984$\pm$0.015 & 0.991$\pm$0.012 & 0.146$\pm$0.215 & 0.179$\pm$0.247 \\
NGMA, fully Euclidean & 0.949$\pm$0.024 & 0.972$\pm$0.016 & 0.954$\pm$0.011 & 0.972$\pm$0.010 & 0.026$\pm$0.034 & 0.047$\pm$0.061 \\
\bottomrule
\end{tabular}
}
\end{table}

%% file: tables/full-results/fmnist_results.tex
\begin{table}[t]
\centering
\caption{Instance retrieval on FashionMNIST from disjoint samples ($n=3{,}000$ per side), every method, scored as in Table~\ref{tab:alignment}. Mean $\pm$ s.d.\ is the mean over model pairings (six for Gaussian and VampPrior, four for cross-prior) of each pairing's mean, with the s.d.\ over all draws. NGMA, Rosetta and mini-vec2vec use five draws per pairing, the other rows at least two. CSLS@1 is top-1 retrieval under the hubness correction of \citet{conneau2018word}. RelReps uses class-centroid anchors, so RelReps-Centroid is the same computation as RelReps $a\to b$ here and is not repeated. NGMA is ahead of every baseline not given the true pairs in all three columns: 0.719 against 0.710 (Rosetta), 0.979 against 0.656 (Rosetta), 0.871 against 0.098 (RelGeo). With the cross-prior pair, NGMA from GW on the Euclidean cost reaches 0.643 and from the LDD-LP coupling 0.614.}
\label{tab:full-fmnist}
\vspace{4pt}
\scriptsize
\resizebox{\textwidth}{!}{%
\begin{tabular}{lcccccc}
\toprule
 & \multicolumn{2}{c}{Gaussian} & \multicolumn{2}{c}{VampPrior} & \multicolumn{2}{c}{cross-prior} \\
\cmidrule(lr){2-3} \cmidrule(lr){4-5} \cmidrule(lr){6-7}
Method & R@1 & CSLS@1 & R@1 & CSLS@1 & R@1 & CSLS@1 \\
\midrule
\multicolumn{7}{l}{\emph{Given the true pairs}} \\
NGMA, fitted on the true pairs & 0.997$\pm$0.001 & 0.999$\pm$0.000 & 1.000$\pm$0.000 & 1.000$\pm$0.000 & 0.998$\pm$0.001 & 0.999$\pm$0.000 \\
Rigid map, fitted on the true pairs & 0.782$\pm$0.032 & 0.846$\pm$0.031 & 0.747$\pm$0.029 & 0.847$\pm$0.024 & 0.071$\pm$0.004 & 0.062$\pm$0.005 \\
\midrule
\multicolumn{7}{l}{\emph{Label-informed (class labels or anchors)}} \\
RelReps & 0.151$\pm$0.053 & 0.180$\pm$0.059 & 0.062$\pm$0.013 & 0.073$\pm$0.016 & 0.002$\pm$0.001 & 0.002$\pm$0.001 \\
RelGeo & 0.159$\pm$0.134 & 0.182$\pm$0.151 & 0.355$\pm$0.131 & 0.394$\pm$0.146 & 0.098$\pm$0.033 & 0.123$\pm$0.040 \\
LFM, class labels & 0.011$\pm$0.004 & 0.014$\pm$0.006 & 0.020$\pm$0.010 & 0.033$\pm$0.016 & 0.001$\pm$0.001 & 0.002$\pm$0.001 \\
LFM, 50 paired anchors & 0.060$\pm$0.034 & 0.080$\pm$0.046 & 0.029$\pm$0.027 & 0.051$\pm$0.041 & 0.001$\pm$0.001 & 0.002$\pm$0.001 \\
\midrule
\multicolumn{7}{l}{\emph{Given no pairs}} \\
LFM, unsupervised & 0.001$\pm$0.000 & 0.001$\pm$0.001 & 0.000$\pm$0.000 & 0.001$\pm$0.001 & 0.000$\pm$0.000 & 0.000$\pm$0.000 \\
Rosetta & 0.710$\pm$0.040 & 0.761$\pm$0.037 & 0.656$\pm$0.045 & 0.762$\pm$0.044 & 0.065$\pm$0.019 & 0.178$\pm$0.047 \\
mini-vec2vec & 0.504$\pm$0.274 & 0.547$\pm$0.292 & 0.624$\pm$0.030 & 0.738$\pm$0.028 & 0.018$\pm$0.010 & 0.044$\pm$0.026 \\
GENOT-Q, Euclidean cost, $a\to b$ & 0.007$\pm$0.005 & -- & 0.021$\pm$0.012 & -- & 0.002$\pm$0.002 & -- \\
GENOT-Q, pullback cost, $a\to b$ & 0.009$\pm$0.001 & -- & 0.030$\pm$0.002 & -- & 0.006$\pm$0.001 & -- \\
\textbf{NGMA} & 0.719$\pm$0.112 & 0.812$\pm$0.107 & 0.979$\pm$0.027 & 0.989$\pm$0.017 & 0.871$\pm$0.049 & 0.965$\pm$0.017 \\
NGMA, GW coupling & 0.733$\pm$0.107 & 0.824$\pm$0.102 & 0.982$\pm$0.023 & 0.990$\pm$0.015 & 0.872$\pm$0.063 & 0.964$\pm$0.024 \\
NGMA, LDD-LP coupling & 0.268$\pm$0.097 & 0.340$\pm$0.117 & 0.953$\pm$0.065 & 0.971$\pm$0.054 & 0.614$\pm$0.076 & 0.797$\pm$0.072 \\
NGMA, GW on the Euclidean cost & 0.561$\pm$0.317 & 0.622$\pm$0.330 & 0.917$\pm$0.198 & 0.929$\pm$0.189 & 0.643$\pm$0.329 & 0.738$\pm$0.377 \\
NGMA, fully Euclidean & 0.536$\pm$0.284 & 0.593$\pm$0.297 & 0.815$\pm$0.217 & 0.862$\pm$0.212 & 0.051$\pm$0.029 & 0.078$\pm$0.045 \\
\bottomrule
\end{tabular}
}
\end{table}

%% file: tables/full-results/celeba_results.tex
\begin{table}[t]
\centering
\caption{Instance retrieval on CelebA from disjoint samples ($n=7{,}500$ per side), every method, scored as in Table~\ref{tab:alignment}. Mean $\pm$ s.d.\ is the mean over model pairings (three for Gaussian, two for VampPrior and two for cross-prior) of each pairing's mean, with the s.d.\ over all draws. NGMA, Rosetta and mini-vec2vec use five draws per pairing, the other rows at least two. CSLS@1 is top-1 retrieval under the hubness correction of \citet{conneau2018word}. NGMA is ahead of every baseline not given the true pairs in all three columns: 0.804 against 0.567 (Rosetta), 0.818 against 0.616 (mini-vec2vec), 0.818 against 0.757 (Rosetta). With the cross-prior pair, NGMA from GW on the Euclidean cost reaches 0.006 and from the LDD-LP coupling 0.643. Where a draw of NGMA has R@1 below $0.3$, the parentheses count the draws that reach it. From NGMA's coupling, 28 of the 35 draws reach it; from GW alone, 25.}
\label{tab:full-celeba}
\vspace{4pt}
\scriptsize
\resizebox{\textwidth}{!}{%
\begin{tabular}{lcccccc}
\toprule
 & \multicolumn{2}{c}{Gaussian} & \multicolumn{2}{c}{VampPrior} & \multicolumn{2}{c}{cross-prior} \\
\cmidrule(lr){2-3} \cmidrule(lr){4-5} \cmidrule(lr){6-7}
Method & R@1 & CSLS@1 & R@1 & CSLS@1 & R@1 & CSLS@1 \\
\midrule
\multicolumn{7}{l}{\emph{Given the true pairs}} \\
NGMA, fitted on the true pairs & 0.995$\pm$0.000 & 0.995$\pm$0.000 & 0.996$\pm$0.000 & 0.997$\pm$0.000 & 0.996$\pm$0.000 & 0.996$\pm$0.000 \\
Rigid map, fitted on the true pairs & 0.993$\pm$0.000 & 0.997$\pm$0.000 & 0.995$\pm$0.000 & 0.998$\pm$0.000 & 0.990$\pm$0.000 & 0.996$\pm$0.000 \\
\midrule
\multicolumn{7}{l}{\emph{Label-informed (class labels or anchors)}} \\
RelReps-Centroid & 0.270$\pm$0.028 & 0.352$\pm$0.031 & 0.249$\pm$0.065 & 0.305$\pm$0.077 & 0.000$\pm$0.000 & 0.000$\pm$0.000 \\
RelReps & 0.445$\pm$0.033 & 0.575$\pm$0.030 & 0.282$\pm$0.070 & 0.352$\pm$0.084 & 0.000$\pm$0.000 & 0.000$\pm$0.000 \\
RelGeo & 0.340$\pm$0.118 & 0.424$\pm$0.150 & 0.387$\pm$0.139 & 0.479$\pm$0.166 & 0.220$\pm$0.120 & 0.278$\pm$0.151 \\
LFM, class labels & 0.060$\pm$0.015 & 0.131$\pm$0.033 & 0.024$\pm$0.008 & 0.068$\pm$0.019 & 0.003$\pm$0.002 & 0.012$\pm$0.005 \\
LFM, 50 paired anchors & 0.042$\pm$0.034 & 0.100$\pm$0.075 & 0.108$\pm$0.092 & 0.247$\pm$0.188 & 0.006$\pm$0.004 & 0.023$\pm$0.013 \\
\midrule
\multicolumn{7}{l}{\emph{Given no pairs}} \\
LFM, unsupervised & 0.000$\pm$0.000 & 0.001$\pm$0.001 & 0.000$\pm$0.000 & 0.001$\pm$0.000 & 0.000$\pm$0.000 & 0.000$\pm$0.000 \\
Rosetta & 0.567$\pm$0.453 & 0.586$\pm$0.444 & 0.365$\pm$0.434 & 0.388$\pm$0.421 & 0.757$\pm$0.372 & 0.795$\pm$0.367 \\
mini-vec2vec & 0.251$\pm$0.353 & 0.278$\pm$0.360 & 0.616$\pm$0.415 & 0.648$\pm$0.400 & 0.439$\pm$0.474 & 0.475$\pm$0.482 \\
GENOT-Q, Euclidean cost, $a\to b$ & 0.000$\pm$0.000 & -- & 0.003$\pm$0.002 & -- & 0.000$\pm$0.000 & -- \\
GENOT-Q, pullback cost, $a\to b$ & 0.001$\pm$0.000 & -- & 0.004$\pm$0.002 & -- & 0.001$\pm$0.001 & -- \\
\textbf{NGMA} & 0.804$\pm$0.375\,{\scriptsize(12/15)} & 0.814$\pm$0.367 & 0.818$\pm$0.372\,{\scriptsize(8/10)} & 0.823$\pm$0.363 & 0.818$\pm$0.372\,{\scriptsize(8/10)} & 0.823$\pm$0.363 \\
NGMA, GW coupling & 0.684$\pm$0.440\,{\scriptsize(10/15)} & 0.697$\pm$0.430 & 0.818$\pm$0.372\,{\scriptsize(8/10)} & 0.823$\pm$0.362 & 0.729$\pm$0.427\,{\scriptsize(7/10)} & 0.736$\pm$0.417 \\
NGMA, LDD-LP coupling & 0.467$\pm$0.434 & 0.515$\pm$0.424 & 0.733$\pm$0.328 & 0.759$\pm$0.311 & 0.643$\pm$0.378 & 0.670$\pm$0.362 \\
NGMA, GW on the Euclidean cost & 0.424$\pm$0.446 & 0.458$\pm$0.442 & 0.464$\pm$0.457 & 0.478$\pm$0.446 & 0.006$\pm$0.018 & 0.010$\pm$0.028 \\
NGMA, fully Euclidean & 0.387$\pm$0.430 & 0.423$\pm$0.420 & 0.451$\pm$0.464 & 0.468$\pm$0.452 & 0.000$\pm$0.001 & 0.001$\pm$0.002 \\
\bottomrule
\end{tabular}
}
\end{table}

%% file: figures/celeba_coupling.tex
\begin{figure}[t]
\centering
\includegraphics[width=0.48\textwidth]{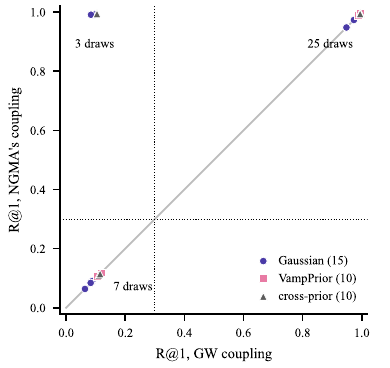}
\caption{NGMA on CelebA, every draw of every model pairing (35 draws): R@1 from NGMA's coupling against R@1 from entropic GW alone, on the same samples with the same map. Dotted lines mark R@1 $0.3$. 28 draws reach it with NGMA's coupling and 25 with GW; the coupling recovers 3 draws on which GW fails and loses 0, and no draw is lower by more than 0.004.}
\label{fig:celeba-coupling}
\end{figure}

%% file: tables/equivalence.tex
\begin{table}[t]
\centering
\caption{Likelihood and generated distributions of the two models of every pair: the difference of their importance-weighted (IWAE) bounds on $2000$ test images ($K=1000$ importance samples), and the squared maximum mean discrepancy (MMD$^2$) between their generated images, and of each against the data, with the permutation $p$-value of the first. Ranges over pairings; in the cross-prior settings $a$ is the VampPrior model. The cross-prior pairs differ by 10.2--25.2 nats, the same-prior pairs by at most 4.3. The two generators of a cross-prior pair are further apart than the VampPrior model is from the data on 7 of 8 pairs, and than the Gaussian model is from the data on 0. The MMD test also tells 18 of 24 same-prior pairs apart ($p<0.05$; 0.005 is the smallest $p$ that 200 permutations can return), so the IWAE gap, not the MMD, is what sets the cross-prior pairs apart. They are isometric (Table~\ref{tab:invariance}) without being equivalent, as Proposition~\ref{prop:alignment} allows.}
\label{tab:equivalence}
\vspace{4pt}
\scriptsize
\resizebox{\textwidth}{!}{%
\begin{tabular}{lccccccc}
\toprule
 & & \multicolumn{2}{c}{IWAE (nats)} & \multicolumn{4}{c}{Generator MMD$^2$ ($\times 10^{3}$)} \\
\cmidrule(lr){3-4} \cmidrule(lr){5-8}
Setting & pairs & $a-b$ & s.e. & $a$ vs.\ $b$ & $a$ vs.\ data & $b$ vs.\ data & $p$ \\
\midrule
MNIST, Gaussian & 6 & $-4.3$--$+0.5$ & 0.6--0.8 & 0.4--1.3 & 8.0--10.5 & 8.1--8.7 & 0.005--0.010 \\
MNIST, VampPrior & 6 & $-0.6$--$+3.9$ & 0.6--0.8 & 0.8--1.6 & 2.5--2.7 & 2.3--2.9 & 0.005 \\
MNIST, cross-prior & 4 & $+22.3$--$+25.2$ & 0.7--0.9 & 2.6--5.2 & 2.5--2.7 & 7.4--8.7 & 0.005 \\
FashionMNIST, Gaussian & 6 & $-1.0$--$+1.6$ & 0.3--0.4 & 0.5--1.2 & 13.7--16.5 & 12.7--13.7 & 0.005--0.055 \\
FashionMNIST, VampPrior & 6 & $-1.1$--$+0.2$ & 0.3--0.4 & 0.0--0.7 & 3.1--3.3 & 2.6--3.0 & 0.030--0.428 \\
FashionMNIST, cross-prior & 4 & $+10.2$--$+11.1$ & 0.5 & 6.2--7.3 & 3.1--3.3 & 12.7--13.2 & 0.005 \\
\bottomrule
\end{tabular}
}
\end{table}

%% file: tables/invariance.tex
\begin{table}[t]
\centering
\caption{Pairwise distances in one model against the same pairs in the other, under the true correspondence, on the samples with $100\%$ overlap (one draw, $1500$ representations). Pullback distances are segment lengths under the decoder ensemble (Equation~\ref{eq:chord}). \emph{corr.}\ is the Pearson correlation of the two distance matrices, \emph{scale} the factor by which one exceeds the other, and \emph{same neighbour} the share of images whose nearest neighbour is the same image in both models. Ranges are over pairings. Pullback distances agree across all 32 pairs (corr.\ 0.973--0.996, scale 0.98--1.01). Euclidean distances do not on the cross-prior pairs, with corr.\ 0.71, 0.61 and scale 2.92, 5.01 (MNIST, FashionMNIST). A random pairing gives $|$corr.$|\le 0.01$.}
\label{tab:invariance}
\vspace{4pt}
\small
\resizebox{\textwidth}{!}{%
\begin{tabular}{lccccccc}
\toprule
 & & \multicolumn{3}{c}{Pullback distances} & \multicolumn{3}{c}{Euclidean distances} \\
\cmidrule(lr){3-5} \cmidrule(lr){6-8}
Setting & pairs & corr. & scale & same neighbour & corr. & scale & same neighbour \\
\midrule
MNIST, Gaussian & 6 & 0.987--0.991 & 0.98--0.99 & 0.84--0.87 & 0.903--0.941 & 0.99--1.00 & 0.58--0.63 \\
MNIST, VampPrior & 6 & 0.992--0.994 & 0.99--1.00 & 0.86--0.90 & 0.961--0.973 & 0.98--1.02 & 0.74--0.75 \\
MNIST, cross-prior & 4 & 0.976--0.983 & 1.00--1.01 & 0.84--0.86 & 0.696--0.714 & 2.89--2.95 & 0.43--0.44 \\
FashionMNIST, Gaussian & 6 & 0.979--0.989 & 1.00 & 0.81--0.84 & 0.905--0.928 & 0.99--1.00 & 0.57--0.60 \\
FashionMNIST, VampPrior & 6 & 0.994--0.996 & 0.98--1.01 & 0.85--0.86 & 0.972--0.978 & 1.00--1.02 & 0.72--0.77 \\
FashionMNIST, cross-prior & 4 & 0.973--0.976 & 0.99--1.00 & 0.79--0.81 & 0.598--0.623 & 4.98--5.04 & 0.31--0.33 \\
\bottomrule
\end{tabular}
}
\end{table}

%% file: figures/coupling_overlap_all.tex
\begin{figure}[t]
\centering
\includegraphics[width=\textwidth]{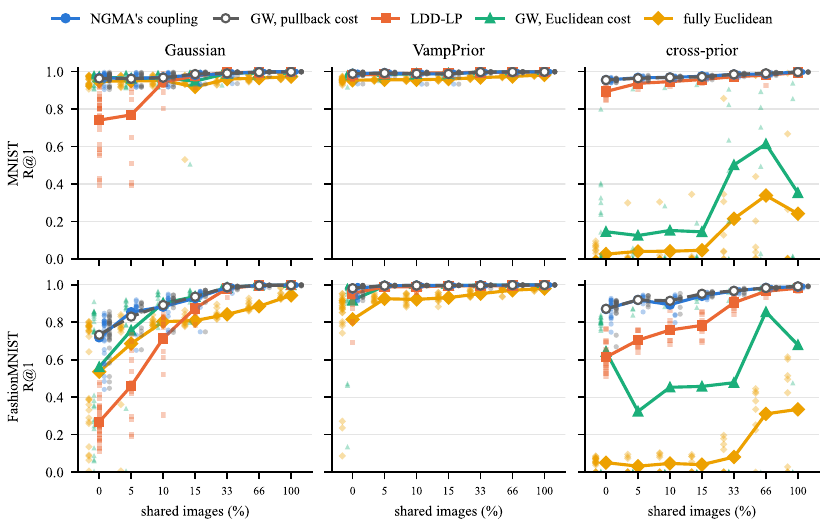}
\caption{NGMA's R@1 against the share of images the two samples have in common, fitted from five couplings, every MNIST and FashionMNIST setting. Lines join the means over pairings and draws, and dots are single draws. At $0\%$ the samples are the disjoint samples of Table~\ref{tab:alignment}, with five draws per pairing; every other share has two. The three GW couplings are plain entropic GW, without NGMA's LDD-LP stages. With all images shared, NGMA's coupling reaches 0.990--0.999 and GW on the pullback cost 0.991--0.999 in every setting, and GW on the Euclidean cost reaches 0.353 on the MNIST cross-prior pair and 0.679 on the FashionMNIST cross-prior pair. The LDD-LP needs shared images where LDDs are poorly separated, as on FashionMNIST with Gaussian priors (0.268 at $0\%$, 0.874 at $15\%$ and 0.980 at $33\%$).}
\label{fig:coupling-overlap-all}
\end{figure}

%% file: tables/shared_baselines.tex
\begin{table}[p]
\centering
\caption{R@1 against the share of images the two samples have in common, every method on every MNIST and FashionMNIST pairing: the mean over a setting's pairings of the mean over draws. At $0\%$ the samples are the disjoint samples of Table~\ref{tab:alignment}, with five draws per pairing (two for RelReps and RelGeo); every other share has two draws per pairing. RelReps and RelGeo are label-informed. Rosetta and mini-vec2vec use no pairs and no labels, like NGMA. The rigid map fitted on the true pairs does not depend on the share, so each setting's header gives its R@1. Rosetta and mini-vec2vec centre and normalise the codes before fitting their rotation, so they can exceed it. The best R@1 in each column of a setting is in bold. NGMA has the highest R@1 in every column of every setting. On single pairings, a baseline matches or exceeds NGMA at 4 of 896 points (method, pairing, share): Rosetta on FashionMNIST, Gaussian ac at 0\% (0.652 against 0.537); Rosetta on MNIST, Gaussian ac at 5\% (0.921 against 0.917); Rosetta on MNIST, Gaussian ac at 10\% (0.926 against 0.919); Rosetta on MNIST, Gaussian ad at 15\% (0.934 against 0.932).}
\label{tab:shared}
\vspace{4pt}
\small
\begin{tabular}{lccccccc}
\toprule
 & \multicolumn{7}{c}{Share of images in common (\%)} \\
\cmidrule(lr){2-8}
Method & 0 & 5 & 10 & 15 & 33 & 66 & 100 \\
\midrule
\multicolumn{8}{l}{\emph{MNIST, Gaussian}; rigid map on the true pairs 0.955} \\
NGMA & \textbf{0.967} & \textbf{0.963} & \textbf{0.969} & \textbf{0.982} & \textbf{0.989} & \textbf{0.999} & \textbf{0.999} \\
RelReps & 0.525 & 0.531 & 0.531 & 0.539 & 0.549 & 0.573 & 0.600 \\
RelGeo & 0.512 & 0.491 & 0.485 & 0.482 & 0.451 & 0.496 & 0.476 \\
Rosetta & 0.941 & 0.945 & 0.950 & 0.952 & 0.956 & 0.957 & 0.958 \\
mini-vec2vec & 0.849 & 0.805 & 0.861 & 0.792 & 0.862 & 0.887 & 0.945 \\
\midrule
\multicolumn{8}{l}{\emph{MNIST, VampPrior}; rigid map on the true pairs 0.936} \\
NGMA & \textbf{0.989} & \textbf{0.992} & \textbf{0.988} & \textbf{0.988} & \textbf{0.997} & \textbf{0.998} & \textbf{0.999} \\
RelReps & 0.297 & 0.299 & 0.300 & 0.299 & 0.301 & 0.300 & 0.305 \\
RelGeo & 0.552 & 0.665 & 0.673 & 0.661 & 0.672 & 0.722 & 0.759 \\
Rosetta & 0.930 & 0.927 & 0.929 & 0.929 & 0.931 & 0.930 & 0.931 \\
mini-vec2vec & 0.922 & 0.921 & 0.922 & 0.922 & 0.925 & 0.926 & 0.927 \\
\midrule
\multicolumn{8}{l}{\emph{MNIST, cross-prior}; rigid map on the true pairs 0.512} \\
NGMA & \textbf{0.955} & \textbf{0.967} & \textbf{0.969} & \textbf{0.975} & \textbf{0.985} & \textbf{0.988} & \textbf{0.998} \\
RelReps & 0.004 & 0.003 & 0.003 & 0.003 & 0.002 & 0.003 & 0.004 \\
RelGeo & 0.319 & 0.351 & 0.353 & 0.342 & 0.349 & 0.347 & 0.385 \\
Rosetta & 0.571 & 0.581 & 0.590 & 0.597 & 0.615 & 0.626 & 0.633 \\
mini-vec2vec & 0.393 & 0.463 & 0.460 & 0.407 & 0.520 & 0.506 & 0.534 \\
\midrule
\multicolumn{8}{l}{\emph{FashionMNIST, Gaussian}; rigid map on the true pairs 0.782} \\
NGMA & \textbf{0.719} & \textbf{0.854} & \textbf{0.883} & \textbf{0.928} & \textbf{0.986} & \textbf{0.996} & \textbf{0.998} \\
RelReps & 0.151 & 0.160 & 0.161 & 0.161 & 0.164 & 0.172 & 0.186 \\
RelGeo & 0.159 & 0.148 & 0.154 & 0.153 & 0.159 & 0.165 & 0.193 \\
Rosetta & 0.710 & 0.724 & 0.728 & 0.732 & 0.745 & 0.756 & 0.764 \\
mini-vec2vec & 0.504 & 0.663 & 0.582 & 0.629 & 0.586 & 0.652 & 0.699 \\
\midrule
\multicolumn{8}{l}{\emph{FashionMNIST, VampPrior}; rigid map on the true pairs 0.747} \\
NGMA & \textbf{0.979} & \textbf{0.995} & \textbf{0.996} & \textbf{0.996} & \textbf{0.997} & \textbf{0.999} & \textbf{0.999} \\
RelReps & 0.062 & 0.062 & 0.061 & 0.063 & 0.065 & 0.068 & 0.066 \\
RelGeo & 0.355 & 0.354 & 0.363 & 0.369 & 0.385 & 0.411 & 0.435 \\
Rosetta & 0.656 & 0.685 & 0.689 & 0.691 & 0.700 & 0.705 & 0.712 \\
mini-vec2vec & 0.624 & 0.639 & 0.644 & 0.650 & 0.662 & 0.669 & 0.680 \\
\midrule
\multicolumn{8}{l}{\emph{FashionMNIST, cross-prior}; rigid map on the true pairs 0.071} \\
NGMA & \textbf{0.871} & \textbf{0.919} & \textbf{0.894} & \textbf{0.941} & \textbf{0.967} & \textbf{0.984} & \textbf{0.990} \\
RelReps & 0.002 & 0.002 & 0.002 & 0.002 & 0.002 & 0.002 & 0.002 \\
RelGeo & 0.098 & 0.107 & 0.101 & 0.101 & 0.097 & 0.104 & 0.082 \\
Rosetta & 0.065 & 0.073 & 0.074 & 0.074 & 0.079 & 0.081 & 0.085 \\
mini-vec2vec & 0.018 & 0.016 & 0.013 & 0.016 & 0.019 & 0.019 & 0.025 \\
\bottomrule
\end{tabular}
\end{table}

%% file: tables/mirror_test.tex
\begin{table}[t]
\centering
\caption{The mirror test. Every test image is flipped left to right and encoded by both models. True R@1 scores the image of a representation under a map against the representations of the unflipped images, and mirror R@1 against those of the flipped images, with the image's own mirror as the partner. The self-mirror rate is the fraction of images whose own mirror is the nearest flipped representation in the same model. Ranges are over draws. The 70 CelebA alignments are the seven pairings with draws 1--5, each from NGMA's coupling and from GW alone, and a draw fails when its R@1 is below $0.3$. The controls are MNIST and FashionMNIST, Gaussian pairing $ab$, draws 1--5.}
\label{tab:mirror}
\vspace{4pt}
\scriptsize
\resizebox{\textwidth}{!}{%
\begin{tabular}{lcccccc}
\toprule
 & & \multicolumn{2}{c}{$a\to b$} & \multicolumn{2}{c}{$b\to a$} & \\
\cmidrule(lr){3-4} \cmidrule(lr){5-6}
Draws & $n$ & true R@1 & mirror R@1 & true R@1 & mirror R@1 & self-mirror rate \\
\midrule
\multicolumn{7}{l}{\emph{CelebA, failed draws}} \\
NGMA's coupling & 7 & 0.054--0.113 & 0.905--0.992 & 0.055--0.111 & 0.910--0.992 & 0.101--0.116 \\
GW alone & 10 & 0.052--0.113 & 0.870--0.993 & 0.053--0.111 & 0.893--0.993 & 0.101--0.116 \\
\midrule
\multicolumn{7}{l}{\emph{CelebA, successful draws}} \\
NGMA's coupling & 28 & 0.948--0.995 & 0.057--0.106 & 0.947--0.994 & 0.057--0.103 & 0.101--0.116 \\
GW alone & 25 & 0.948--0.995 & 0.057--0.107 & 0.947--0.994 & 0.057--0.103 & 0.101--0.116 \\
\midrule
\multicolumn{7}{l}{\emph{Controls, both couplings}} \\
MNIST, Gaussian & 10 & 0.873--0.916 & 0.004--0.005 & 0.891--0.922 & 0.005--0.006 & 0.007 \\
FashionMNIST, Gaussian & 10 & 0.595--0.745 & 0.155--0.210 & 0.615--0.760 & 0.176--0.213 & 0.276--0.305 \\
\bottomrule
\end{tabular}
}
\end{table}

%% file: tables/full-results/scvi_lung.tex
\begin{table}[t]
\centering
\caption{scVI models of Tabula Sapiens lung cells (Appendix~\ref{app:lung}): NGMA against the unpaired baselines, on the disjoint samples ($n=3000$ cells per side) of two models trained on different cells or on the same cells with different seeds, by training size. Retrieval of the true partner among the $10{,}000$ evaluation cells (R@1, chance $10^{-4}$) and cell-type accuracy of the nearest cell (chance 0.124), scored as in Table~\ref{tab:alignment}. GENOT-Q runs on its own GW coupling with the Euclidean cost or with the pullback cost NGMA uses. NGMA, one decoder, reads the pullback metric of the scVI decoder alone instead of the ensemble of eight. The last two rows of each block are not methods. They are NGMA's map class and the rigid map, both fitted on the true pairs and scored $a \to b$. Mean $\pm$ s.d.\ over two pairings and three draws; the best method in each column is in bold. $^\dagger$RelGeo builds its anchors from the cell-type labels of both models.}
\label{tab:lung_baselines}
\vspace{4pt}
\small
\setlength{\tabcolsep}{3.5pt}
\begin{tabular}{lcccccc}
\toprule
 & \multicolumn{3}{c}{different cells} & \multicolumn{3}{c}{same cells} \\
\cmidrule(lr){2-4}\cmidrule(lr){5-7}
Training size & 4{,}900 & 9{,}800 & 19{,}600 & 4{,}900 & 9{,}800 & 19{,}600 \\
\midrule
\multicolumn{7}{l}{\emph{R@1}} \\
NGMA & \textbf{0.221}{\scriptsize$\pm$0.035} & \textbf{0.369}{\scriptsize$\pm$0.026} & \textbf{0.378}{\scriptsize$\pm$0.015} & \textbf{0.274}{\scriptsize$\pm$0.023} & \textbf{0.401}{\scriptsize$\pm$0.031} & \textbf{0.480}{\scriptsize$\pm$0.016} \\
NGMA, Euclidean & 0.119{\scriptsize$\pm$0.064} & 0.002{\scriptsize$\pm$0.002} & 0.047{\scriptsize$\pm$0.066} & 0.062{\scriptsize$\pm$0.058} & 0.181{\scriptsize$\pm$0.118} & 0.146{\scriptsize$\pm$0.146} \\
NGMA, one decoder & 0.195{\scriptsize$\pm$0.031} & 0.348{\scriptsize$\pm$0.020} & 0.294{\scriptsize$\pm$0.075} & 0.257{\scriptsize$\pm$0.013} & 0.372{\scriptsize$\pm$0.050} & 0.389{\scriptsize$\pm$0.073} \\
\addlinespace[2pt]
\multicolumn{7}{l}{\quad unpaired baselines} \\
Rosetta & 0.077{\scriptsize$\pm$0.033} & 0.141{\scriptsize$\pm$0.020} & 0.128{\scriptsize$\pm$0.011} & 0.088{\scriptsize$\pm$0.024} & 0.154{\scriptsize$\pm$0.025} & 0.210{\scriptsize$\pm$0.010} \\
mini-vec2vec & 0.028{\scriptsize$\pm$0.021} & 0.062{\scriptsize$\pm$0.065} & 0.060{\scriptsize$\pm$0.039} & 0.027{\scriptsize$\pm$0.035} & 0.071{\scriptsize$\pm$0.063} & 0.124{\scriptsize$\pm$0.027} \\
GENOT-Q, Eucl.\ cost & 0.028{\scriptsize$\pm$0.019} & 0.001{\scriptsize$\pm$0.001} & 0.002{\scriptsize$\pm$0.003} & 0.020{\scriptsize$\pm$0.016} & 0.007{\scriptsize$\pm$0.010} & 0.015{\scriptsize$\pm$0.015} \\
GENOT-Q, pullback cost & 0.024{\scriptsize$\pm$0.007} & 0.030{\scriptsize$\pm$0.001} & 0.027{\scriptsize$\pm$0.001} & 0.028{\scriptsize$\pm$0.002} & 0.033{\scriptsize$\pm$0.005} & 0.032{\scriptsize$\pm$0.002} \\
\addlinespace[2pt]
\multicolumn{7}{l}{\quad label-informed} \\
RelGeo$^\dagger$ & 0.027{\scriptsize$\pm$0.005} & 0.045{\scriptsize$\pm$0.012} & 0.047{\scriptsize$\pm$0.013} & 0.036{\scriptsize$\pm$0.005} & 0.059{\scriptsize$\pm$0.009} & 0.051{\scriptsize$\pm$0.024} \\
\addlinespace[2pt]
\multicolumn{7}{l}{\quad fitted on the true pairs} \\
map & 0.363{\scriptsize$\pm$0.036} & 0.510{\scriptsize$\pm$0.005} & 0.624{\scriptsize$\pm$0.009} & 0.466{\scriptsize$\pm$0.019} & 0.621{\scriptsize$\pm$0.011} & 0.692{\scriptsize$\pm$0.007} \\
rigid map & 0.099{\scriptsize$\pm$0.029} & 0.185{\scriptsize$\pm$0.006} & 0.181{\scriptsize$\pm$0.003} & 0.110{\scriptsize$\pm$0.020} & 0.179{\scriptsize$\pm$0.014} & 0.254{\scriptsize$\pm$0.007} \\
\midrule
\multicolumn{7}{l}{\emph{cell-type accuracy}} \\
NGMA & \textbf{0.891}{\scriptsize$\pm$0.023} & \textbf{0.924}{\scriptsize$\pm$0.008} & \textbf{0.931}{\scriptsize$\pm$0.006} & 0.893{\scriptsize$\pm$0.011} & 0.911{\scriptsize$\pm$0.016} & \textbf{0.938}{\scriptsize$\pm$0.014} \\
NGMA, Euclidean & 0.701{\scriptsize$\pm$0.305} & 0.062{\scriptsize$\pm$0.052} & 0.449{\scriptsize$\pm$0.278} & 0.567{\scriptsize$\pm$0.239} & 0.667{\scriptsize$\pm$0.209} & 0.443{\scriptsize$\pm$0.423} \\
NGMA, one decoder & 0.883{\scriptsize$\pm$0.025} & 0.923{\scriptsize$\pm$0.006} & 0.897{\scriptsize$\pm$0.046} & \textbf{0.897}{\scriptsize$\pm$0.008} & \textbf{0.914}{\scriptsize$\pm$0.019} & 0.915{\scriptsize$\pm$0.034} \\
\addlinespace[2pt]
\multicolumn{7}{l}{\quad unpaired baselines} \\
Rosetta & 0.773{\scriptsize$\pm$0.028} & 0.775{\scriptsize$\pm$0.044} & 0.789{\scriptsize$\pm$0.015} & 0.766{\scriptsize$\pm$0.048} & 0.774{\scriptsize$\pm$0.008} & 0.814{\scriptsize$\pm$0.019} \\
mini-vec2vec & 0.563{\scriptsize$\pm$0.169} & 0.577{\scriptsize$\pm$0.205} & 0.636{\scriptsize$\pm$0.155} & 0.487{\scriptsize$\pm$0.245} & 0.497{\scriptsize$\pm$0.288} & 0.727{\scriptsize$\pm$0.079} \\
GENOT-Q, Eucl.\ cost & 0.633{\scriptsize$\pm$0.193} & 0.144{\scriptsize$\pm$0.100} & 0.346{\scriptsize$\pm$0.282} & 0.551{\scriptsize$\pm$0.240} & 0.337{\scriptsize$\pm$0.271} & 0.397{\scriptsize$\pm$0.350} \\
GENOT-Q, pullback cost & 0.777{\scriptsize$\pm$0.014} & 0.774{\scriptsize$\pm$0.018} & 0.768{\scriptsize$\pm$0.003} & 0.773{\scriptsize$\pm$0.003} & 0.771{\scriptsize$\pm$0.015} & 0.768{\scriptsize$\pm$0.007} \\
\addlinespace[2pt]
\multicolumn{7}{l}{\quad label-informed} \\
RelGeo$^\dagger$ & 0.825{\scriptsize$\pm$0.013} & 0.848{\scriptsize$\pm$0.027} & 0.852{\scriptsize$\pm$0.011} & 0.845{\scriptsize$\pm$0.011} & 0.858{\scriptsize$\pm$0.008} & 0.854{\scriptsize$\pm$0.007} \\
\addlinespace[2pt]
\multicolumn{7}{l}{\quad fitted on the true pairs} \\
map & 0.935{\scriptsize$\pm$0.006} & 0.950{\scriptsize$\pm$0.001} & 0.960{\scriptsize$\pm$0.003} & 0.943{\scriptsize$\pm$0.002} & 0.960{\scriptsize$\pm$0.002} & 0.968{\scriptsize$\pm$0.001} \\
rigid map & 0.793{\scriptsize$\pm$0.025} & 0.800{\scriptsize$\pm$0.035} & 0.820{\scriptsize$\pm$0.010} & 0.802{\scriptsize$\pm$0.017} & 0.792{\scriptsize$\pm$0.006} & 0.825{\scriptsize$\pm$0.019} \\
\bottomrule
\end{tabular}
\end{table}